\documentclass[11pt]{article}

\usepackage{amsmath,amsbsy,amsfonts,amssymb,amsthm,dsfont,fullpage,units}
\usepackage{graphicx,psfrag,epsfig,epsf}
\usepackage{algorithm,algorithmic}
\usepackage{mathtools}

\usepackage{color,cases}
\usepackage{tikz}
\usetikzlibrary{fit}
\usetikzlibrary{calc,shapes}
\usetikzlibrary{positioning}
\usepackage{enumitem}
\usepackage[round]{natbib}
\usepackage{hyperref}
\usepackage{sublabel}
\usepackage{appendix}

\usepackage[caption=false]{subfig}
\renewcommand{\algorithmicrequire}{\textbf{Input:}}
\renewcommand{\algorithmicensure}{\textbf{Output:}}

 \newcommand{\IGNORE}[1]{}

\def\nn{\nonumber}

\newcommand\E{\mathbb{E}}
\newcommand\R{\mathbb{R}}

\def\Pc{\mathcal{P}}

\def\Pc{{\cal S}}

\newcommand\tl{\tilde}

\newcommand\poly{\operatorname{poly}}

\def\tl{\tilde}

\newcommand\inner[1]{\ensuremath{\langle #1 \rangle}}

 \DeclareMathOperator*{\argmin}{arg\,min}

\def\tha{{\mbox{\tiny th}}}

\DeclareMathOperator{\Diag}{Diag}

 \def\0{{\bf 0}}

\def\nn{\nonumber}

\def\qed{\hfill\hbox{${\vcenter{\vbox{
    \hrule height 0.4pt\hbox{\vrule width 0.4pt height 6pt
    \kern5pt\vrule width 0.4pt}\hrule height 0.4pt}}}$}}

\def\tcr{\textcolor{red}}
\def\tcb{\textcolor{blue}}

\definecolor{myred}{rgb}{0.3,0.0,0.7}
\definecolor{dkg}{rgb}{0.1,0.7,0.2}
\definecolor{dkb}{rgb}{0.0,0.2,0.8}
\definecolor{brm}{rgb}{1,0.0,1}

\def\tcdkg{\textcolor{dkg}}
\def\tcbrm{\textcolor{brm}}

 \def\ha{\hat{a}}
 \def\hb{\hat{b}}

 \def\hf{\hat{f}}
 
 \def\hh{\hat{h}}

 \def\hv{\widehat{v}}

 \def\hy{\hat{y}}
 
 \def\hA{\hat{A}}

 \def\hM{\widehat{M}}

\def\hT{\widehat{T}}

\def\hW{\widehat{W}}

\def\Sc{{\cal S}}

\def\Ebb{{\mathbb E}}

\def\Rbb{{\mathbb R}}

\newcommand{\bprfof}{\begin{proof_of}}
\newcommand{\eprfof}{\end{proof_of}}
\newcommand{\bprf}{\begin{myproof}}
\newcommand{\eprf}{\end{myproof}}
\newcommand{\bp}{\begin{psfrags}}
\newcommand{\ep}{\end{psfrags}}
\newcommand{\bl}{\begin{lemma}}
\newcommand{\el}{\end{lemma}}
\newcommand{\bt}{\begin{theorem}}
\newcommand{\et}{\end{theorem}}
\newcommand{\bc}{\begin{center}}
\newcommand{\ec}{\end{center}}
\newcommand{\bi}{\begin{itemize}}
\newcommand{\ei}{\end{itemize}}
\newcommand{\ben}{\begin{enumerate}}
\newcommand{\een}{\end{enumerate}}
\newcommand{\bd}{\begin{definition}}
\newcommand{\ed}{\end{definition}}
\def\beq{\begin{equation}}
\def\eeq{\end{equation}\noindent}
\def\beqn{\begin{eqnarray}}
\def\eeqn{\end{eqnarray} \noindent}
\def\beqnn{  \begin{eqnarray*}}
\def\eeqnn{\end{eqnarray*}  \noindent}
\def\bcase{  \begin{numcases}}
\def\ecase{\end{numcases}   \noindent}
\def\bsbcase{  \begin{subnumcases}}
\def\esbcase{\end{subnumcases}   \noindent}

\newtheorem{theorem}{Theorem}
\newtheorem{corollary}{Corollary}
\newtheorem{lemma}[theorem]{Lemma}

\newtheorem{definition}{Definition}

\newtheorem{remark}{Remark}

\newenvironment{myproof}{\noindent{\bf Proof:} \hspace*{1em}}{
    \hspace*{\fill} $\Box$ }
\newenvironment{proof_of}[1]{\noindent {\bf Proof of #1: }}{\hspace*{\fill} $\Box$ }

\newcommand{\matplottc}[1]{               
        \unitlength .45truein
        \begin{center}
        \includegraphics{#1.ps}
        \end{picture}
        \end{center}
}

\def\psfancypar#1#2{\begingroup\def\par{\endgraf\endgroup\lineskiplimit=0pt}
               \setbox2=\hbox{\large\sc #2}
               \newdimen\tmpht \tmpht \ht2 \advance\tmpht by \baselineskip
               \font\hhuge=Times-Bold at \tmpht
               \setbox1=\hbox{{\hhuge #1}}
               \count7=\tmpht \count8=\ht1
               \divide\count8 by 1000 \divide\count7 by \count8
               \tmpht=.001\tmpht\multiply\tmpht by \count7
               \font\hhuge=Times-Bold at \tmpht
               \setbox1=\hbox{{\hhuge #1}}
               \noindent
                \hangindent1.05\wd1
               \hangafter=-2 {\hskip-\hangindent
               \lower1\ht1\hbox{\raise1.0\ht2\copy1}%
                \kern-0\wd1}\copy2\lineskiplimit=-1000pt}

\def\Kout{\setbox1=\hbox{\Huge\bf K}\hbox to
1.05\wd1{\hspace{.05\wd1}
}}
\def\Sout{\setbox1=\hbox{\Huge\bf S}\hbox to 1.05\wd1{\hspace{.05\wd1}
}}

\newcommand{\torestate}[3]{%
\expandafter \def \csname BBRESTATE #2 \endcsname{#3}
\theoremstyle{plain}
\newtheorem{BBRESTATETHMNUM#2}[theorem]{#1}
\begin{BBRESTATETHMNUM#2}\label{#2}\csname BBRESTATE #2 \endcsname   \end{BBRESTATETHMNUM#2}
\newtheorem*{BBRESTATETHMNONNUM#2}{{#1}~\ref{#2}}
}

\newcommand{\restate}[1]{\begin{BBRESTATETHMNONNUM#1}[Restated] \csname BBRESTATE #1 \endcsname
\end{BBRESTATETHMNONNUM#1}}

\definecolor{blue1}{HTML}{0066FF}
\definecolor{lpurple}{cmyk}{.05,0.18,0,0}

\date{}
\author{Majid Janzamin\footnote{University of California, Irvine. Email: mjanzami@uci.edu} \and Hanie Sedghi\footnote{University of California, Irvine. Email: sedghih@uci.edu} \and Anima Anandkumar\footnote{University of California, Irvine. Email: a.anandkumar@uci.edu}}

\title{Beating the Perils of Non-Convexity: \\ Guaranteed Training of Neural
Networks using Tensor Methods}

\begin{document}
\maketitle

\begin{abstract}
Training  neural networks is a challenging non-convex optimization problem, and backpropagation or gradient descent can get stuck in spurious local optima. We propose a novel algorithm based on tensor decomposition   for guaranteed training of  two-layer neural networks. We provide risk bounds for our proposed method, with a polynomial sample complexity in the relevant parameters, such as input dimension and number of neurons. While learning arbitrary target functions is NP-hard, we provide transparent conditions on the function and the input for learnability. Our training method is based on tensor decomposition, which provably converges  to the global optimum, under a set of mild non-degeneracy conditions. It consists of simple embarrassingly parallel linear and multi-linear operations, and is competitive with standard stochastic gradient descent (SGD), in terms of computational complexity. Thus, we propose a computationally efficient method with guaranteed risk bounds for training neural networks with one hidden layer.
\end{abstract}




\section{Introduction}

Neural networks have revolutionized performance across multiple domains such as computer vision and speech recognition. They are flexible models   trained   to approximate any arbitrary target function, e.g., the label function for classification tasks. They  are composed of multiple layers of {\em neurons} or {\em activating functions}, which are applied recursively on the input data, in order to predict the output. While neural networks have been extensively employed in practice,  a complete theoretical understanding  is currently lacking.

Training a neural network can be framed as an optimization problem, where the network parameters are chosen to minimize   a given loss function, e.g.,   the {\em quadratic loss}  function over the error in predicting the output. The performance of training algorithms   is typically measured through the notion of {\em risk}, which is the expected loss function over unseen test data. A natural question to ask is the hardness of   training a neural network with a bounded risk. The findings are mostly negative~\citep{rojas1996neural,vsima2002training,blum1993training,bartlett1999hardness,kuhlmann2000hardness}. Training even a simple network  is NP-hard, e.g.,  a network with a single neuron~\citep{vsima2002training}. 


The computational hardness of training is due to the non-convexity of the loss function. In general, the loss function has many {\em critical points}, which include {\em spurious local optima} and {\em saddle points}. In addition, we face {\em curse of dimensionality}, and the number of critical points grows exponentially with the input dimension for general non-convex problems~\citep{dauphin2014identifying}. Popular local search methods  such as gradient descent or {\em backpropagation}    can get stuck in bad local optima and experience arbitrarily  slow convergence. Explicit examples of its failure and the presence of bad local optima in even simple separable settings  have been documented before~\citep{brady1989back,gori1992problem,frasconi1997successes};  see Section~\ref{sec:toy-example} for a discussion.

Alternative methods for training neural networks have been mostly limited to specific activation functions (e.g., linear or quadratic), specific target functions (e.g., polynomials)~\citep{andoni2014learning}, or assume strong assumptions on the input (e.g., Gaussian or product distribution)~\citep{andoni2014learning},  see related work for details.
Thus, up until now,   there is no unified framework for training networks with   general  input, output and activation functions, for  which we can provide  guaranteed risk bound.

%
%
%

In this paper, for the first time, we present a guaranteed framework for learning general target functions using neural networks, and simultaneously overcome computational, statistical, and approximation challenges. In other words, our method has a low computational and sample complexity, even as the dimension of the optimization grows, and in addition, can also handle approximation errors, when the target function may not be generated by a given neural network. We  prove a  guaranteed risk bound for our proposed method.  NP-hardness   refers to the computational complexity of training worst-case instances. Instead,  we provide transparent conditions on the target functions and the inputs for tractable  learning.

Our   training method is  based on the method of moments, which involves  decomposing the empirical cross moment between output and some function of input.
While pairwise  moments are represented using a matrix, higher order moments require tensors, and the learning problem can be formulated as tensor decomposition. A CP (CanDecomp/Parafac) decomposition of a tensor involves finding a succinct  sum of  rank-one components that best fit the input tensor. Even though it  is a non-convex problem, the global optimum of tensor decomposition can be achieved using computationally efficient techniques, under a set of mild non-degeneracy conditions~\citep{JanzaminEtal:Altmin14,JMLR:v15:anandkumar14b,anandkumar2014guaranteed,
Overcomplete:COLT2015,bhaskara2013smoothed}. These methods   have been recently employed for learning a wide range of  latent variable models~\citep{JMLR:v15:anandkumar14b,AnandkumarEtal:community12COLT}.

Incorporating tensor methods for training neural networks requires addressing a number of non-trivial questions: What form of moments are informative about network parameters? Earlier works using tensor methods for learning assume a linear relationship between the hidden and observed variables. However, neural networks possess non-linear activation functions. How do we adapt tensor methods for this setting? How do these methods behave in the presence of approximation and sample perturbations? How can we establish risk bounds? We address these questions shortly.

\subsection{Summary of Results}

The main contributions are: (a) we propose an efficient algorithm for training neural networks, termed as  Neural Network-LearnIng using Feature Tensors (NN-LIFT), (b) we demonstrate that the method is embarrassingly parallel and is competitive with standard SGD in terms of computational complexity, and as a main result,  (c) we establish that it has bounded risk, when the number of training samples   scales polynomially in relevant parameters such as input dimension and number of neurons.

 %

We analyze training of a two-layer feedforward neural network, where the second layer has a linear activation function. This is the classical neural network considered in a number of works~\citep{cybenko1989approximation,hornik1991approximation,barron1994approximation}, and a natural starting point for the analysis of any learning algorithm.  Note that training even this two-layer network is non-convex, and finding a computationally efficient method with guaranteed risk bound has been an open problem up until now.

At a high level,    NN-LIFT   estimates the weights of the first layer using tensor CP decomposition. It then uses these estimates to learn   the bias parameter of first layer using a simple Fourier technique, and finally estimates the parameters of last layer using linear regresion. NN-LIFT consists of  simple linear and multi-linear operations~\citep{JMLR:v15:anandkumar14b,anandkumar2014guaranteed,Overcomplete:COLT2015}, Fourier analysis and ridge regression analysis, which are parallelizable to large-scale data sets. The computational complexity is comparable to that of the standard SGD; in fact, the parallel time complexity for both the methods is  in the same order, and our method requires more processors than SGD by a multiplicative factor that scales linearly in the input dimension.

\paragraph{Generative vs.\ discriminative models:}Generative models incorporate a joint distribution $p(x,y)$ over both the input $x$ and label $y$. On the other hand, discriminative models such as neural networks only incorporate the conditional distribution $p(y|x)$. While training neural networks for general input $x$ is NP-hard, {\bf does  knowledge about the input distribution $p(x)$ make  learning tractable?}

In this work, we assume knowledge of   the input density $p(x)$, which can be any continuous differentiable function.
Unlike many theoretical works, e.g.,~\citep{andoni2014learning}, we do not limit ourselves to distributions such as product or Gaussian distributions for the input. While unsupervised learning, i.e., estimation of density $p(x)$, is itself a hard problem for general models, in this work, we investigate how $p(x)$ can be exploited to make training of neural networks tractable. The knowledge of $p(x)$ is naturally available in the {\em experimental design} framework, where the person designing the experiments has the ability to choose the input distribution. Examples include   conducting polling, carrying out drug trials, collecting survey information, and so on.

\paragraph{Utilizing generative models on the input via score functions:}
We utilize the knowledge about the input density $p(x)$ (up to normalization)\footnote{We do not require the knowledge of the normalizing constant or the partition function, which is $\#P$ hard to compute~\citep{wainwright2008graphical}.} to obtain certain (non-linear) transformations of the input, given by the class of score functions.   {\em Score functions} are normalized derivatives of the input pdf; see~\eqref{eqn:diffoperator}.  If the input is a vector (the typical case), the first order score function (i.e., the first  derivative) is a vector, the second order score is a matrix, and the higher order scores are tensors. 
In our NN-LIFT method, we first estimate the cross-moments between the output and the input score functions, and then decompose it to rank-1 components.







\paragraph{Risk bounds:}
Risk bound includes both approximation and estimation errors. The approximation error is the error in fitting the target function to a neural network of given architecture, and the estimation error is the error in estimating the weights of that neural network using the given samples.

We first consider the {\em realizable setting} where the target function is generated by a two-layer neural network (with hidden layer of neurons consisting of any general sigmoidal activations), and a linear output layer.
Note that the approximation error is zero in this setting.
Let $A_1 \in \R^{d \times k}$ be the weight matrix of first layer (connecting the input to the neurons) with $k$ denoting the number of neurons and $d$ denoting the input dimension. Suppose these weight vectors are non-degenerate, i.e., the weight matrix $A_1$ (or its tensorization) is full column rank. 
We assume continuous input distribution with access to  score functions, which are bounded on any set of non-zero measure. We allow for any general sigmoidal activation functions with non-zero third derivatives in expectation, and satisfying Lipschitz property.
Let $s_{\min}(\cdot)$ be the minimum singular value operator, and $M_3(x) \in \R^{d \times d^2}$ denote the matricization of input score function tensor $\Sc_3(x) \in \R^{d \times d \times d}$; see~\eqref{eqn:matricization}~and~\eqref{eqn:diffoperator} for the definitions. For  the Gaussian input
$x \sim \mathcal{N} (0,I_d)$, we have $\E \left[ \left\| M_3(x) M_3^\top(x) \right\| \right] = \tl{O} \left( d^{3} \right)$. We have the following learning result in the realizable setting where the target function is generated by a two layer neural network (with one hidden layer).

\begin{theorem}[Informal result for realizable setting]
Our method NN-LIFT learns a realizable target function up to error $\epsilon$ when the number of samples is lower bounded as\footnote{Here, only the dominant   terms in the sample complexity are noted; see~\eqref{eqn:samplecomp} for the full details.},
\begin{align*}
n \geq
\tl{O} \left(  \frac{k}{\epsilon^2} \cdot \E \left[ \left\| M_3(x) M_3^\top(x) \right\| \right]
\cdot
\frac{s_{\max}^2(A_1)}{s_{\min}^6(A_1)} \right).	
\end{align*}
\end{theorem}

Thus, we can efficiently learn the neural network parameters with polynomial sample complexity using NN-LIFT algorithm. In addition, the method has polynomial computational complexity, and in fact, its parallel time complexity is the same as stochastic gradient descent (SGD) or backpropagation. See Theorem~\ref{thm:guarantees-sample} for the formal result.

We then extend our results to the {\em non-realizable setting} where the target function {\em need not be} generated by a neural network. For our method NN-LIFT to succeed, we require the approximation error to be sufficiently small under the given network architecture. Note that it is not of practical interest to consider functions with large approximation errors, since  classification performance in that case is poor~\citep{bartlett1998sample}. We state the informal version of the result as follows.

We assume the following: the target function $f(x)$ has a continuous Fourier spectrum and is sufficiently smooth, i.e., the parameter $C_f$ (see~\eqref{eqn:C_f} for the definition) is sufficiently small as specified in~\eqref{eqn:Cf-bound}. This implies that the approximation error of the target function  can be controlled, i.e., there exists a neural network of given size that can fit the target function with  bounded approximation error. Let the input $x$ be bounded as $\|x\| \leq r$. Our informal result is as follows.
See Theorem~\ref{thm:approx-guarantees} for the formal result.

\begin{theorem}[Informal result for non-realizable setting] \label{thm:approx-guarantees-informal}
The arbitrary target function $f(x)$ is approximated by the neural network $\hf(x)$ which is learnt using  NN-LIFT algorithm such that the risk bound satisfies w.h.p.\
\begin{equation*}
\Ebb_x [|f(x) - \hf(x)|^2] \leq O(r^2 C_f^2) \cdot \left(\frac{1}{\sqrt{k}} + \delta_1 \right)^2 + O(\epsilon^2),
\end{equation*}
where $k$ is the number of neurons in the neural network, and $\delta_\tau$ is defined in~\eqref{eqn:delta_tau-Def}.
\end{theorem}

In the above bound, we require  for the target  function $f(x)$  to have bounded first order moment in  the Fourier spectrum; see~\eqref{eqn:Cf-bound}. As an example,  we show that this bound is satisfied for the  class of  scale and location mixtures of the Gaussian kernel function.

\begin{corollary}[Learning  mixtures of Gaussian kernels] \label{cor:kernel}
Let $f(x) :=  \int K(\alpha(x+\beta)) G(d\alpha,d\beta)$, $\alpha>0$, $\beta \in \R^d$, be a location and scale mixture of the Gaussian kernel function  $K(x) = \exp \left( - \frac{\|x\|^2}{2} \right)$,
  the input be Gaussian as $x \sim \mathcal{N} (0,\sigma_x^2 I_d)$, and the activations be step functions,  then, our algorithm trains a neural network with risk bounds as in  Theorem~\ref{thm:approx-guarantees-informal},   when
\begin{align*}
\int |\alpha| \cdot |G|(d\alpha,d\beta) \leq
\poly \left( \frac{1}{d},\frac{1}{k}, \epsilon, \frac{1}{\sigma_x}, \exp \left(-1/\sigma_x^2\right) \right).
\end{align*}
\end{corollary}


We observe that when the kernel mixtures correspond to smoother functions (smaller $\alpha$), the above bound is more likely to be satisfied. This is intuitive since smoother functions have lower amount of high frequency content. Also, notice that the above bound has a dependence on  the variance of the Gaussian input  $\sigma_x$. We obtain the  most relaxed bound (r.h.s.\ of above bound) for middle values of $\sigma_x$, i.e.,   when $\sigma_x$ is neither too large nor too small.
See Appendix~\ref{appendix:kernel} for more discussion and the proof of the corollary.

\paragraph{Intuitions behind the conditions for the risk bound:}Since there exist worst-case instances where learning is hard, it is natural to expect that NN-LIFT has guarantees only  when certain conditions are met.   We assume that the input has a regular continuous    probability density function (pdf); see~\eqref{eqn:regularity-cond} for the details. This is a reasonable assumption, since under Boolean inputs (a special case of discrete input), it reduces to learning parity with noise which is a hard problem~\citep{kearns1994introduction}. We assume that the activating functions are {\em sufficiently non-linear}, since if they are linear,   then the network  can be collapsed into a single layer~\citep{baldi1989neural}, which is non-identifiable.  We precisely characterize how the   estimation error   depends on the non-linearity of the activating function through its third order derivative.

Another condition for providing the risk bound is  non-redundancy of the neurons. If the neurons are redundant, it is an over-specified network. In the realizable setting, where the target function is generated by a neural network with the given number of neurons $k$, we require   (tensorizations of) the weights of first layer to be linearly independent. In the non-realizable setting, we require this to be satisfied by   $k$ vectors randomly  drawn from the Fourier magnitude distribution (weighted by the norm of frequency vector) of the target function $f(x)$. More precisely, the random frequencies are drawn from probability distribution $\Lambda(\omega) := \|\omega\| \cdot |F(\omega)|/C_f$ where $F(\omega)$ is the Fourier transform of arbitrary function $f(x)$, and $C_f$ is the normalization factor; see~\eqref{eqn:Lambda-distribution} and corresponding discussions for more details. This is a mild condition which holds when the   distribution is continuous in some domain. Thus,  our conditions for achieving bounded risk are mild and encompass a large class of target functions and input distributions. 

\paragraph{Why tensors are required?} We employ the cross-moment tensor which encodes the correlation between the third order score function and the output. We then decompose the moment tensor as a sum of rank-$1$ components to yield the weight vectors of the first layer. We require at least a third order tensor to learn the neural network weights for the following reasons: while  a matrix decomposition is only identifiable up to orthogonal components, tensors can have identifiable non-orthogonal components. In general, it is not realistic to assume that the weight vectors are orthogonal, and hence, we require tensors to learn the weight vectors. Moreover, through tensors, we can learn overcomplete networks, where the number of hidden neurons can exceed the input/output dimensions. Note that matrix factorization methods are unable to learn overcomplete models, since the rank of the matrix cannot exceed its dimensions.  Thus, it is critical to incorporate tensors for  training neural networks. A recent set of papers have analyzed the tensor methods in detail, and established convergence and perturbation guarantees~\citep{JanzaminEtal:Altmin14,JMLR:v15:anandkumar14b,anandkumar2014guaranteed,
Overcomplete:COLT2015,bhaskara2013smoothed}, despite non-convexity of the decomposition problem. Such strong theoretical guarantees are essential for deriving provable risk bounds for NN-LIFT.

\paragraph{Extensions: }Our algorithm NN-LIFT can   be extended to more layers, by  recursively estimating the weights layer by layer.  In principle, our analysis can be extended   by controlling the perturbation introduced due to layer-by-layer estimation. Establishing precise guarantees is an exciting open problem.

In this work, we assume knowledge of the generative model for the input. As argued before, in many settings such as experimental design or polling, the design of the input pdf $p(x)$ is under the control of the learner. Even if $p(x)$ is not known, a recent flurry of research activity has shown  that a wide class of probabilistic models can be trained consistently using a suite of different efficient algorithms: convex relaxation methods~\citep{chandrasekaran2010latent},  spectral and tensor methods~\citep{JMLR:v15:anandkumar14b}, alternating minimization~\citep{AnandkumarEtal:COLT14},
and they require only polynomial sample and computational complexity, with respect to the input and hidden dimensions.
These methods can learn a rich class of models which also includes   latent or hidden variable models.

Another aspect not addressed in this work is the issue of regularization for our NN-LIFT algorithm. In this work, we assume that the number of neurons is chosen appropriately to balance bias and variance through cross validation. Designing implicit regularization methods such as {\em dropout}~\citep{hinton2012improving} or {\em early stopping}~\citep{morgan1989generalization} for tensor factorization and analyzing them rigorously is another exciting open research problem.

\subsection{Related works}

We first review some works regarding the analysis of backpropagation, and then provide some theoretical results on training neural networks.
 
\paragraph{Analysis of backpropagation and loss surface of optimization:}~\citet{baldi1989neural} show that if the activations are linear, then  backpropagation has a unique local optimum, and it corresponds to the principal components of the covariance matrix of the training examples. However, it is known that there exist networks with non-linear activations where backpropagation  fails; for instance, \citet{brady1989back} construct    simple cases of linearly separable classes that backpropagation fails. Note that the simple perceptron algorithm will succeed here due to linear separability. ~\citet{gori1992problem} argue that such examples are artificial and that backpropagation succeeds in reaching the global optimum for linearly separable classes in practical settings. However, they show that under non-linear separability, backpropagation can get stuck in local optima. For a detailed survey, see~\cite{frasconi1997successes}. 

Recently,~\citet{DBLP:journals/corr/ChoromanskaHMAL14}
analyze the loss surface of a multi-layer ReLU network by relating it to a spin glass system. They make several assumptions such as variable independence for the input, equally likely paths from input to output, redundancy in network parameterization and uniform distribution for unique weights, which are far from realistic.   Under these assumptions, the network reduces to a random {\em spin glass model}, where it is known  that the lowest critical values of the random loss function form a layered structure, and the number of local minima outside that band diminishes exponentially with the network size~\citep{auffinger2013complexity}. However,  this does not imply computational efficiency: there is no guarantee that  we can find such a good local optimal point using computationally cheap algorithms, since there are still exponential number of such points.

~\citet{DBLP:journals/corr/HaeffeleV15} provide a general framework for characterizing when local optima become global in deep learning and other scenarios. The idea is that if the network is sufficiently overspecified (i.e., has enough hidden neurons) such that there exist local optima where some of the neurons have zero contribution, then such local optima are in fact, global. This provides a simple and a unified characterization of local optima which are global. However, in general, it is not clear how to design algorithms that can reach these efficient optimal points.

\paragraph{Previous theoretical works for training neural networks:}
Analysis of risk for neural networks is a classical problem. Approximation error of two layer neural network has been analyzed in a number of works~\citep{cybenko1989approximation,hornik1991approximation,barron1994approximation}. ~\citet{barron1994approximation} provides a bound on  the approximation error and combines it with the estimation error to obtain a  risk bound, but   for a computationally inefficient method. The sample complexity for neural networks have been extensively analyzed in~\cite{barron1994approximation,bartlett1998sample}, assuming convergence to the globally optimal solution, which in general is intractable. See \citet{anthony2009neural,shalev2014understanding} for an exposition of classical results on neural networks.

~\citet{andoni2014learning} learn polynomial target functions using a two-layer neural network   under Gaussian/uniform input distribution. They argue that the weights for the first layer
   can be selected randomly, and only the second layer weights, which are linear, need to be fitted optimally.  However, in practice, Gaussian/uniform distributions are never encountered in classification problems. For general distributions, random weights in the first layer is not sufficient. Under our framework, we impose only mild non-degeneracy conditions on the weights.~\citet{livni2014computational} make the observation that networks with quadratic activation functions can be trained in a computationally efficient manner in an incremental manner. This is because with quadratic activations, greedily adding one neuron at a time can be solved efficiently through eigen decomposition. However, the standard sigmoidal networks   require a large depth polynomial network, which is not practical. After we posted the initial version of this paper,~\citet{DBLP:journals/corr/ZhangLJ15} extended this framework to improper learning scenario, where the output predictor need not be a neural network. They show that if the $\ell_1$ norm of the incoming weights in each layer is bounded, then learning is efficient. However, for the usual neural networks with sigmoidal activations, the $\ell_1$ norm of the weights scales with dimension and in this case, the algorithm is no longer polynomial time.~\citet{arora2013provable} provide bounds for leaning a class of deep representations. They use layer-wise learning where the neural network is learned layer-by-layer in an unsupervised manner. They assume sparse edges with random bounded weights, and 0/1 threshold functions in hidden nodes. The difference is here, we are considering the supervised setting where there is both input and output, and we allow for general sigmoidal functions at the hidden neurons.
   
Recently, after posting the initial version of this paper,~\citet{hardt2015train}   provided an analysis of stochastic gradient descent and its generalization error in convex and non-convex problems such as training neural networks. They show that the generalization error can be controlled under mild conditions. However, their work does not address about reaching a  solution with small risk bound using SGD, and the SGD in general can get stuck in a spurious local optima. On the other hand, we show that in addition to having a small generalization error,  our method  yields a neural network with a small risk bound. Note that our method is moment-based estimation, and these methods come with stability bounds that guarantee good generalization error.

Closely related to this paper,~\citet{Sedghi:SparseNet} consider learning neural networks with sparse connectivity. They employ the cross-moment between the (multi-class) label and (first order) score function of the input. They show that they can provably learn the weights of the first layer,
 as long as the weights are sparse enough, and there are enough number of input dimensions and output classes (at least linear up to log factor in the number of neurons in any layer).  In this paper, we remove these restrictions and allow for the output to be just binary class (and indeed, our framework applies for multi-class setting as well, since the amount of information increases with more label classes from the algorithmic perspective), and for the number of neurons to exceed the input/output dimensions (overcomplete setting). Moreover, we extend beyond the realizable setting, and do not require the target functions to be generated from the class of neural networks under consideration.

\section{Preliminaries and Problem Formulation} \label{sec:model}

We first introduce some notations and then propose the problem formulation.

\subsection{Notation}

Let $[n] := \{1,2,\dotsc,n\}$, and $\|u\|$ denote the $\ell_2$ or Euclidean norm of vector $u$, and 
$\inner{u,v}$ denote the inner product of vectors $u$ and $v$.
For matrix $C \in \R^{d \times k}$, the $j$-th column is referred by $C_j$ or $c_j$, $j \in [k]$.
 Throughout this paper, $\nabla_x^{(m)}$ denotes the $m$-th order derivative operator w.r.t.\ variable $x$. 
For matrices $A, B \in \R^{d \times k}$, the {\em Khatri-Rao product} $C := A \odot B \in \R^{d^2 \times k}$ is defined such that $C(l+ (i-1)d,j) = A(i,j) \cdot B(l,j)$, for $i,l \in [d], j \in [k]$.

\paragraph{Tensor:}
A real \emph{$m$-th order tensor} $T \in \bigotimes^m \R^d$ is a member of the outer product of Euclidean spaces $\R^{d}$.
The different dimensions of the tensor are referred to as {\em modes}. For instance, for a matrix, the first mode refers to columns and the second mode refers to rows.

\paragraph{Tensor matricization:}
For a third order tensor $T \in \R^{d \times d \times d}$, the matricized version along first mode denoted by $M \in \R^{d \times d^2}$ is defined such that
\begin{equation} \label{eqn:matricization}
T(i,j,l) = M(i,l+(j-1)d), \quad i,j,l \in [d].
\end{equation}

\paragraph{Tensor rank:}A $3$rd order tensor $T \in \Rbb^{d \times d \times d}$ is said to be rank-$1$ if it can be written in the form
\begin{equation} \label{eqn:rank-1 tensor}
T= w \cdot a \otimes b\otimes c \Leftrightarrow T(i,j,l) = w \cdot a(i) \cdot b(j) \cdot c(l),
\end{equation}
where $\otimes$  represents the {\em outer product}, and $a, b , c \in \Rbb^d$ are unit vectors.
A tensor $T  \in \Rbb^{d \times d \times d}$ is said to have a CP (Candecomp/Parafac) {\em rank} $k$ if it can be (minimally) written as the sum of $k$ rank-$1$ tensors
\begin{equation}\label{eqn:tensordecomp}
T = \sum_{i\in [k]} w_i a_i \otimes b_i \otimes c_i, \quad w_i \in \Rbb, \ a_i,b_i,c_i \in \Rbb^d.
\end{equation}

\paragraph{Derivative:}
For function $g(x): \R^d \rightarrow \R$ with vector input $x \in \R^d$, the $m$-th order derivative w.r.t.\ variable $x$ is denoted by $\nabla_x^{(m)} g(x) \in \bigotimes^{m} \R^d$ (which is a $m$-th order tensor) such that
\begin{equation} \label{eqn:derivativedef}
\left[ \nabla_x^{(m)} g(x) \right]_{i_1,\dotsc,i_m} := \frac{\partial g(x)}{\partial x_{i_1} \partial x_{i_2} \dotsb \partial x_{i_m}}, \quad i_1,\dotsc,i_m \in [d].
\end{equation}
When it is clear from the context, we drop the subscript $x$ and write the derivative as $\nabla^{(m)} g(x)$.

\paragraph{Fourier transform:}For a function $f(x): \R^d \rightarrow \R$, the multivariate Fourier transform $F(\omega): \R^d \rightarrow \R$ is defined as
\begin{equation} \label{eqn:FourierDef}
F(\omega) := \int_{\R^d} f(x) e^{-j \inner{\omega,x}} dx,
\end{equation}
where variable $\omega \in \R^d$ is called the frequency variable, and $j$ denotes the imaginary unit. We also denote the Fourier pair $(f(x),F(\omega))$ as $f(x) \xleftrightarrow{\text{Fourier}} F(\omega)$.

\paragraph{Function notations:}
Throughout the paper, we use the following convention to distinguish different types of functions.
We use  $f(x)$ (or $y$) to denote an arbitrary function and exploit $\tl{f}(x)$ (or $\tl{y}$) to denote  the output of a realizable neural network. This helps us to differentiate between them. We also use notation $\hf(x)$ (or $\hy$) to denote the estimated (trained) neural networks using finite number of samples.

\subsection{Problem formulation} \label{sec:problem-formulation}

We now introduce the problem of training a  neural network in realizable and non-realizable settings, and elaborate on the notion of risk bound on how the trained neural network approximates an arbitrary function. 
It is known that continuous functions with compact domain can be arbitrarily well approximated by feedforward neural networks with one hidden layer and sigmoidal nonlinear functions~\citep{Cybenko1989,HornikEtal1989,Barron93}.

The input (feature) is denoted by variable $x$, and output (label) is denoted by variable $y$. We assume the input and output are generated according to some joint density function $p(x,y)$ such that $(x_i,y_i) \overset{\operatorname{i.i.d.}}{\sim} p(x,y)$, where $(x_i,y_i)$ denotes the $i$-th sample. We assume knowledge of the input density $p(x)$, and demonstrate how it can be used to train a neural network to approximate the conditional density $p(y|x)$ in a computationally efficient manner. We discuss in Section~\ref{sec:ScoreFunc} how the input density $p(x)$ can be estimated through numerous methods such as score matching or spectral methods. In settings such as experimental design, the input density $p(x)$ is known to the learner since she designs the density function, and our framework is directly applicable there. In addition, we   do not need to know the normalization factor or the partition function of the input distribution $p(x)$, and the estimation up to normalization factor suffices.

\paragraph{Risk bound:}
In this paper, we propose a new algorithm for training neural networks and provide risk bounds with respect to an arbitrary target function.
Risk is the expected loss over the joint probability density function of input $x$ and output $y$. Here, we consider the squared $\ell_2$ loss and bound the {\em risk error}
\begin{equation} \label{eqn:riskdef}
\Ebb_x [|f(x) - \hf(x)|^2],
\end{equation}
where $f(x)$ is an arbitrary function which we want to approximate by $\hf(x)$ denoting the estimated (trained) neural network. This notion of risk is also called mean integrated squared error.
The proposed risk error for a neural network consists of two parts: approximation error and estimation error. {\em Approximation error} is the error in fitting the target function $f(x)$ to a neural network with the given architecture $\tl{f}(x)$, and {\em estimation error} is the error in training that network with finite number of samples denoted by $\hf(x)$. Thus, the risk error measures the ability of the trained neural network to generalize to new data generated by function $f(x)$. We now introduce the realizable and non-realizable settings, which elaborates more these sources of error.

\subsubsection{Realizable setting}
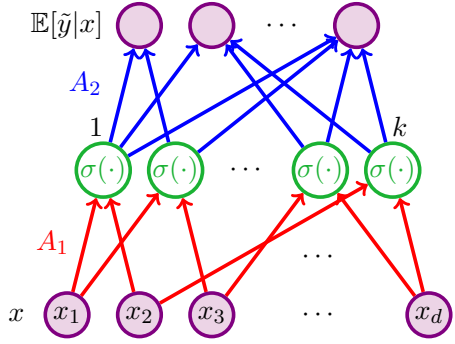
\begin{figure}[t]
\begin{center}
\resizebox{0.38\textwidth}{!}{
\begin{tikzpicture}
  [
    scale=1,
    observed/.style={circle,minimum size={width("$x_{d}$")+6pt},inner
sep=0mm,draw=violet,fill=lpurple,line width=.5mm},
    hidden/.style={circle,minimum size=0.6cm,inner sep=0mm,draw=dkg,line width=.5mm},
        func/.style={circle,minimum size=0.6cm,inner sep=0mm,draw=blue1,dashed, line width=.5mm},
        vdots/.style={min, node distance=.5mm},
  ]
  \node [hidden,name=f1] at ($(-2,0)$) {$\tcdkg{\sigma(\cdot)}$};
  \node [hidden,name=f2] at ($(-1,0)$) {$\tcdkg{\sigma(\cdot)}$};
  \node [hidden,name=fk] at ($(2,0)$) {$\tcdkg{\sigma(\cdot)}$};
 \node [hidden,name=fn] at ($(1,0)$) {$\tcdkg{\sigma(\cdot)}$};
  \node [] at ($(-2.1,0.6)$) {$1$};
  \node [] at ($(2.1,0.6)$) {$k$};
  \node[observed,name=y1] at ($(-1.5,2)$){};
  \node [observed,name=yk] at ($(-.5,2)$){};
  \node [observed,name=y2] at ($(1.5,2)$){};
 \node [observed,name=x11] at ($(-2.5,-2)$) {$x_1$}; 
   \node [observed,name=x1] at ($(-1.5,-2)$) {$x_2$}; 
  \node [observed,name=x2] at ($(-0.5,-2)$) {$x_3$}; 
  \node [observed,name=xd2] at ($(2.5,-2)$) {$x_{d}$}; 
  \node [] at ($(-3.2,-2)$) {$x$};
  \node [] at ($(-2.5,2)$) {$\E[\tl{y}|x]$};
  \node [] at ($(-2.25,1.2)$) {\tcb{$A_2$}};
   \node [] at ($(-2.7,-1)$) {\tcr{$A_1$}};
    \node at ($(0.5,2)$) {$\dotsb$};
        \node at ($(1,-1.2)$) {$\dotsb$};
  \node at ($(0,0)$) {$\dotsb$};
   \node at ($(1,-2)$) {$\dotsb$};

  \draw [blue, line width=.5mm, ->] (f1) to (y1);
  \draw [blue, line width=.5mm, ->] (f1) to (y2);
  \draw [blue, line width=.5mm, ->] (f1) to (yk);
 \draw [blue, line width=.5mm,->] (f2) to (y2);
  \draw [blue, line width=.5mm, ->] (f2) to (y1);
  \draw [blue, line width=.5mm, ->] (fk) to (yk);
  \draw [blue, line width=.5mm, ->] (fk) to (y2);
   \draw [blue, line width=.5mm, ->] (fn) to (y2);
   \draw [blue, line width=.5mm, ->] (fn) to (yk);

    \draw [red, line width=.5mm, <-] (f1) to (x1);
  \draw [red, line width=.5mm, <-] (f1) to (x11);
   \draw [red, line width=.5mm, <-] (f2) to (x11);
  \draw [red, line width=.5mm,<-] (f2) to (x2);
  \draw [red, line width=.5mm,<-] (fk) to (xd2);
  \draw [red, line width=.5mm, <-] (fk) to (x1);
  \draw [red, line width=.5mm, <-] (fn) to (x2);
  \draw [red, line width=.5mm,<-] (fn) to (xd2);

\end{tikzpicture}
}
\end{center}
\caption{\small Graphical representation of a neural network, $\Ebb[\tl{y}|x] = A_2^\top \sigma(A_1^\top x+b_1)+b_2$. Note that this representation is for general vector output $\tl{y}$ which can be also written as $\Ebb[\tl{y}|x]= \inner{a_2,\sigma(A_1^\top x+b_1)}+b_2$ in the case of scalar output $\tl{y}$.}
\label{fig:DeepNet}
\end{figure}

In the realizable setting, the output is generated by a neural network. We consider a  neural network with one hidden layer of dimension $k$. Let the output $\tl{y} \in \{0,1\}$ be the binary label, and $x \in \R^d$ be the feature vector; see Section~\ref{sec:discussion} for generalization to higher dimensional output (multi-label and multi-class), and also the continuous output case. 
We consider the label generating model
\begin{equation} \label{eq:nn2}
\quad \tl{f}(x) := \Ebb[\tl{y}|x]= \inner{a_2,\sigma(A_1^\top x+b_1)}+b_2,
\end{equation}
where $\sigma(\cdot)$ is (linear/nonlinear) elementwise function. 
See Figure~\ref{fig:DeepNet} for a schematic representation of label-function in~\eqref{eq:nn2} in the general case of vector output $\tl{y}$. 

In the realizable setting, the goal is to train the neural network in~\eqref{eq:nn2}, i.e., to learn the weight matrices (vectors) $A_1 \in \R^{d \times k}$, $a_2 \in \R^k$ and bias vectors $b_1 \in \R^k$, $b_2 \in \R$. This only involves the estimation analysis where we have a label-function $\tl{f}(x)$ specified in~\eqref{eq:nn2} with fixed unknown parameters $A_1, b_1, a_2, b_2$, and the goal is to learn these parameters and finally bound the overall function estimation error $\E_x[|\tl{f}(x) - \hf(x)|^2]$, where $\hf(x)$ is the estimation of fixed neural network $\tl{f}(x)$ given finite samples.
For this task, we propose a computationally efficient algorithm which requires only polynomial number of samples for  bounded estimation error. This is the first time that such a result has been established for any neural network.


\subsubsection{Non-realizable setting}
In the non-realizable setting, the output is an arbitrary function $f(x)$ which is not necessarily a neural network. We want to approximate $f(x)$ by $\hf(x)$ denoting the estimated (trained) neural network. In this setting, the additional approximation analysis is also required. In this paper, we combine the estimation result in realizable setting with the approximation bounds in~\citet{Barron93} leading to risk bounds with respect to the target function $f(x)$; see~\eqref{eqn:riskdef} for the definition of risk.  The detailed results are provided in Section~\ref{sec:Generalization}. 


\section{NN-LIFT Algorithm}


In this section, we introduce our proposed method for learning neural networks using tensor, Fourier and regression techniques. 
Our method is shown in Algorithm~\ref{algo:NN-LIFT} named NN-LIFT (Neural Network LearnIng using Feature Tensors). The algorithm has three main components. The first component involves estimating the weight matrix of the first layer denoted by $A_1 \in \R^{d \times k}$ by a tensor decomposition method. The second component involves estimating the bias vector of the first layer $b_1 \in \R^k$ by a Fourier method. We finally estimate the parameters of last layer $a_2 \in \R^k$ and $b_2 \in \R$ by linear regression.

Note that most of the unknown parameters (compare the dimensions of matrix $A_1$, vectors $a_2$, $b_1$, and scalar $b_2$) are estimated in the first part, and thus, this is the main part of the algorithm.
Given this fact, we also provide an alternative method for the  estimation of other parameters of the model, given the estimate of $A_1$ from the tensor method. This is based on incrementally adding neurons, one by one, whose first layer weights are given by $A_1$ and the remaining parameters are updated using brute force search on a grid. Since each update involves just updating the corresponding bias term $b_1$, and its contribution to the final output, this is low dimensional, and can be done efficiently; details are in Section~\ref{sec:ft-alt}.

We now explain the steps of Algorithm~\ref{algo:NN-LIFT} in more details.
%

\begin{algorithm}[t]
\caption{NN-LIFT (Neural Network LearnIng using Feature Tensors)}
\label{algo:NN-LIFT}
\begin{algorithmic}[1]
\renewcommand{\algorithmicrequire}{\textbf{input}}
\renewcommand{\algorithmicensure}{\textbf{output}}
\REQUIRE Labeled samples $\{(x_i,y_i): i \in [n]\}$, parameter $\tl{\epsilon}_1$, parameter $\lambda$.
\REQUIRE Third order score function $\Pc_3(x)$ of the input $x$; see Equation~\eqref{eqn:diffoperator} for the definition.

\STATE Compute $\widehat{T}:= \frac{1}{n} \sum_{i \in [n]} y_i \cdot \Pc_3(x_i)$. 
\STATE $\lbrace (\hat{A}_1)_j  \rbrace_{j \in [k]}=\text{tensor decomposition}(\widehat{T})$; see Section~\ref{sec:tensordecomp} and Appendix~\ref{appendix:tensordecomp} for details. \label{line:TensorDecomposition}

\STATE  $\hb_1 = \text{Fourier method}(\{(x_i,y_i): i \in [n]\},\hA_1,\tl{\epsilon}_1)$; see Procedure~\ref{algo:Fourier}.

\STATE $(\ha_2, \hb_2) = \text{Ridge regression}(\{(x_i,y_i): i \in [n]\},\hA_1,\hb_1,\lambda)$; see Procedure~\ref{algo:LLSQ}.

\RETURN $\hat{A}_1, \ha_2, \hb_1, \hb_2$.

\end{algorithmic}
\end{algorithm}

\subsection{Score function} \label{sec:ScoreFunc}
The $m$-th order score function $\Pc_m(x) \in \bigotimes^m \R^d$ is defined as~\citep{janzamin2014matrix}
\begin{equation} \label{eqn:diffoperator}
\Pc_m(x) := (-1)^m \frac{\nabla_x^{(m)} p(x)}{p(x)},
\end{equation}
where $p(x)$ is the probability density function of random vector $x \in \R^d$. In addition, $\nabla_x^{(m)}$ denotes the $m$-th order derivative operator; see~\eqref{eqn:derivativedef} for the precise definition. The main property of score functions as yielding differential operators that enables us to estimate the weight matrix $A_1$ via tensor decomposition is discussed in the next subsection; see Equation~\eqref{eqn:cross-moment-repeat}.

In this paper, we assume   access to a sufficiently good approximation of the input pdf $p(x)$ and the corresponding  score functions $\Sc_2(x)$, $\Sc_3(x)$. Indeed, estimating these quantities in general is a hard problem, but there exist numerous instances where this becomes tractable. Examples include spectral methods for learning latent variable models such as Gaussian mixtures, topic or admixture models, independent component analysis (ICA) and so on~\citep{JMLR:v15:anandkumar14b}. Moreover, there have been recent  advances in non-parametric score matching methods~\citep{sriperumbudur2013density} for density estimation in infinite dimensional exponential families with guaranteed convergence rates. These methods can be used to estimate the input pdf in an unsupervised manner. Below, we discuss in detail about score function estimation methods. In this paper, we focus on how we can use the input generative information to make   training of  neural networks tractable.  For simplicity, in the subsequent analysis, we assume that these quantities are perfectly known; it is possible to extend the perturbation analysis to take into account the errors in estimating the input pdf; see Remark~\ref{remark:score-error}.

\paragraph{Estimation of score function:}There are various efficient methods for estimating the score function. The framework of score matching is popular for parameter estimation  in probabilistic models~\citep{hyvarinen2005estimation, swersky2011autoencoders}, where the criterion is to fit parameters based on matching the data score function. \citet{swersky2011autoencoders} analyze the score matching for latent energy-based models.
In deep learning, the framework of auto-encoders attempts to find encoding and decoding functions which minimize the reconstruction error under added noise; the so-called Denoising Auto-Encoders (DAE). This is an unsupervised framework involving only unlabeled samples. \citet{alain2012regularized} argue that the DAE   approximately learns the first order score function of the input, as the noise variance goes to zero. ~\citet{sriperumbudur2013density} propose non-parametric score matching methods for density estimation in infinite dimensional exponential families with guaranteed convergence rates. Therefore, we can use any of these methods for estimating $\Pc_1(x)$ and use the recursive form~\citep{janzamin2014matrix}
%
$$\Pc_m(x) = - \Pc_{m-1}(x) \otimes \nabla_x \log p(x) - \nabla_x \Pc_{m-1}(x)$$ to estimate higher order score functions.

\subsection{Tensor decomposition} \label{sec:tensordecomp}
The score functions are new representations (extracted features) of input data $x$ that can be used for training neural networks.
We   use score functions and labels of training data to  form the empirical cross-moment $\widehat{T}=\frac{1}{n} \sum_{i \in [n]} y_i \cdot \Pc_3(x_i)$.   We decompose tensor $\widehat{T}$ to estimate the columns of $A_1$. The following discussion reveals why tensor decomposition is relevant for this task.

The score functions have  the property of yielding  differential operators with respect to the input distribution. 
More precisely, for label-function $f(x) := \E[y|x]$, \citet{janzamin2014matrix} show that
$$\E[y \cdot \Sc_3(x)] = \E[\nabla_x^{(3)} f(x)].$$
Now for the neural network output in~\eqref{eq:nn2}, note that the function $\tl{f}(x)$ is a non-linear function of both input $x$ and weight matrix $A_1$. The expectation operator $\Ebb[\cdot]$ averages out the dependency on $x$, and the derivative acts as a {\em linearization operator} as follows. In the neural network output~\eqref{eq:nn2}, we observe that  the columns of weight vector $A_1$ are the linear coefficients involved with input variable $x$. When taking the derivative of this function, by the chain rule, these linear coefficients shows up in the final form.
In particular, we show in Lemma~\ref{lem:moment} (see Section~\ref{sec:proofsketch}) that for neural network in~\eqref{eq:nn2}, we have
\begin{equation} \label{eqn:cross-moment-repeat}
\Ebb \left[ \tl{y} \cdot \Pc_3(x) \right] = \sum_{j \in [k]} \lambda_j \cdot (A_1)_j \otimes (A_1)_j \otimes (A_1)_j \in \R^{d \times d \times d},
\end{equation}
where $(A_1)_j \in \R^d$ denotes the $j$-th column of $A_1$, and $\lambda_j \in \R$ denotes the coefficient; refer to Equation~\eqref{eqn:tensordecomp} for the notion of tensor rank and its rank-1 components. This clarifies how the score function acts as a linearization operator while the final output is nonlinear in terms of $A_1$. The above form also clarifies the reason behind using tensor decomposition in the learning framework.

\paragraph{Tensor decomposition algorithm:}The goal of tensor decomposition algorithm is to recover the rank-1 components of tensor. 
For this step, we use the tensor decomposition algorithm proposed in~\citep{JMLR:v15:anandkumar14b,anandkumar2014guaranteed}; see Appendix~\ref{appendix:tensordecomp} for details.
The main step of the tensor decomposition method is the {\em tensor power iteration} which is the generalization of matrix power iteration to $3$rd order tensors. The tensor power iteration is given by
\begin{equation*} 
u \leftarrow \frac{T(I,u,u)}{\|T(I,u,u)\|},
\end{equation*}
where $u \in \R^d$, $T(I,u,u) :=  \sum_{j,l \in [d]} u_j u_l T(:,j,l) \in \R^d$ is a {\em multilinear} combination of tensor {\em fibers}.\footnote{Tensor fibers are the vectors which are derived by fixing all the indices of the tensor except one of them, e.g., $T(:,j,l)$ in the above expansion.}
The convergence guarantees of tensor power iteration for orthogonal tensor decomposition have been developed in the literature~\citep{ZG01,JMLR:v15:anandkumar14b}. Note that we first orthogonalize the tensor via whitening procedure and then apply the tensor power iteration. Thus, the original tensor decomposition need not to be orthogonal.

\paragraph{Computational Complexity:} It is popular to perform the tensor decomposition in  a stochastic manner which reduces the computational complexity. This is done by splitting the data into mini-batches. Starting with the first mini-batch, we perform a small number of tensor power iterations, and then use the result as initialization for the next mini-batch, and so on. As mentioned earlier, we assume that a sufficiently good approximation of score function tensor is given to us. For specific cases where we have this tensor in factor form, we can reduce the computational complexity of NN-LIFT by not computing the whole tensor explicitly. By having factor form, we mean when we can write the score function tensor in terms of summation of rank-1 components which could be the summation over samples, or from other existing structures in the model. We now state a few examples when we have the factor form, and provide the computational complexity.
For example, if input follows Gaussian distribution, the score function has a simple polynomial form, and  the computational complexity of tensor decomposition is $O(nkdR)$, where $n$ is the number of samples and $R$ is the number of initializations for the tensor decomposition. Similar argument follows when the input distribution is mixture of Gaussian distributions.

We can also analyze complexity for more complex inputs. If we fit the input data into a Restricted Boltzmann Machines (RBM) model, the computational complexity of our method is $O(nkd d_h R)$.  Here $d_h$ is the number of neurons of the first layer of the RBM used for approximating the input distribution. Tensor methods are also embarrassingly parallelizable. When performed in parallel, the computational complexity would be $O(\log (\min \{d, d_h\}))$ with $O(nkd d_h R/\log(\min(d,d_h)))$ processors. Alternatively, we can also exploit recent tensor sketching approaches~\citep{wang2015fast}  for computing tensor decompositions efficiently.~\citet{wang2015fast}  build on the idea of count sketches and show that the running time is linear in the input dimension and the number of samples, and is independent in the order of the tensor.  Thus, tensor decompositions can be computed efficiently.


\subsection{Fourier method} \label{sec:Fourier}
The second part of the algorithm estimates the first layer bias vector $b_1 \in \R^k$. This step is very different from the previous step for estimating $A_1$ which was based on tensor decomposition methods. This is a Fourier-based method where  complex variables are formed using labeled data and random frequencies in the Fourier domain; see Procedure~\ref{algo:Fourier}.  We prove in Lemma~\ref{lem:v-mean} that the entries of $b_1$ can be estimated from the phase of these complex numbers. We also observe in Lemma~\ref{lem:v-mean} that the magnitude of these complex numbers can be used to estimate $a_2$; this is discussed in Appendix~\ref{appendix:Fourier}.

\paragraph{Polynomial-time random draw from set $\Omega_l$:}
Note that the random frequencies are drawn from a $d-1$ dimensional manifold denoted by $\Omega_l$ which is the intersection of sphere $\|\omega\|=\frac{1}{2}$ and cone $\bigl|  \inner{\omega,(\hA_1)_l} \bigr| \geq \frac{1-\tl{\epsilon}_1^2/2}{2}$ in $\R^d$. This manifold is actually the surface of a spherical cap. In order to draw these frequencies in polynomial time, we consider the $d$-dimensional spherical coordinate system such that one of the angles is defined based on the cone axis. We can then directly impose the cone constraint by limiting the corresponding angle in the random draw.
In addition, \citet{kothariMeka2014} propose a method for generating pseudo-random variables from the spherical cap in logarithmic time.

\floatname{algorithm}{Procedure}
\begin{algorithm}[t]
\caption{Fourier method for estimating $b_1$}
\label{algo:Fourier}
\begin{algorithmic}[1]
\renewcommand{\algorithmicrequire}{\textbf{input}}
\renewcommand{\algorithmicensure}{\textbf{output}}
\REQUIRE Labeled samples $\{(x_i,y_i): i \in [n]\}$, estimate $\hA_1$, parameter $\tl{\epsilon}_1$.
\REQUIRE Probability density function $p(x)$ of the input $x$.
\FOR {$l = 1$ to $k$}
\STATE Let $\Omega_l := \left\{ \omega \in \R^d : \|\omega\|=\frac{1}{2},  \bigl|  \inner{\omega,(\hA_1)_l} \bigr| \geq \frac{1-\tl{\epsilon}_1^2/2}{2} \right\}$, and $|\Omega_l|$ denotes the surface area of $d-1$ dimensional manifold $\Omega_l$.
\STATE Draw $n$ i.i.d.\ random frequencies $\omega_i, i \in [n],$ uniformly from set $\Omega_l$.
\STATE Let $v := \frac{1}{n} \sum_{i \in [n]} \frac{y_i}{p(x_i)} e^{-j \inner{\omega_i, x_i}}$ which is a complex number as $v = |v| e^{j \angle v}$. The operators $|\cdot|$ and $\angle \cdot$ respectively denote the magnitude and phase operators. \label{line:v}
\STATE Let $\hb_1(l) := \frac{1}{ \pi} (\angle v - \angle \Sigma(1/2))$, where $\sigma(x) \xleftrightarrow{\text{Fourier}} \Sigma(\omega)$.
\ENDFOR
\RETURN $\hb_1$.
\end{algorithmic}
\end{algorithm}

\begin{remark}[Knowledge of input distribution only up to normalization factor]
The computation of score function and the Fourier method  both involve knowledge about input pdf $p(x)$. However,  we do not need to know the normalization factor, also known as partition function, of the input pdf. For the score function, it is immediately seen from the definition in~\eqref{eqn:diffoperator} since the normalization factor is canceled out by the division by $p(x)$, and thus, the estimation of score function is at most as hard as estimation of input pdf up to normalization factor. In the Fourier method, we can use the non-normalized estimation of input pdf which leads to a normalization mismatch in the estimation of corresponding  complex number. 
This is not a problem since we only use the phase information of these complex numbers.
\end{remark}


\subsection{Ridge regression method}
For the neural network model in~\eqref{eq:nn2}, given a good estimation of neurons, we can estimate the parameters of last layer by linear regression. We provide Procedure~\ref{algo:LLSQ} in which we use ridge regression algorithm to estimate the parameters of last layer $a_2$ and $b_2$. See Appendix~\ref{appendix:LLSQ} for the details of ridge regression and the corresponding analysis and guarantees.

\floatname{algorithm}{Procedure}
\begin{algorithm}[t]
\caption{Ridge regression method for estimating $a_2$ and $b_2$}
\label{algo:LLSQ}
\begin{algorithmic}[1]
\renewcommand{\algorithmicrequire}{\textbf{input}}
\renewcommand{\algorithmicensure}{\textbf{output}}
\REQUIRE Labeled samples $\{(x_i,y_i): i \in [n]\}$, estimates $\hA_1$ and $\hb_1$, regularization parameter $\lambda$. 
\STATE Let $\hh_i := \sigma ( \hA_1^\top x_i + \hb_1 )$, $i \in [n]$, denote the estimation of the neurons.
\STATE Append each neuron $\hh_i$ by the dummy variable 1 to represent the bias, and thus, $\hh_i \in \R^{k+1}$.
\STATE Let $\hat{\Sigma}_{\hh} := \frac{1}{n} \sum_{i \in [n]}  \hh_i \hh_i^\top \in \R^{(k+1) \times (k+1)}$ denote the empirical covariance of $\hh$.
\STATE Let $\hat{\beta}_\lambda \in \R^{k+1}$ denote the estimated parameters 
by $\lambda$-regularized ridge regression as
\begin{equation} \label{eqn:a2-est}
\hat{\beta}_\lambda = \left( \hat{\Sigma}_{\hh} + \lambda I_{k+1} \right)^{-1} \cdot
\frac{1}{n} \left( \sum_{i \in [n]} y_i \hh_i \right),
\end{equation}
where $I_{k+1}$ denotes the $(k+1)$-dimensional identity matrix.
\RETURN $\ha_2 := \hat{\beta}_\lambda(1:k)$, $\hb_2 := \hat{\beta}_\lambda(k+1)$.
\end{algorithmic}
\end{algorithm}


\section{Risk Bound in the Realizable Setting}

In this section, we provide  guarantees in the realizable setting, where the function $\tl{f}(x):=\Ebb[\tl{y}|x]$ is generated by a neural network as  in~\eqref{eq:nn2}. We   provide the estimation error bound on the overall function recovery $\E_x [ |\tl{f}(x)-\hf(x)|^2 ]$ when the estimation is done by Algorithm~\ref{algo:NN-LIFT}.

We provide  guarantees in the following settings. 1) In the basic case, we consider the {\em undercomplete} regime $k \leq d$, and provide the results assuming $A_1$ is full column rank. 2) In the second case, we form higher order cross-moments and tensorize it into a lower order tensor. This enables us to learn the network in the overcomplete regime $k>d$, when the Khatri-Rao product $A_1 \odot A_1 \in \R^{d^2 \times k}$ is full column rank. We call this the {\em overcomplete} setting and this can handle up to $k =O(d^2)$. Similarly, we can extend to larger $k$ by tensorizing higher order moments in the expense of additional computational complexity.



We define the following quantity for label function $\tl{f}(\cdot)$ as
$$\tl{\zeta}_{\tl{f}} := \int_{\R^d} \tl{f}(x) dx.$$
Note that in the binary classification setting ($\tl{y} \in \{0,1\}$), we have $\E[\tl{y}|x] := \tl{f}(x) \in [0,1]$ which is always positive, and there is no square of $\tl{f}(x)$ considered in the above quantity.

Let $\eta$ denote the noise in the neural network model in~\eqref{eq:nn2} such that the output is $$\tl{y} = \tl{f}(x) + \eta.$$
Note that the noise $\eta$ is not necessarily independent of $x$; for instance, in the classification setting or binary output $\tl{y} \in \{0,1\}$, the noise in dependent on $x$.




\paragraph{Conditions for Theorem~\ref{thm:guarantees-sample}:}
\vspace{-0.05in}
\bi[itemsep=-0.1mm]
\item {\bf Non-degeneracy of weight vectors}: In the undercomplete setting  $(k \leq d)$, the weight matrix $A_1 \in \R^{d \times k}$ is full column rank and $s_{\min}(A_1)> \epsilon$, where $s_{\min}(\cdot)$ denotes the minimum singular value, and $\epsilon>0$ is related to the target error in recovering the columns of $A_1$. In the overcomplete setting $(k \leq d^2)$, the Khatri-Rao product $A_1 \odot A_1 \in \R^{d^2 \times k}$ is full column rank\footnote{It is shown in~\citet{bhaskara2013smoothed} that this condition is satisfied under smoothed analysis.}, and $s_{\min}(A_1\odot A_1)> \epsilon$; see Remark~\ref{remark:TensorizingGeneralization} for generalization.
\item {\bf Conditions on nonlinear activating function $\sigma(\cdot)$}: the coefficients
\begin{align*}
\lambda_j  := \E \left[ \sigma'''(z_j) \right] \cdot a_2(j), \quad
\tl{\lambda}_j  := \E \left[ \sigma''(z_j) \right] \cdot a_2(j), \quad  j \in [k],
\end{align*}
in~\eqref{eqn:cross-moment-coeffs} and \eqref{eqn:cross-moment2-coeffs} are nonzero. Here, $z := A_1^\top x+b_1$ is the input to the nonlinear operator $\sigma(\cdot)$.
In addition, $\sigma(\cdot)$ satisfies the {\em Lipschitz} property\footnote{ If the step function $\sigma(u) = 1_{\{u>0\}}(u)$ is used as the activating function, the Lipschitz property does not hold because of the non-continuity at $u=0$. But, we can assume the Lipschitz property holds in the linear continuous part, i.e., when $u,u'>0$ or $u,u'<0$. We then argue that the input to the step function $1_{\{u>0\}}(u)$ is w.h.p.\ in the linear interval (where the Lipschitz property holds). 
} with constant $L$ such that $|\sigma(u) - \sigma(u')| \leq L \cdot |u-u'|$, for $u,u' \in \R$.
Suppose that the   nonlinear activating function $\sigma(z)$ satisfies the property such that $\sigma(z) = 1 - \sigma(-z)$. Many popular activating functions such as step function, sigmoid function and tanh function satisfy this last property.
\item {\bf Subgaussian noise}: There exists a finite $\sigma_{\text{noise}} \geq 0$ such that, almost surely,
$$
\E_{\eta}[\exp(\alpha \eta)| x] \leq \exp(\alpha^2 \sigma_{\text{noise}}^2/2), \quad \forall \alpha \in \R,
$$
where $\eta$ denotes the noise in the output $\tl{y}$.
\item {\bf Bounded statistical leverage}: There exists a finite $\rho_\lambda \geq 1$ such that, almost surely,
$$\frac{\sqrt{k}}{\sqrt{(\inf\{\lambda_j\} + \lambda)k_{1,\lambda}}} \leq \rho_\lambda,$$
where $k_{1,\lambda}$ denotes the effective dimensions of the hidden layer $h := \sigma ( A_1^\top x + b_1)$ as
$
k_{1,\lambda} := \sum_{j \in [k]} \frac{\lambda_j}{\lambda_j + \lambda}.
$
Here, $\lambda_j$'s denote the (positive) eigenvalues of the hidden layer covariance matrix $\Sigma_h$, and $\lambda$ is the regularization parameter of ridge regression.
\ei

We now elaborate on these conditions. The {\em non-degeneracy of weight vectors} are required for the tensor decomposition analysis in the estimation of $A_1$.
In the undercomplete setting, the algorithm first orthogonalizes (through whitening procedure) the tensor given in~\eqref{eqn:cross-moment-repeat}, and then decomposes it through tensor power iteration. Note that the convergence of power iteration for orthogonal tensor decomposition is guaranteed~\citep{ZG01,JMLR:v15:anandkumar14b}. For the orthogonalization procedure to work, we need the tensor components (the columns of matrix $A_1$) to be linearly independent. In the overcomplete setting, the algorithm performs the same steps with the additional tensorizing procedure; see Appendix~\ref{appendix:tensordecomp} for details. In this case, a higher order tensor is given to the algorithm and it is first tensorized before performing the same steps as in the undercomplete setting. Thus, the same conditions are now imposed on $A_1 \odot A_1$. 

In addition to the non-degeneracy condition on weight matrix $A_1$, the {\em coefficients condition} on $\lambda_j$'s is also required to ensure the corresponding rank-1 components in~\eqref{eqn:cross-moment-repeat} do not vanish, and thus, the tensor decomposition algorithm recovers them.
Similarly, the coefficients $\tl{\lambda}_j$ should be also nonzero to enable us using the second order moment $\tl{M}_2$ in~\eqref{eqn:second-moment-score} in the whitening step of tensor decomposition algorithm.  If one of the coefficients vanishes, we use the other option to perform the whitening; see Remark~\ref{remark:whitening} and Procedure~\ref{algo:whitening} for details.
Note that the amount of non-linearity of $\sigma(\cdot)$ affects the magnitude of the coefficients. 
It is also worth mentioning that although we use the third derivative notation $\sigma'''(\cdot)$ in characterizing the coefficients $\lambda_j$ (and similarly $\sigma''(\cdot)$ in $\tl{\lambda}_j$), we do not need the differentiability of non-linear function $\sigma(\cdot)$ in all points. In particular, when input $x$ is a continuous variable, we use Dirac delta function $\delta(\cdot)$ as the derivative in non-continuous points; for instance, for the derivative of step function $1_{\{x>0\}}(x)$, we have $\frac{d}{dx} 1_{\{x>0\}}(x) = \delta(x)$. Thus, in general, we only need the expectations $\E \left[ \sigma'''(z_j) \right]$ and $\E \left[ \sigma''(z_j) \right]$ to exist for these type of functions and the corresponding higher order derivatives.

We impose the {\em Lipschitz} condition on the non-linear activating function to limit the error propagated in the hidden layer, when the first layer parameters are estimated by the neural network and Fourier methods. The condition $\sigma(z) = 1 - \sigma(-z)$ is also assumed to tackle the sign issue in recovering the columns of $A_1$; see Remark~\ref{remark:sign} for the details. The {\em subgaussian noise} and the {\em bounded statistical leverage} conditions are standard conditions, required for ridge regression, which is used for estimating the parameters of the second layer of the neural network. Both parameters $\sigma_{\text{noise}}$, and $\rho_\lambda$ affect the sample complexity in the final guarantees.

Imposing additional bounds on the parameters of the neural network are useful in learning these parameters with computationally efficient algorithms since it limits the searching space for training these parameters.
In particular, for the Fourier analysis, we assume the following conditions. Suppose the columns of weight matrix $A_1$ are normalized, i.e., $\|(A_1)_j\|=1, j \in [k]$, and the entries of first layer bias vector $b_1$ are bounded as $|b_1(l)| \leq 1, l \in [k]$.
Note that the normalization condition  on the columns of $A_1$ is also needed for identifiability of the parameters. For instance, if the non-linear operator is the step function $\sigma(z) = 1_{\{z>0\}}(z)$, then matrix $A_1$ is only identifiable up to its norm, and thus, such normalization condition is required for identifiability.
The estimation of entries of the bias vector $b_1$ is obtained from the phase of a complex number through Fourier analysis; see Procedure~\ref{algo:Fourier} for details. Since there is ambiguity in the phase of a complex number\footnote{A complex number does not change if  an integer multiple of $2 \pi$ is added to its phase.}, we impose the bounded assumption on the entries of $b_1$ to avoid this ambiguity. 

Let $p(x)$ satisfy some mild regularity conditions on the boundaries of the support of $p(x)$. In particular, all the entries of (matrix-output) functions
\begin{equation} \label{eqn:regularity-cond}
\tl{f}(x) \cdot \nabla^{(2)} p(x), \quad
\nabla \tl{f}(x) \cdot \nabla p(x)^\top, \quad
\nabla^{(2)} \tl{f}(x) \cdot  p(x)
\end{equation}
should go to zero on the boundaries of support of $p(x)$.
These regularity conditions are required for the properties of the score function to hold; see~\citet{janzamin2014matrix} for more details.

In addition to the above main conditions, we also need some mild conditions which are not crucial for the recovery guarantees and are mostly assumed to simplify the presentation of the main results. These conditions can be relaxed more. Suppose the input $x$ is bounded, i.e.,  $x \in B_r$, where  $B_r := \{x: \|x\| \leq r\}$.
Assume  the input probability density function $p(x) \geq \psi$ for some $\psi >0$, and for any $x \in B_r$.
The regularity conditions in~\eqref{eqn:regularity-cond} might seem contradictory with the lower bound condition $p(x) \geq \psi$, but there is an easy fix for that.
The lower bound on $p(x)$ is required for the analysis of the Fourier part of the algorithm. We can have a continuous $p(x)$, while in the Fourier part, we only use $x$'s such that $p(x) \geq \psi$, and ignore the rest. This only introduces a probability factor $\Pr[x: p(x) \geq \psi]$ in the analysis.

\paragraph{Settings of algorithm in Theorem~\ref{thm:guarantees-sample}:}
\bi[itemsep=-1mm]
\item No.\ of iterations in Algorithm~\ref{alg:robustpower}: $N = \Theta \left( \log \frac{1}{\epsilon} \right)$.
\item No.\ of initializations in Procedure~\ref{algo:SVD init}: $R  \geq \poly(k)$.
\item Parameter $\tl{\epsilon}_1 = \tl{O} \left(\frac{1}{\sqrt{n}}\right)$ in Procedure~\ref{algo:Fourier}, where $n$ is the number of samples.
\item  We exploit the empirical second order moment $\widehat{M}_2 := \frac{1}{n} \sum_{i \in [n]} y_i \cdot \Pc_2(x_i)$, in the whitening Procedure~\ref{algo:whitening}, which is the first option stated in the procedure. See Remark~\ref{remark:whitening} for further discussion about the other option.
\ei

\begin{theorem}[NN-LIFT guarantees: estimation bound in the realizable setting] \label{thm:guarantees-sample}
Assume the above settings and conditions hold.
For $\epsilon > 0$, suppose the number of samples $n$ satisfies
(up to $\log$ factors)
\begin{align} \label{eqn:samplecomp}
n \geq
\tl{O} \left( \frac{k}{\epsilon^2} \right. & \cdot \E \left[ \left\| M_3(x) M_3^\top(x) \right\| \right] \\
& \hspace{-2em} \left.
\cdot \poly \left(
\tl{y}_{\max},
\frac{\E \left[ \left\| \Sc_2(x) \Sc_2^\top(x) \right\| \right]}{\E \left[ \left\| M_3(x) M_3^\top(x) \right\| \right]},
 \frac{\tl{\zeta}_{\tl{f}}}{\psi},
\frac{\tl{\lambda}_{\max}}{\tl{\lambda}_{\min}},
\frac{1}{\lambda_{\min}},
\frac{s_{\max}(A_1)}{s_{\min}(A_1)},
|\Omega_l|, L, \frac{\|a_2\|}{(a_2)_{\min}}, |b_2|, \sigma_{\text{noise}}, \rho_\lambda \right)
\right). \nn
\end{align}
See~\eqref{eqn:samplecomp-tensordecomp}, \eqref{eqn:samplecomp-fourier}, \eqref{eqn:samplecomp-ridge1} and \eqref{eqn:samplecomp-ridge2} for the complete form of sample complexity. 
Here, $M_3(x) \in \R^{d \times d^2}$ denotes the matricization of score function tensor $\Sc_3(x) \in \R^{d \times d \times d}$; see~\eqref{eqn:matricization} for the definition of matricization. Furthermore, $\lambda_{\min} := \min_{j \in [k]} |\lambda_j|$, $\tl{\lambda}_{\min} := \min_{j \in [k]} |\tl{\lambda}_j|$, $\tl{\lambda}_{\max} := \max_{j \in [k]} |\tl{\lambda}_j|$, $(a_2)_{\min} := \min_{j \in [k]} |a_2(j)|$, and $\tl{y}_{\max}$ is such that  $|\tl{f}(x)| \leq \tl{y}_{\max}$, for $x \in B_r$.
Then the  function estimate $\hf(x) := \inner{\ha_2,\sigma(\hA_1^\top x+\hb_1)}+\hb_2$ using the estimated parameters 
$\hA_1,\hb_1,\ha_2,\hb_2$ (output of NN-LIFT Algorithm~\ref{algo:NN-LIFT}) satisfies  
the estimation error
$$\E_x[|\hf(x) - \tl{f}(x)|^2] \leq \tl{O} (\epsilon^2).$$
\end{theorem}

%
%
%

See Section~\ref{sec:proofsketch} and Appendix~\ref{appendix:mainproof} for the proof of theorem. Thus, we estimate the neural network in polynomial time and sample complexity. This is one of the first results to provide a guaranteed method for training neural networks with efficient computational and statistical complexity. Note that although the sample complexity in~\citep{Barron93} is smaller as $n \geq \tl{O} \left( \frac{kd}{\epsilon^2} \right)$,   the proposed algorithm in~\citep{Barron93} is not computationally efficient.

\begin{remark}[Sample complexity for Gaussian input]
If the input $x$ is Gaussian as
$x \sim \mathcal{N} (0,I_d)$,
then we know that $\E \left[ \left\| M_3(x) M_3^\top(x) \right\| \right] = \tl{O} \left( d^{3} \right)$ and $\E \left[ \left\| \Sc_2(x) \Sc_2^\top(x) \right\| \right] = \tl{O} \left( d^{2} \right)$, and the above sample complexity is simplified.
\end{remark}

\begin{remark}[Higher order tensorization] \label{remark:TensorizingGeneralization}
We stated that by tensorizing higher order tensors to lower order ones, we can estimate overcomplete models where the hidden layer dimension $k$ is larger than the input dimension $d$. We can generalize this idea to higher order tensorizing such that $m$ modes of the higher order tensor are tensorized into a single mode in the resulting lower order tensor. This enables us to estimate the models up to $k=O(d^m)$ assuming the matrix $A_1 \odot \dotsb \odot A_1$ ($m$ Khatri-Rao products) is full column rank. This is possible with the higher computational complexity.
\end{remark}


\begin{remark}[Effect of erroneous estimation of $p(x)$] \label{remark:score-error}
The input probability density function $p(x)$ is directly used in the Fourier part of the algorithm, and also  indirectly used in the tensor decomposition part   to compute the score function $\Sc_3(x)$; see~\eqref{eqn:diffoperator}. In the above analysis, to simplify the presentation, we assume we exactly know these functions, and thus, there is no additional error introduced by estimating them. It is straightforward to incorporate the corresponding errors in estimating input density into the final bound.
\end{remark}

 \begin{remark}[Alternative whitening prodecure] \label{remark:whitening}
In whitening Procedure~\ref{algo:whitening}, two options are provided for constructing the second order moment $M_2$. In the above analysis, we used the first option which exploits the second order score function. If any coefficient $\tl{\lambda}_j, j \in [k]$, in~\eqref{eqn:second-moment-score} vanishes, we cannot use the second order score function in the whitening procedure, and we use the other option for whitening; see Procedure~\ref{algo:whitening} for the details.
\end{remark}

\section{Risk Bound in the Non-realizable Setting} \label{sec:Generalization} 

In this section, we provide the risk bound for training the neural network with respect to an arbitrary target function; see Section~\ref{sec:problem-formulation} for the definition of the risk. 

In order to provide the risk bound with respect to an arbitrary target function, we also need to argue the approximation error in addition to the estimation error. For an arbitrary function $f(x)$, we need to find a neural network whose error in approximating the function can be bounded. We then combine it with the estimation error in training that neural network. This yields the final risk bound.

The approximation problem is about finding a neural network that approximates an arbitrary function $f(x)$ with bounded error. Thus, this is different from the realizable setting  where there is a fixed neural network and we only analyze its estimation.
\citet{Barron93} provides an approximation bound for the two-layer neural network and we exploit that here.
His result is based on the Fourier properties of function $f(x)$. Recall from~\eqref{eqn:FourierDef} the definition of Fourier transform of $f(x)$, denoted by $F(\omega)$, where $\omega$ is called the frequency variable. Define the first absolute moment of the Fourier magnitude distribution as
\begin{equation} \label{eqn:C_f}
C_f := \int_{\R^d} \|\omega\|_2 \cdot |F(\omega)| d \omega.
\end{equation}
\citet{Barron93} analyzes the approximation properties of 
\begin{equation} \label{eq:nn2-bounds}
\tl{f}(x) = \sum_{j \in [k]} a_2(j) \sigma \bigl( \inner{(A_1)_j, x} + b_1(j) \bigr), \quad \|(A_1)_j\|=1, |b_1(j)| \leq 1, |a_2(j)| \leq 2C_f, j \in [k],
\end{equation}
where the columns of weight matrix $A_1$ are the normalized version of random frequencies drawn from the Fourier magnitude distribution $|F(\omega)|$ weighted by the norm of the frequency vector. More precisely,
\begin{equation} \label{eqn:random_freq}
\omega_j \overset{\operatorname{i.i.d.}}{\sim} \frac{\|\omega\|}{C_f} |F(\omega)|, \quad (A_1)_j = \frac{\omega_j}{\|\omega_j\|}, \quad  j \in [k].
\end{equation}
See Section~\ref{sec:Proof-Approx1} for a detailed discussion on this connection between the columns of weight matrix $A_1$ and the random frequency draws from the Fourier magnitude distribution, and see how this is argued in the proof of the approximation bound.
The other parameters $a_2,b_1$ need to be also found. He then shows the following approximation bound for \eqref{eq:nn2-bounds}.
\begin{theorem}[Approximation bound, Theorem 3 of~\citet{Barron93}] \label{thm:approx-Barron}
For a function $f(x)$ with bounded $C_f$, there exists an approximation $\tl{f}(x)$ in the form of \eqref{eq:nn2-bounds} that satisfies the approximation bound 
$$
\Ebb_x [|\overline{f}(x) - \tl{f}(x)|^2] \leq O(r^2 C_f^2) \cdot \left(\frac{1}{\sqrt{k}} + \delta_1 \right)^2,
$$
where $\overline{f}(x) = f(x)-f(0)$.
Here, for $\tau>0$,
\begin{equation} \label{eqn:delta_tau-Def}
\delta_\tau := \inf_{0<\xi \leq 1/2} \left\{ 2 \xi + \sup_{|z| \geq \xi} \left| \sigma(\tau z) - 1_{\{z>0\}}(z) \right| \right\}
\end{equation}
is a distance between the unit step function $1_{\{z>0\}}(z)$ and the scaled sigmoidal function $\sigma(\tau z)$. 
\end{theorem}

See \citet{Barron93} for the complete proof of the above theorem. For completeness, we have also reviewed the main ideas of this proof in Section~\ref{sec:Proof-Approx}.
We now provide the formal statement of our risk bound.

\paragraph{Conditions for Theorem~\ref{thm:approx-guarantees}:}
\vspace{-0.05in}
\bi[itemsep=-0.1mm]
\item  The nonlinear activating function $\sigma(\cdot)$ is an arbitrary sigmoidal function satisfying the aforementioned Lipschitz condition. Note that a sigmoidal function is a bounded measurable function on the real line for which $\sigma(z) \rightarrow 1$ as $z \rightarrow \infty$ and $\sigma(z) \rightarrow 0$ as $z \rightarrow -\infty$.
\item For $\epsilon > 0$, suppose the number of samples $n$ satisfies (up to log factors)
\begin{align} \label{eqn:samplecomp-risk}
n \geq
\tl{O} \left(  \frac{k}{\epsilon^2} \right. & \cdot \E \left[ \left\| M_3(x) M_3^\top(x) \right\| \right] \\
& \hspace{-2em} \left.
\cdot \poly \left(
y_{\max},
\frac{\E \left[ \left\| \Sc_2(x) \Sc_2^\top(x) \right\| \right]}{\E \left[ \left\| M_3(x) M_3^\top(x) \right\| \right]},
 \frac{\zeta_{f}}{\psi},
\frac{\tl{\lambda}_{\max}}{\tl{\lambda}_{\min}},
\frac{1}{\lambda_{\min}},
\frac{s_{\max}(A_1)}{s_{\min}(A_1)},
|\Omega_l|, L, \frac{\|a_2\|}{(a_2)_{\min}},
|b_2|, \sigma_{\text{noise}}, \rho_\lambda \right)
\right), \nn
\end{align}
where $\zeta_{f} := \int_{\R^d} f(x)^2 dx$; notice the difference with $\tl{\zeta}_{\tl{f}}$.
Note that this is the same sample complexity as in Theorem~\ref{thm:guarantees-sample} with $\tl{y}_{\max}$ substituted with $y_{\max}$ and  $\tl{\zeta}_{\tl{f}}$ substituted with $\zeta_{f}$.
\item The target function $f(x)$ is bounded, and for $\epsilon > 0$, it has bounded $C_f$ as
\begin{align} \label{eqn:Cf-bound}
C_f \leq
\tl{O} \left(  \left( \frac{1}{\sqrt{k}} + \delta_1 \right)^{-1} \right. & \cdot \left( \frac{1}{\sqrt{k}} + \epsilon \right) \cdot \frac{1}{\sqrt{\E \left[ \left\| \Sc_3(x) \right\|^2 \right]}} \\
& \hspace{-5em} \left.
\cdot \poly \left(
\frac{1}{r},
\frac{\E \left[ \left\| \Sc_3(x) \right\|^2 \right]}{\E \left[ \left\| \Sc_2(x) \right\|^2 \right]},
\psi,
\frac{\tl{\lambda}_{\min}}{\tl{\lambda}_{\max}},
\lambda_{\min},
\frac{s_{\min}(A_1)}{s_{\max}(A_1)},
\frac{1}{|\Omega_l|}, \frac{1}{L}, \frac{(a_2)_{\min}}{\|a_2\|},
\frac{1}{|b_2|}, \frac{1}{\sigma_{\text{noise}}}, \frac{1}{\rho_\lambda} \right)
\right).	 \nn
\end{align}
See~\eqref{eqn:Cf-bound-tensordecomp} and \eqref{eqn:Cf-bound-fourier} for the complete form of bound on $C_f$.
For Gaussian input $x \sim \mathcal{N}(0,I_d)$, we have $\sqrt{\E \left[ \|\Sc_3(x)\|^2 \right]} = \tl{O} \left( d^{1.5} \right)$, and $r=\tl{O}(\sqrt{d})$.

See Corollary~\ref{cor:kernel} for examples of functions that satisfy this bound, and thus, we can learn them by the proposed method.
\item The coefficients
$\lambda_j  := \E \left[ \sigma'''(z_j) \right] \cdot a_2(j)$, and $\tl{\lambda}_j  := \E \left[ \sigma''(z_j) \right] \cdot a_2(j)$, $ j \in [k]$, in~\eqref{eqn:cross-moment-coeffs} and \eqref{eqn:cross-moment2-coeffs} are non-zero.
\item $k$ random i.i.d.\ draws of frequencies in Equation~\eqref{eqn:random_freq} are linearly independent. Note that the draws are from Fourier magnitude distribution\footnote{Note that it should be normalized to be a probability distribution as in~\eqref{eqn:random_freq}.} $\|\omega\| \cdot |F(\omega)|$. For more discussions on this condition, see Section~\ref{sec:Proof-Approx1} and earlier explanations in this section. In the overcomplete regime, $(k>d)$, the linear independence property needs to hold for appropriate tensorizations of the frequency draws. 
\ei

The above requirements on the number of samples $n$ and parameter $C_f$   depend on the parameters of the neural network $A_1$, $a_2$, $b_1$ and $b_2$. Note that there is also a dependence on these parameters through coefficients $\lambda_j$ and $\tl{\lambda}_j$. Since this is the non-realizable setting,  these neural network parameters correspond to the neural networks that satisfy the approximation bound proposed in Theorem~\ref{thm:approx-Barron} and are generated via random draws from the frequency spectrum of the function $f(x)$.

The proposed bound on $C_f$ in~\eqref{eqn:Cf-bound} is stricter when the number of hidden units $k$ increases. This might seem counter-intuitive, since the approximation result in Theorem~\ref{thm:approx-Barron} suggests that increasing $k$ leads to smaller approximation error. But, note that the approximation result in Theorem~\ref{thm:approx-Barron} does not consider efficient training of the neural network. The result in Theorem~\ref{thm:approx-guarantees} also deals with the efficient estimation of the neural network. This imposes additional constraint on the parameter $C_f$ such that when the number of neurons increases, the problem of learning the network weights   is more challenging for the tensor method to resolve.

\begin{theorem}[NN-LIFT guarantees: risk bound] \label{thm:approx-guarantees}
Suppose the above conditions hold.
Then the target function $f$ is approximated by the neural network $\hf$ which is learnt using  NN-LIFT in Algorithm~\ref{algo:NN-LIFT} satisfying   w.h.p.
\begin{equation*}
\Ebb_x [|f(x) - \hf(x)|^2] \leq O(r^2 C_f^2) \cdot \left(\frac{1}{\sqrt{k}} + \delta_1 \right)^2 + O(\epsilon^2),
\end{equation*}
where $\delta_\tau$ is defined in~\eqref{eqn:delta_tau-Def}.
Recall $x \in B_r$, where  $B_r := \{x: \|x\| \leq r\}$.
\end{theorem}

The theorem is mainly proved by combining the estimation bound guarantees in Theorem~\ref{thm:guarantees-sample}, and the approximation bound results for neural networks provided in Theorem~\ref{thm:approx-Barron}. But note that the approximation bound provided in Theorem~\ref{thm:approx-Barron} holds for a specific class of neural networks which are not generally recovered by the NN-LIFT algorithm. In addition, the estimation guarantees  in Theorem~\ref{thm:guarantees-sample} is for the realizable setting where the observations are the outputs of a fixed neural network, while in Theorem~\ref{thm:approx-guarantees} we observe samples of arbitrary function $f(x)$. Thus,  the approximation analysis in Theorem~\ref{thm:approx-Barron} can not be directly applied  to Theorem~\ref{thm:guarantees-sample}. For this, we need additional assumptions to ensure the NN-LIFT algorithm recovers a neural network which is close to one of the neural networks that satisfy the approximation bound in Theorem~\ref{thm:approx-Barron}. Therefore, we impose the bound on quantity $C_f$, and the full column rank assumption proposed in Theorem~\ref{thm:approx-Barron}. See Appendix~\ref{appendix:proof-risk} for the complete proof of Theorem~\ref{thm:approx-guarantees}.



The above risk bound includes two terms. The first term $O(r^2 C_f^2) \cdot \left(\frac{1}{\sqrt{k}} + \delta_1 \right)^2$ represents the approximation error on how the arbitrary function $f(x)$ with quantity $C_f$ can be approximated by the  neural network, whose weights are drawn from the Fourier magnitude distribution; see Theorem~\ref{thm:approx-Barron} for the formal statement. From the definition of $C_f$ in~\eqref{eqn:C_f},  this bound is weaker when the Fourier spectrum of target $f(x)$  has more energy in higher frequencies. This makes intuitive sense since it should be easier to approximate a function which is more smooth and has less fluctuations. The second term $O(\epsilon^2)$ is from estimation error for NN-LIFT algorithm, which is analyzed in Theorem~\ref{thm:guarantees-sample}. The polynomial factors for sample complexity in our estimation error are slightly worse than the bound provided in~\citet{barron1994approximation}, but note that we provide an estimation method which is both computationally and statistically efficient, while the method in~\citet{barron1994approximation} is not computationally efficient. Thus, for the first time, we have a computationally efficient method with guaranteed risk bounds for training neural networks.

\paragraph{Discussion on $\delta_\tau$ in the approximation bound:}
The approximation bound involves a term $\delta_\tau$ which is a constant and does not shrink with increasing the neuron size $k$. Recall that $\delta_\tau$ measures the  distance between the unit step function $1_{\{z>0\}}(z)$ and the scaled sigmoidal function $\sigma(\tau z)$ (which is used in the neural network specified in~\eqref{eq:nn2}). We now provide the following two observations

The above risk bound is only provided for the case $\tau=1$. We can generalize this result by imposing different constraint on the norm of columns of $A_1$ in~\eqref{eq:nn2-bounds}. In general, if we impose $\|(A_1)_j\|=\tau, j \in [k]$, for some $\tau>0$, then we have the approximation bound\footnote{Note that this change also needs some straightforward appropriate modifications in the algorithm.} $O(r^2 C_f^2) \cdot \left(\frac{1}{\sqrt{k}} + \delta_\tau \right)^2$. Note that $\delta_\tau \rightarrow 0$ when $\tau \rightarrow \infty$ (the scaled sigmoidal function $\sigma(\tau z)$ converges to the unit step function), and thus, this constant approximation error vanishes.

If the sigmoidal function is the unit step function as $\sigma(z) = 1_{\{z>0\}}(z)$, then $\delta_\tau=0$ for all $\tau>0$, and hence, there is no such constant approximation error.

%
%
%

\section{Discussions and Extensions}\label{sec:discussion}

In this section, we provide additional discussions. We first propose a toy example contrasting the hardness of optimization problems backpropagation and tensor decomposition. We then discuss the generalization of learning guarantees to higher dimensional output, and also the continuous output case.
We then discuss an alternative approach for estimating the low-dimensional parameters of the model.

\subsection{Contrasting the loss surface of backpropagation with tensor decomposition} \label{sec:toy-example}

We discussed that the computational hardness of training a neural network is due to the non-convexity of the loss function, and thus, popular local search methods such as backpropagation can get stuck in spurious local optima. We now provide a toy example highlighting this issue, and contrast it with the tensor decomposition approach.

\begin{figure}[t]
\subfloat[a][\small Classification setup]
{\begin{minipage}{2.2in}
\bc
\bp\psfrag{x1}[l]{$x_1$}\psfrag{x2}[l]{$x_2$}
\psfrag{y=1}[c]{\small $y\!=\!\tcb{1}$}\psfrag{y=-1}[c]{\small $\ \ \ y\!=\!\tcbrm{-1}$}
\psfrag{l}[c]{\tiny \quad\quad \tcr{Local optimum}}\psfrag{g}[l]{\tiny \tcdkg{Global optimum}}
\includegraphics[width=2.3in]{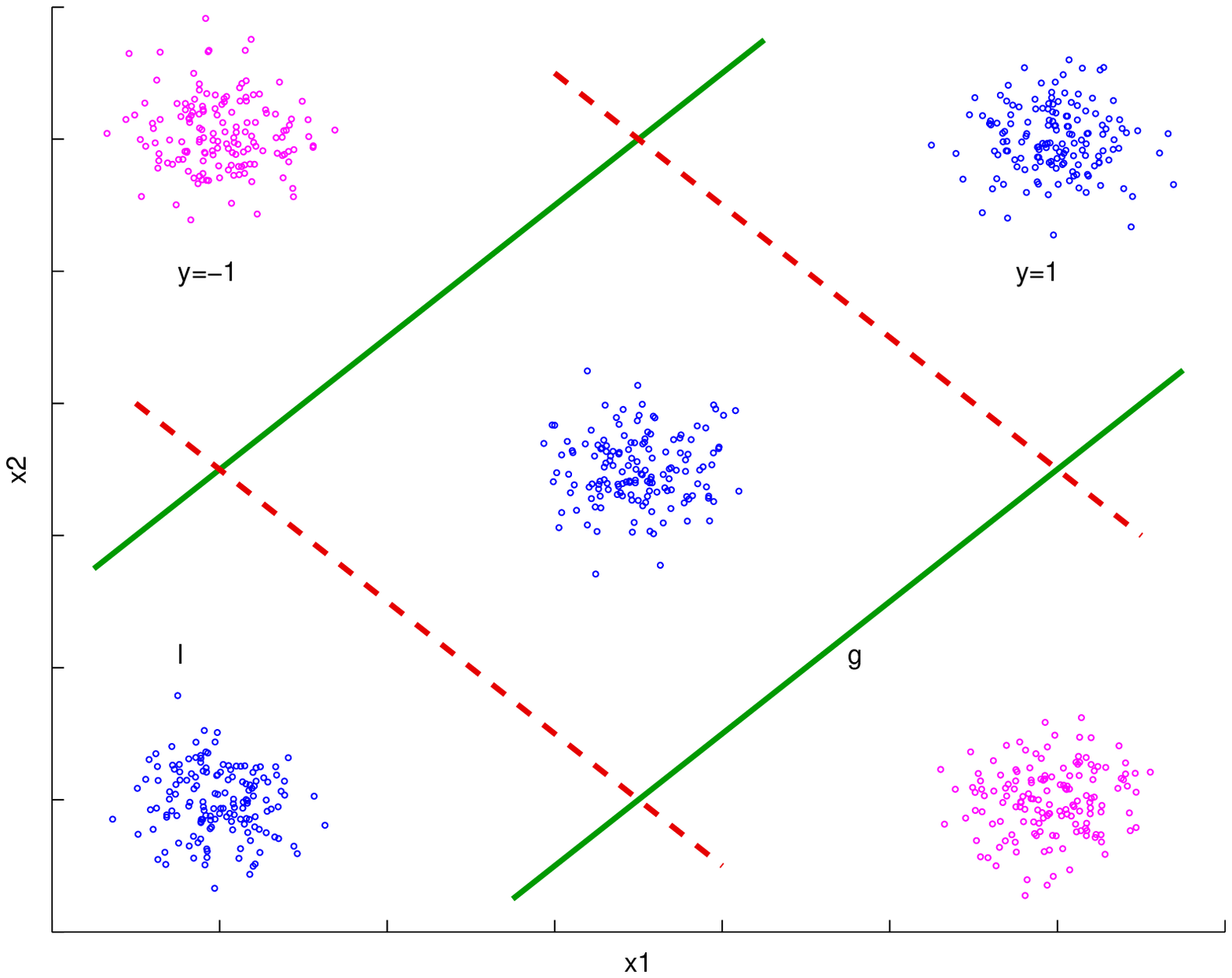}\ep
\ec
\end{minipage}
}\subfloat[(b)][\small Loss surface for backprop.]{\begin{minipage}{2.2in}
\bc
\bp\psfrag{w11}[c]{\footnotesize $A_1(1,1)$}\psfrag{w12}[l]{\footnotesize $A_1(2,1)$}
\includegraphics[width=2.3in]{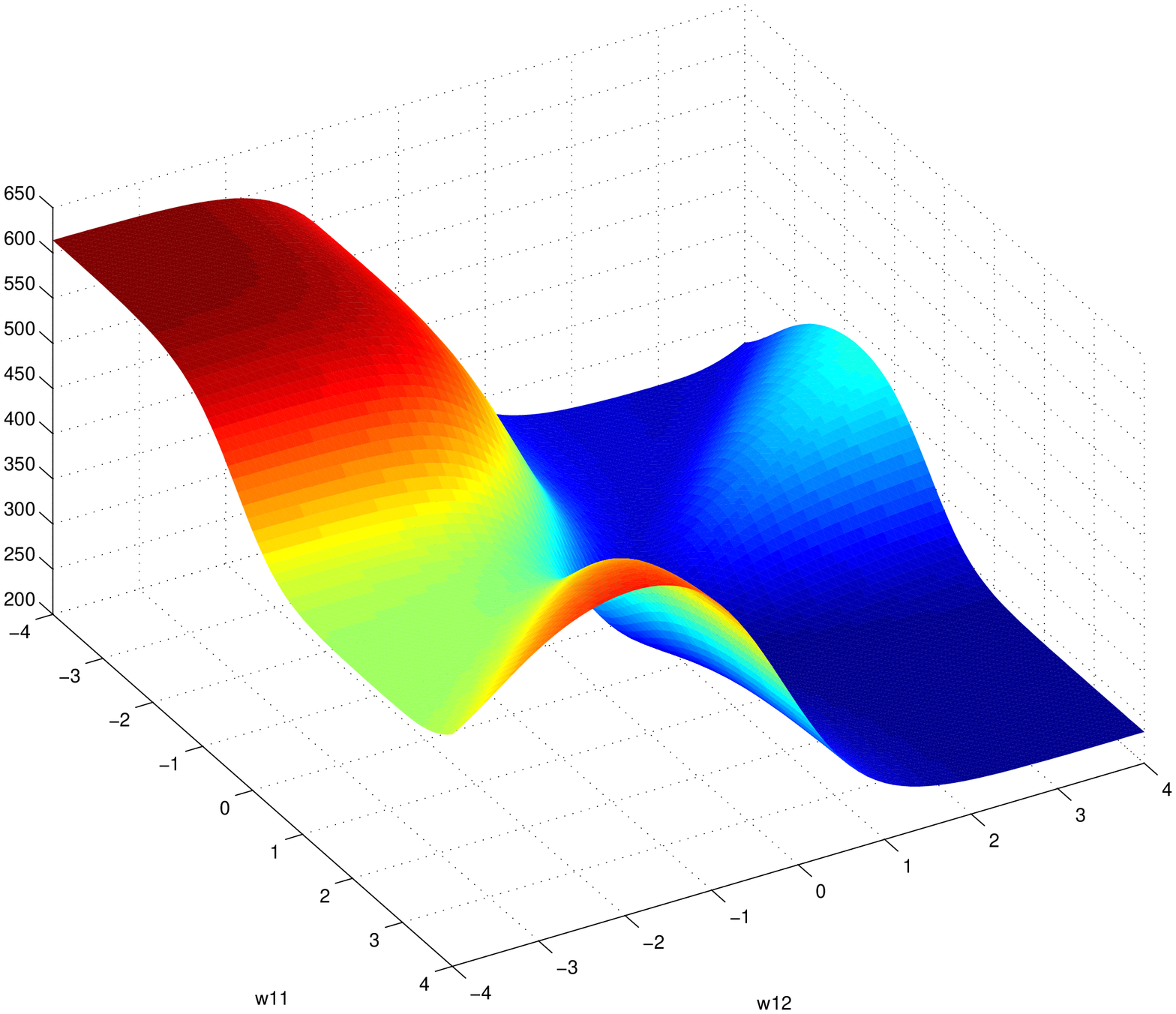}\ep
\ec\end{minipage}
}\subfloat[(c)][\small Loss surface for tensor method]{\begin{minipage}{2.2in}
\bc
\bp\psfrag{w11}[c]{\footnotesize $A_1(1,1)$}\psfrag{w12}[l]{\footnotesize $A_1(2,1)$}
\includegraphics[width=2.3in]{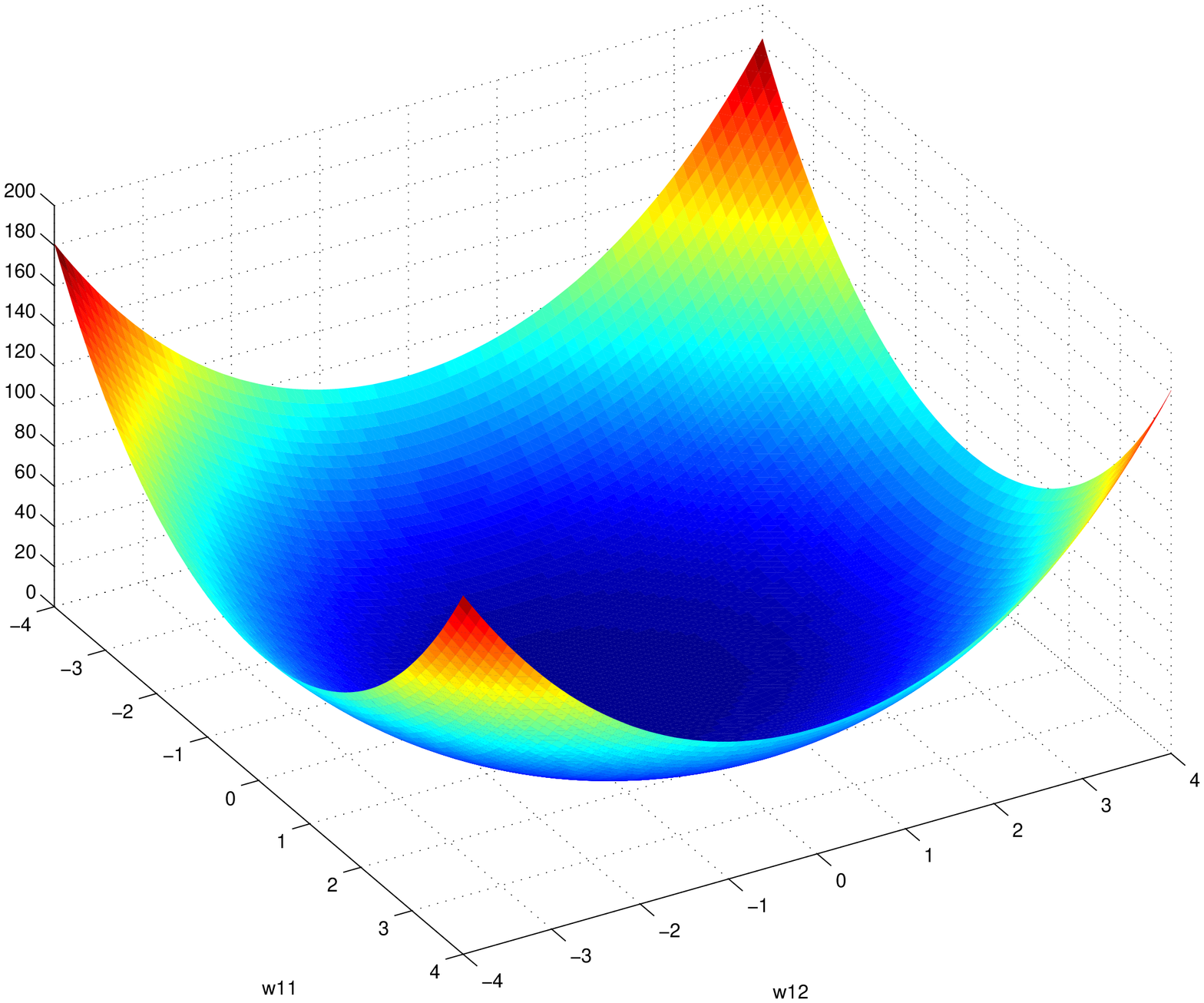}\ep
\ec\end{minipage}
}
\caption{\small (a) Classification task: two colors correspond to   binary labels. A two-layer neural network with two hidden neurons is used. Loss surface in terms of the first layer weights of one neuron (i.e., weights connecting the inputs to the neuron) is plotted while other  parameters are fixed. (b) Loss surface for usual square loss objective has spurious local optima. In part (a), one of the spurious local optima is drawn as red dashed lines and the global optimum is drawn as green solid lines. (c) Loss surface for  tensor factorization objective is free of spurious local optima. }\label{fig:toy-example}
\end{figure}

We consider a simple binary classification task shown in Figure~\ref{fig:toy-example}.a, where blue and magneta data points correspond to two different classes. It is clear that these two classes can be classified by a mixture of two linear classifiers which are drawn as green solid lines in the figure. For this task, we consider a two-layer neural network with two hidden neurons. The loss surfaces for backpropagation and tensor decomposition are shown in Figures~\ref{fig:toy-example}.b and \ref{fig:toy-example}.c, respectively. They are shown in terms of the weight parameters of inputs to the first neuron, i.e., the first column of matrix $A_1$, while the weight parameters to the second neuron are randomly drawn, and then fixed.

The stark contrast between the optimization landscape of tensor objective function, and the usual square loss objective used for backpropagation are observed, where even for a very simple classification task, backpropagation suffers from spurious local optima (one set of them is drawn as red dashed lines), which is not the case with tensor methods that is at least locally convex. This comparison highlights the algorithmic advantage of tensor decomposition compared to backpropagation in terms of the optimization they are performing.

\subsection{Extensions to cases beyond binary classification} \label{sec:discussion}
We earlier limited ourselves to the case where the output of neural network $\tl{y} \in \{0,1\}$ is  binary. These results can be easily extended to more complicated cases such as higher dimensional output (multi-label and multi-class), and also the continuous outputs (i.e., regression setting).   In the rest of this section, we discuss about the necessary changes in the algorithm to adapt it for these cases.

In the multi-dimensional case, the output label $\tl{y}$ is a vector generated as
$$\Ebb[\tl{y}|x] = A_2^\top \sigma(A_1^\top x+b_1)+b_2,$$
where the output is either discrete (multi-label and multi-class) or continuous.
Recall that the algorithm includes three main parts: tensor decomposition, Fourier and ridge regression components.

\paragraph{Tensor decomposition:}
For the tensor decomposition part, we first form the empirical version of $\tl{T}=\E \left[ \tl{y} \otimes \Sc_3(x) \right]$; note that $\otimes$ is used here (instead of scalar product used earlier) since $\tl{y}$ is not a scalar anymore. By the properties of score function, this tensor has decomposition form
$$\tl{T} = \Ebb \left[ \tl{y} \otimes \Pc_3(x) \right] = \sum_{j \in [k]} \E \left[ \sigma'''(z_j) \right] \cdot (A_2)^j \otimes (A_1)_j \otimes (A_1)_j \otimes (A_1)_j,$$
where $(A_2)^j$ denotes the $j^{\tha}$ row of matrix $A_2$. This is proved similar to  Lemma~\ref{lem:moment}. The tensor $\tl{T}$   is a fourth order tensor, and   we contract the first mode by multiplying it with a random vector $\theta$ as $\tl{T}(\theta,I,I,I)$ leading to the same form in~\eqref{eqn:cross-moment} as
$$\tl{T}(\theta,I,I,I) = \sum_{j \in [k]} \lambda_j \cdot (A_1)_j \otimes (A_1)_j \otimes (A_1)_j,$$
with $\lambda_j$ changed to $\lambda_j = \E \left[ \sigma'''(z_j) \right] \cdot \inner{(A_2)^j,\theta}$. Therefore, the same tensor decomposition guarantees in the binary case also hold here when the empirical version of $\tl{T}(\theta,I,I,I)$ is the input to the algorithm.

\paragraph{Fourier method:}
Similar to the scalar case, we can use one of the entries of output to estimate the entries of $b_1$.
There is an additional difference in the continuous case. Suppose that the output is generated as $\tl{y} = \tl{f}(x) + \eta$ where $\eta$ is noise vector which is independent of input $x$. In this case, the parameter $\tl{\zeta}_{\tl{f}}$ corresponding to $l^{\tha}$ entry of output $\tl{y}_l$ is changed to
$\tl{\zeta}_{\tl{f}} := \int_{\R^d} \tl{f}(x)_l^2 dx + \int_{\R} \eta_l^2 dt.$

\paragraph{Ridge regression:}
The ridge regression method and analysis can be immediately generalized to non-scalar output by applying the method independently to different entries of output vector to recover different columns of matrix $A_2$ and different entries of vector $b_2$.

\subsection{An alternative for estimating low-dimensional parameters} \label{sec:ft-alt}
Once we have an estimate of the first layer weights $A_1$, we can  greedily (i.e., incrementally) add neurons with the weight vectors $(A_1)_j$ for $j\in [k]$, and choose the bias $b_1(j)$ through grid search, and learn its contribution $a_2(j)$ for its final output. This is on the lines of the method proposed in~\citet{Barron93}, with one crucial difference that in our case, the first layer weights $A_1$ are already estimated by the tensor method. \citet{Barron93} proposes optimizing for each weight vector $(A_1)_j$ in $d$-dimensional space, whose computational complexity can scale exponentially in $d$ in the worst case. But, in our setup here, since we have already estimated the high-dimensional parameters (i.e., the columns of $A_1$), we only need to estimate a few low dimensional parameters. For the new hidden unit indexed by $j$, these parameters include the bias from input layer to the neuron (i.e., $b_1(j)$), and the weight from the neuron to the output (i.e., $a_2(j)$). This makes the approach computationally tractable, and we can even use brute-force or exhaustive search to find the best parameters on a finite set and get guarantees akin to~\citep{Barron93}.

\section{Proof Sketch} \label{sec:proofsketch}

In this section, we provide key ideas for proving the main results in Theorems~\ref{thm:guarantees-sample}~and~\ref{thm:approx-Barron}.

\subsection{Estimation bound}

The estimation bound is proposed in Theorem~\ref{thm:guarantees-sample}, and the complete proof is provided in Appendix~\ref{appendix:mainproof}.
Recall that NN-LIFT algorithm includes a tensor decomposition part for estimating $A_1$, a Fourier technique for estimating $b_1$, and a linear regression for estimating $a_2,b_2$. The application of linear regression in the last layer is immediately clear. In this section, we propose two main lemmas which clarify why the other methods are useful for estimating the unknown parameters $A_1, b_1$ in the realizable setting, where the label $\tl{y}$ is generated by  the neural network with the given architecture.

In the following lemma, we show how the cross-moment between label and score function as $\E[\tl{y} \cdot \Sc_3(x)]$ leads to a tensor decomposition form for estimating weight matrix $A_1$.

\begin{lemma} \label{lem:moment}
For the two-layer neural network specified in~\eqref{eq:nn2}, we have
\begin{equation} \label{eqn:cross-moment}
\Ebb \left[ \tl{y} \cdot \Pc_3(x) \right] = \sum_{j \in [k]} \lambda_j \cdot (A_1)_j \otimes (A_1)_j \otimes (A_1)_j,
\end{equation}
where $(A_1)_j \in \R^d$ denotes the $j$-th column of $A_1$, and 
\begin{equation} \label{eqn:cross-moment-coeffs}
\lambda_j = \E \left[ \sigma'''(z_j) \right] \cdot a_2(j),
\end{equation}
for vector $z := A_1^\top x+b_1$ as the input to the nonlinear operator $\sigma(\cdot)$.
\end{lemma}

This is proved  by the main property of score functions as yielding differential operators, where for label-function $f(x) := \E[y|x]$, we have $\E[y \cdot \Sc_3(x)] = \E[\nabla_x^{(3)} f(x)]$~\citep{janzamin2014matrix}; see Section~\ref{appendix:tensor-proof} for a complete proof of the lemma. This lemma shows that by decomposing the cross-moment tensor $\E[\tl{y} \cdot \Sc_3(x)]$, we can recover the columns of $A_1$.



We also exploit the phase of complex number $v$ to estimate the bias vector $b_1$; see Procedure~\ref{algo:Fourier}. 
The following lemma clarifies this. The perturbation analysis is provided in the appendix.


\torestate{Lemma}{lem:v-mean}{
Let 
\begin{equation} \label{eqn:v-appendix}
\tl{v} := \frac{1}{n} \sum_{i \in [n]} \frac{\tl{y}_i}{p(x_i)} e^{-j \inner{\omega_i, x_i}}.
\end{equation}
Notice this is a realizable of $v$ in Procedure~\ref{algo:Fourier} where the output corresponds to a neural network $\tl{y}$.
If $\omega_i$'s are uniformly i.i.d.\ drawn from set $\Omega_l$, then $\tl{v}$ has mean (which is computed over $x$, $\tl{y}$ and $\omega$)
\begin{equation} \label{eqn:v-mean}
\E[\tl{v}] = \frac{1}{|\Omega_l|} \Sigma \left( \frac{1}{2} \right) a_2(l) e^{j \pi b_1(l)},
\end{equation}
where $|\Omega_l|$ denotes the surface area of $d-1$ dimensional manifold $\Omega_l$, and $\Sigma(\cdot)$  denotes  the Fourier transform of $\sigma(\cdot)$.
}

This lemma is proved in Appendix~\ref{appendix:Fourier}.

\subsection{Approximation bound} \label{sec:Proof-Approx}


We exploit the approximation bound argued in~\citet{Barron93} provided in Theorem~\ref{thm:approx-Barron}. We first discuss his main result arguing an approximation bound $O(r^2C_f^2/k)$ for a function $f(x)$ with bounded parameter $C_f$; see~\eqref{eqn:C_f} for the definition of $C_f$. Note that this corresponds to the first term in the approximation error proposed in Theorem~\ref{thm:approx-Barron}. For this result, \citet{Barron93} does not consider any bound on the parameters of first layer $A_1$ and $b_1$. He then provides a refinement of this result where he also bounds the parameters of neural network as we also do in~\eqref{eq:nn2-bounds}. This leads to the additional term involving $\delta_\tau$ in the approximation error as seen in Theorem~\ref{thm:approx-Barron}. Note that bounding the parameters of neural network is also useful in learning these parameters with computationally efficient algorithms since it limits the searching space for training these parameters. We now provide the main ideas of proving these bounds as follows.


\subsubsection{No bounds on the parameters of the neural network} \label{sec:Proof-Approx1}
We first provide the proof outline when there is no additional constraints on the parameters of neural network; see set $G$ defined in~\eqref{eqn:set-G}, and compare it with the form we use in~\eqref{eq:nn2-bounds} where there are additional bounds. In this case, \citet{Barron93} argues approximation bound $O(r^2C_f^2/k)$ which is proved based on two main results.
The first result says that if a function $f$ is in the closure of the convex hull of a set $G$ in a Hilbert space, then for every $k \geq 1$, there is an $f_k$ as the convex combination of $k$ points in $G$ such that
\begin{equation} \label{eqn:ConvexHull-Lemma}
\E[|f -  f_k|^2] \leq \frac{c'}{k},
\end{equation}
for any constant $c'$ satisfying some lower bound related to the properties of set $G$ and function $f$; see Lemma 1 in~\citet{Barron93} for the precise statement and the proof of this result.

The second part of the proof is to argue that arbitrary function $f \in \Gamma$ (where $\Gamma$ denotes the set of functions with bounded $C_f$) is in the closure of the convex hull of sigmoidal functions 
\begin{equation} \label{eqn:set-G}
G := \bigl\{ \gamma \sigma \bigl( \inner{\alpha, x} + \beta \bigr) : \alpha \in \R^d, \beta \in \R, |\gamma| \leq 2C \bigr\}.
\end{equation}
\citet{Barron93} proves this result by arguing the following chain of inclusions as
$$
\Gamma \subset \operatorname{cl} G_{\operatorname{cos}} \subset \operatorname{cl} G_{\operatorname{step}} \subset \operatorname{cl} G,
$$
where $\operatorname{cl} G$ denotes the closure of set $G$, and sets $G_{\operatorname{cos}}$ and $G_{\operatorname{step}}$ respectively denote set of some sinusoidal and step functions. See Theorem 2 in~\citet{Barron93} for the precise statement and the proof of this result. 

\paragraph{Random frequency draws from Fourier magnitude distribution:}
Recall from Section~\ref{sec:Generalization} that the columns of weight matrix $A_1$ are the normalized version of random frequencies drawn from Fourier magnitude distribution $\|\omega\| \cdot |F(\omega)|$; see Equation~\eqref{eqn:random_freq}. This connection is along the proof of relation $\Gamma \subset \operatorname{cl} G_{\operatorname{cos}}$ that we recap here; see proof of Lemma 2 in~\citet{Barron93} for more details.
By expanding the Fourier transform as magnitude and phase parts $F(\omega) = e^{j \theta(\omega)} |F(\omega)|$, we have
\begin{equation} \label{eqn:cos-proof1}
\overline{f}(x) := f(x) - f(0) = \int g(x,\omega) \Lambda(d\omega),
\end{equation}
where
\begin{equation} \label{eqn:Lambda-distribution}
\Lambda(\omega) := \|\omega\| \cdot |F(\omega)|/C_f
\end{equation} 
is the normalized Fourier magnitude distribution (as a probability distribution) weighted by the norm of frequency vector, and
$$g(x,\omega) := \frac{C_f}{\|\omega\|} \left( \cos(\inner{\omega,x} +\theta(\omega)) - \cos(\theta(\omega)) \right).$$
The integral in~\eqref{eqn:cos-proof1} is an infinite convex combination of functions in the class
$$G_{\operatorname{cos}} := \left\{ \frac{\gamma}{\|\omega\|} \left( \cos (\inner{\omega,x} + \beta) - \cos(\beta) \right) : \omega \neq 0, |\gamma| \leq C, \beta \in \R \right\}.$$
Now if $\omega_1,\omega_2,\dotsc,\omega_k$ is a random sample of $k$ points independently drawn from Fourier magnitude distribution $\Lambda$, then by Fubini's Theorem, we have
$$\Ebb \int_{B_r} \left( f(x) - \frac{1}{k} \sum_{j \in [k]} g(x,\omega_j) \right)^2 \mu(dx) \leq \frac{C^2}{k},$$
where $\mu(\cdot)$ is the probability measure for $x$.
This shows function $\overline{f}$ is in the convex hull of $G_{\operatorname{cos}}$. Note that the bound $\frac{C^2}{k}$ complies the bound in~\eqref{eqn:ConvexHull-Lemma}.

\subsubsection{Bounding the parameters of the neural network}
\citet{Barron93} then imposes additional bounds on the weights of first layer, considering the following class of sigmoidal functions as
\begin{equation} \label{eqn:set-Gtau}
G_\tau := \bigl\{ \gamma \sigma \bigl( \tau ( \inner{\alpha, x} + \beta ) \bigr) : \|\alpha\| \leq 1, |\beta| \leq 1, |\gamma| \leq 2C \bigr\}.
\end{equation}
Note that the approximation proposed in~\eqref{eq:nn2-bounds} is a convex combination of $k$ points in~\eqref{eqn:set-Gtau} with $\tau=1$. \citet{Barron93} concludes Theorem~\ref{thm:approx-Barron} by the following lemma.  
\begin{lemma}[Lemma 5 in~\citet{Barron93}]
If $g$ is a function on $[-1,1]$ with derivative bounded\footnote{Note that the condition on having bounded derivative does not rule out cases such as step function as the sigmoidal function. This is because similar to the analysis for the main case (no bounds on the weights), we first argue that function $f$ is in the closure of functions in $G_{\operatorname{cos}}$ which are univariate functions with bounded derivative.} by a constant $C$, then for every $\tau>0$, we have
\begin{equation*}
\inf_{g_\tau \in \operatorname{cl} G_\tau} \sup_{|z| \leq \tau} |g(z) - g_\tau(z)| \leq 2C \delta_\tau.
\end{equation*}
\end{lemma}
Finally Theorem~\ref{thm:approx-Barron} is proved by applying triangle inequality to bounds argued in the above two cases.

\section{Conclusion}
We have proposed  a novel algorithm   based on tensor decomposition   for training two-layer neural networks. This is a computationally efficient method with guaranteed risk bounds with respect to the target function under polynomial sample complexity in the input and neuron dimensions. The tensor method   is embarrassingly parallel and has a parallel time computational complexity which is logarithmic in input dimension   which is comparable with parallel stochastic backpropagation.  There are number of open problems to consider in future. Extending this framework to a multi-layer network is of great interest. Exploring the score function framework to train other discriminative models is also interesting.

\subsubsection*{Acknowledgements}
We thank Ben Recht for pointing out to us the Barron's work on approximation bounds for neural networks~\citep{Barron93}, and thank Arthur Gretton for discussion about estimation of score function.  We also acknowledge fruitful discussion with Roi Weiss about the presentation of proof of Theorem~\ref{thm:approx-guarantees} on combining estimation and approximation bounds, and his detailed editorial comments about the preliminary version of the draft. We thank Peter Bartlett for detailed discussions and pointing us to several classical results on neural networks. We are very thankful for Andrew Barron for detailed discussion and for encouraging us to explore alternate incremental training methods for estimating remaining parameters  after the tensor decomposition step and we have added this discussion to the paper. We thank Daniel Hsu for discussion on random design analysis of ridge regression. We thank Percy Liang for discussion about score function.

M. Janzamin is supported by NSF BIGDATA award FG16455. H. Sedghi is supported by NSF Career award FG15890. A. Anandkumar is supported in part by Microsoft Faculty Fellowship, NSF Career award CCF-$1254106$,   and ONR Award N00014-14-1-0665.

%

\appendix

\section{Tensor Notation}
In this Section, we provide the additional tensor notation required for the analysis provided in the supplementary material.
\paragraph{Multilinear form:}
The multilinear form for a tensor $T \in \Rbb^{q_1 \times q_2 \times q_3}$ is defined as follows. Consider matrices $M_r \in \R^{q_r \times p_r}, r \in \{1,2,3\}$. Then tensor $T(M_1,M_2,M_3) \in \R^{p_1 \times p_2 \times p_3}$ is defined as
\begin{equation} \label{eqn:MulitilinearDef}
T(M_1,M_2,M_3) := \sum_{j_1 \in [q_1]} \sum_{j_2 \in [q_2]} \sum_{j_3 \in [q_3]} T_{j_1,j_2,j_3} \cdot M_1(j_1, :)  \otimes M_2(j_2, :) \otimes M_3(j_3, :).
\end{equation}

As a simpler case, for vectors $u,v,w \in \R^d$, we have\,\footnote{Compare with the matrix case where for $M \in \R^{d \times d}$, we have $ M(I,u) = Mu := \sum_{j \in [d]} u_j M(:,j)$.}
\begin{equation} \label{eqn:rank-1update2}
 T(I,v,w) := \sum_{j,l \in [d]} v_j w_l T(:,j,l) \ \in \R^d,
\end{equation}
which is a multilinear combination of the tensor mode-$1$ fibers.

\section{Details of Tensor Decomposition Algorithm} \label{appendix:tensordecomp}

\begin{figure}[t]
\bc
\begin{tikzpicture}
[
scale=1,
   nodestyle/.style={fill = gray!30, shape = rectangle, rounded corners, minimum width = 2cm},
]
\small
\matrix [column sep=2mm,row sep=3mm] {
\node[nodestyle](a1){Input: Tensor $T=\sum_{i \in [k]} \lambda_i u_i^{\otimes 3}$}; & \\
\node[nodestyle](w){Whitening procedure (Procedure~\ref{algo:whitening})}; & \\
\node[nodestyle](a){SVD-based Initialization (Procedure~\ref{algo:SVD init})}; & \\
\node[nodestyle, align=center](b){Tensor Power Method (Algorithm~\ref{alg:robustpower} )}; & \\
\node[nodestyle](e){Output: $\lbrace u_i  \rbrace_{i \in [k]}$}; & \\
};
\draw [->, line width = 1pt] (a1) to (w);
\draw [->, line width = 1pt] (w) to (a);
\draw [->, line width = 1pt] (a) to (b);
\draw [->, line width = 1pt] (b) to (e);

\end{tikzpicture}
\ec
\caption{\small Overview of tensor decomposition algorithm for third order tensor (without tensorization).}
\label{fig:TensorDecomposition}
\end{figure}
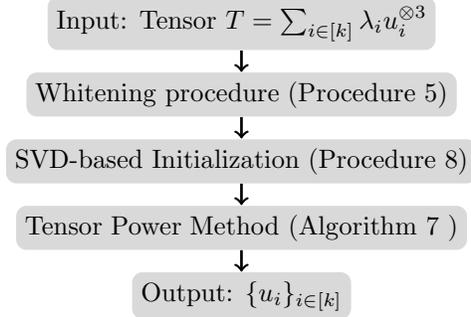

The goal of tensor decomposition algorithm is to recover the rank-1 components of tensor; refer to Equation~\eqref{eqn:tensordecomp} for the notion of tensor rank and its rank-1 components. We exploit the tensor decomposition algorithm proposed in~\citep{JMLR:v15:anandkumar14b,anandkumar2014guaranteed}. 
Figure~\ref{fig:TensorDecomposition} depicts the flowchart of this method where the corresponding algorithms and procedures  are also specified. Similarly, Algorithm~\ref{algo:main} states the high-level steps of tensor decomposition algorithm. 
The main step of the tensor decomposition method is the {\em tensor power iteration} which is the generalization of matrix power iteration to $3$rd order tensors. The tensor power iteration is given by
$$u \leftarrow \frac{T(I,u,u)}{\|T(I,u,u)\|},$$
where $u \in \R^d, T(I,u,u) :=  \sum_{j,l \in [d]} u_j u_l T(:,j,l) \in \R^d$ is a {\em multilinear} combination of tensor {\em fibers}. Note that tensor fibers are the vectors which are derived by fixing all the indices of the tensor except one of them, e.g., $T(:,j,l)$ in the above expansion. 
The initialization for different runs of tensor power iteration is performed by the SVD-based technique proposed in Procedure~\ref{algo:SVD init}. This helps to initialize non-convex tensor power iteration with good initialization vectors. 
The whitening preprocessing is applied to orthogonalize the components of input tensor. Note that the convergence guarantees of tensor power iteration for orthogonal tensor decomposition have been developed in the literature~\citep{ZG01,JMLR:v15:anandkumar14b}.


\floatname{algorithm}{Algorithm}
\begin{algorithm}[t]
\caption{Tensor Decomposition Algorithm Setup}
\label{algo:main}
\begin{algorithmic}[1]
\renewcommand{\algorithmicrequire}{\textbf{input}}
\renewcommand{\algorithmicensure}{\textbf{output}}
\REQUIRE symmetric tensor $T$.

\IF{Whitening}
\STATE Calculate $T=$Whiten($T$); see Procedure~\ref{algo:whitening}.
\ELSIF{Tensorizing}
\STATE Tensorize the input tensor.
\STATE Calculate $T=$Whiten($T$); see Procedure~\ref{algo:whitening}.
\ENDIF
\FOR{$j=1$ to $k$}
\STATE $(v_j , \mu_j, T)=\text{tensor power decomposition}(T)$; see Algorithm~\ref{alg:robustpower}.
\ENDFOR
\STATE $(A_1)_j = \operatorname{Un-whiten}(v_j), j \in [k]$; see Procedure~\ref{algo:Un-whitening}.
\RETURN $\lbrace (A_1)_j \rbrace_{j \in [k]}.$
\end{algorithmic}
\end{algorithm}

The tensorization step works as follows.

\paragraph{Tensorization:}The tensorizing step is applied when we want to decompose overcomplete tensors where the rank $k$ is larger than the dimension $d$. For instance, for getting rank up to $k = O(d^2)$, we first form the 6th order input tensor with decomposition as 
$$T = \sum_{j \in [k]} \lambda_j a_j^{\otimes 6} \in \bigotimes^6 \R^d.$$ 
Given $T$, we form the 3rd order tensor $\tl{T} \in \bigotimes^3 \R^{d^2}$ which is the tensorization of $T$ such that
\begin{equation} \label{eqn:tensorization}
\tl{T} \bigl( i_2+d(i_1-1), j_2+d(j_1-1), l_2+d(l_1-1) \bigr) := T(i_1,i_2,j_1,j_2,l_1,l_2).
\end{equation}
This leads to $\tl{T}$ having decomposition
$$\tl{T} =  \sum_{j \in [k]} \lambda_j (a_j \odot a_j)^{\otimes 3}.$$
We then apply the tensor decomposition algorithm to this new tensor $\tl{T}$. This now clarifies why the full column rank condition is applied to the columns of $A \odot A = [a_1 \odot a_1 \dotsb a_k \odot a_k]$. Similarly, we can perform higher order tensorizations leading to more overcomplete models by exploiting initial higher order tensor $T$; see also Remark~\ref{remark:TensorizingGeneralization}.

\paragraph{Efficient implementation of tensor decomposition given samples:}The main update steps in the tensor decomposition algorithm is the tensor power iteration for which a multilinear operation is performed on tensor $T$.  However, the tensor is not available beforehand, and needs to be estimated using the samples (as in Algorithm~\ref{algo:NN-LIFT} in the main text). Computing and storing the tensor can be enormously expensive for high-dimensional problems. But, it is essential to note that since we can form a factor form of tensor $T$ using the samples and other parameters in the model, we can manipulate the samples directly to perform the power update as {\em multi-linear} operations without explicitly forming the tensor. This leads to efficient computational complexity. See~\citep{JanzaminEtal:Altmin14} for details on these implicit update forms.

\floatname{algorithm}{Procedure}
\begin{algorithm}[t]
\caption{Whitening}
\label{algo:whitening}
\begin{algorithmic}[1]
\renewcommand{\algorithmicrequire}{\textbf{input}}
\renewcommand{\algorithmicensure}{\textbf{output}}
\REQUIRE Tensor $T \in \Rbb^{d \times d \times d}$.
\STATE Second order moment $M_2 \in \R^{d \times d}$ is constructed such that it has the same decompositon form as target tensor $T$ (see Section~\ref{appendix:whitening} for more discussions):
\bi
\item Option 1: constructed using second order score function; see Equation~\eqref{eqn:second-moment-score}.
\item Option 2: computed as $M_2:=T(I,I,\theta) \in \R^{d \times d}$, where $\theta \sim \mathcal{N}(0,I_{d})$ is a random standard Gaussian vector.
\ei
\STATE Compute the rank-k SVD, $M_2=U \Diag(\gamma) U^\top$, where $U \in \R^{d \times k}$ and $\gamma \in \R^k$.
\STATE Compute the whitening matrix $W:=U \Diag(\gamma^{-1/2}) \in \R^{d \times k}$.
\RETURN $T\left(W,W,W\right) \in \R^{k \times k \times k}$.
\end{algorithmic}
\end{algorithm}

\floatname{algorithm}{Procedure}
\begin{algorithm}[t]
\caption{Un-whitening}
\label{algo:Un-whitening}
\begin{algorithmic}[1]
\renewcommand{\algorithmicrequire}{\textbf{input}}
\renewcommand{\algorithmicensure}{\textbf{output}}
\REQUIRE Orthogonal rank-1 components $v_j \in \R^k, j \in [k]$.
\STATE Consider matrix $M_2$ which was exploited for whitening in Procedure~\ref{algo:whitening}, and let $\tl{\lambda}_j, j \in [k]$ denote the corresponding coefficients as $M_2 = A_1 \Diag(\tl{\lambda}) A_1^\top$; see~\eqref{eqn:second-moment-score}.
\STATE Compute the rank-k SVD, $M_2=U \Diag(\gamma) U^\top$, where $U \in \R^{d \times k}$ and $\gamma \in \R^k$.
\STATE Compute
$$(A_1)_j = \frac{1}{\sqrt{\tl{\lambda}_j}} U \Diag(\gamma^{1/2}) v_j, \quad j \in [k].$$
\RETURN $\left\{ (A_1)_j \right\}_{j \in [k]}$.
\end{algorithmic}
\end{algorithm}

\floatname{algorithm}{Algorithm}
\begin{algorithm}[h]
\caption{Robust tensor power method~\citep{JMLR:v15:anandkumar14b}}
\label{alg:robustpower}
\begin{algorithmic}[1]
\renewcommand{\algorithmicrequire}{\textbf{input}}
\renewcommand{\algorithmicensure}{\textbf{output}}
\REQUIRE symmetric tensor $\tilde{T} \in \R^{d' \times d' \times d'}$, number of
iterations $N$, number of initializations $R$. 

\ENSURE the estimated eigenvector/eigenvalue pair; the deflated tensor.

\FOR{$\tau = 1$ to $R$}

\STATE Initialize $\hv_0^{(\tau)}$ with SVD-based method in Procedure~\ref{algo:SVD init}.

\FOR{$t = 1$ to $N$}

\STATE Compute power iteration update
\begin{equation}
\hv_{t}^{(\tau)} :=
\frac{\tilde{T}(I, \hv_{t-1}^{(\tau)}, \hv_{t-1}^{(\tau)})}
{\|\tilde{T}(I, \hv_{t-1}^{(\tau)}, \hv_{t-1}^{(\tau)})\|}
\label{eq:power-update}
\end{equation}

\ENDFOR

\ENDFOR


%
\STATE Let $\tau^* := \arg\max_{\tau \in [R]} \{ \tilde{T}(\hv_{N}^{(\tau)},
\hv_{N}^{(\tau)}, \hv_{N}^{(\tau)}) \}$.

\STATE Do $N$ power iteration updates \eqref{eq:power-update} starting from
$\hv_{N}^{(\tau^*)}$ to obtain $\hv$, and set $\hat\mu :=
\tilde{T}(\hv,\hv,\hv)$.

\RETURN the estimated eigenvector/eigenvalue pair
$(\hv,\hat{\mu})$; the deflated tensor $\tilde{T} - \hat{\mu}\cdot \hat{v}^{\otimes 3}$.

\end{algorithmic}
\end{algorithm} 

\floatname{algorithm}{Procedure}
\begin{algorithm}[h]
\caption{SVD-based initialization~\citep{anandkumar2014guaranteed}}
\label{algo:SVD init}
\begin{algorithmic}[1]
\renewcommand{\algorithmicrequire}{\textbf{input}}
\renewcommand{\algorithmicensure}{\textbf{output}}
\REQUIRE Tensor $T \in \Rbb^{d' \times d' \times d'}$.
\FOR{$\tau = 1$ to $\log (1/\hat{\delta})$}
\STATE Draw a random standard Gaussian vector $\theta^{(\tau)} \sim \mathcal{N}(0,I_{d'}).$
\STATE Compute $u_1^{(\tau)}$ as the top left singular vector of  $T(I,I,\theta^{(\tau)}) \in \R^{d' \times d'}$.
\ENDFOR
\STATE $\hv_0 \leftarrow \max_{\tau \in [\log (1/\hat{\delta})]} \left( u_1^{(\tau)} \right)_{\min}$.
\RETURN $ \hv_0 $.
\end{algorithmic}
\end{algorithm}

\section{Proof of Theorem~\ref{thm:guarantees-sample}} \label{appendix:mainproof}

Proof of Theorem~\ref{thm:guarantees-sample} includes three main pieces which is about arguing the recovery guarantees of three different parts of the algorithm: tensor decomposition, Fourier method, and linear regression. As the first piece, we show that the tensor decomposition algorithm for estimating weight matrix $A_1$ (see Algorithm~\ref{algo:NN-LIFT} for the details) recovers it with the desired error. In the second part, we analyze the performance of Fourier technique for estimating bias vector $b_1$ (see Algorithm~\ref{algo:NN-LIFT} and Procedure~\ref{algo:Fourier} for the details) proving the error in the recovery is small. Finally as the last step, the ridge regression is analyzed to ensure that the parameters of last layer of the neural network are well estimated leading to the estimation of overall function $\tl{f}(x)$.
We now provide the analysis of these three parts.

\subsection{Tensor decomposition guarantees} \label{appendix:tensor-proof}

We first provide a short proof for Lemma~\ref{lem:moment} which shows how the rank-1 components of third order tensor $\E \left[ \tl{y} \cdot \Pc_3(x) \right]$ are the columns of weight matrix $A_1$.

\bprfof{Lemma~\ref{lem:moment}}
It is shown by~\citet{janzamin2014matrix} that the score function yields differential operator such that for label-function $f(x) := \E[y|x]$, we have $$\E[y \cdot \Sc_3(x)] = \E[\nabla_x^{(3)} f(x)].$$
Applying this property to the form of label function $f(x)$ in~\eqref{eq:nn2} denoted by $\tl{f}(x)$, we have
$$ \Ebb \left[ \tl{y} \cdot \Pc_3(x) \right] = \Ebb[\sigma'''(\cdot)(a_2, A_1^\top,A_1^\top, A_1^\top)],$$
where $\sigma'''(\cdot)$ denotes the third order derivative of element-wise function $\sigma(z): \R^k \rightarrow \R^k$. More concretely, with slightly abuse of notation, $\sigma'''(z) \in \R^{k \times k \times k \times k}$ is a diagonal 4th order tensor  with its $j$-th diagonal entry equal to $\frac{\partial^3 \sigma(z_j)}{\partial z_j^3}: \R \rightarrow \R$.
Here two properties are used to compute the third order derivative $\nabla_x^{(3)} \tl{f}(x)$ on the R.H.S.\ of above equation as follows. 1) We apply chain rule to take the derivatives which generates a new factor of $A_1$ for each derivative. Since we take 3rd order derivative, we have 3 factors of $A_1$. 2) The linearity of next layers leads to the derivatives from them being vanished, and thus, we only have the above term as the derivative. Expanding the above multilinear form finishes the proof; see~\eqref{eqn:MulitilinearDef} for the definition of multilinear form.
\eprfof

We now provide the recovery guarantees of weight matrix $A_1$ through tensor decomposition as follows.

\begin{lemma} \label{lem:A1-guarantee}
Among the conditions for Theorem~\ref{thm:guarantees-sample}, consider the rank constraint on $A_1$, and the non-vanishing assumption on coefficients $\lambda_j$'s. Let the whitening to be performed using empirical version of second order score function as specified in~\eqref{eqn:second-moment-score}, and assume the coefficients $\tl{\lambda}_j$'s do not vanish. Suppose the sample complexity
\begin{align} \label{eqn:samplecomp-tensordecomp}
n \geq \max & \left\{
\tl{O} \left( \tl{y}_{\max}^2 \E \left[ \left\| M_3(x) M_3^\top(x) \right\| \right] \frac{\tl{\lambda}_{\max}^4}{\tl{\lambda}_{\min}^4} \frac{s_{\max}^2(A_1)}{\lambda_{\min}^2 \cdot s_{\min}^6(A_1)} \cdot \frac{1}{\tl{\epsilon}_1^2} \right), \right. \\
& \quad \tl{O} \left( \tl{y}_{\max}^2 \cdot \E \left[ \left\| M_3(x) M_3^\top(x) \right\| \right] \cdot \left( \frac{\tl{\lambda}_{\max}}{\tl{\lambda}_{\min}} \right)^3 \frac{1}{\lambda_{\min}^2 \cdot s_{\min}^6(A_1)} \cdot k \right), \nn \\
& \quad \left. \tl{O} \left( \tl{y}_{\max}^2 \cdot \frac{\E \left[ \left\| \Sc_2(x) \Sc_2^\top(x) \right\| \right]^{3/2}}{\E \left[ \left\| M_3(x) M_3^\top(x) \right\| \right]^{1/2}} \cdot \frac{1}{\tl{\lambda}_{\min}^2 \cdot s_{\min}^3(A_1)} \right) \nn
\right\},
\end{align}
holds, where $M_3(x) \in \R^{d \times d^2}$ denotes the matricization of score function tensor $\Sc_3(x) \in \R^{d \times d \times d}$; see~\eqref{eqn:matricization} for the definition of matricization.
Then the estimate $\hA_1$ by NN-LIFT Algorithm~\ref{algo:NN-LIFT} satisfies w.h.p.\
$$\min_{z \in \{ \pm 1 \}} \|(A_1)_j - z \cdot (\hA_1)_j \| \leq \tl{O} \left( \tl{\epsilon}_1 \right), \quad j \in [k],$$
where the recovery guarantee is up to the permutation of columns of $A_1$.
\end{lemma}

\begin{remark}[Sign ambiguity] \label{remark:sign}
We observe that in addition to the permutation ambiguity in the recovery guarantees, there is also a sign ambiguity issue in recovering the columns of matrix $A_1$ through the decomposition of third order tensor in~\eqref{eqn:cross-moment}. This is because the sign of $(A_1)_j$ and coefficient $\lambda_j$ can both change while the overall tensor is still fixed. Note that the coefficient $\lambda_j$ can be positive or negative. According to the Fourier method for estimating $b_1$, mis-calculating the sign of $(A_1)_j$ also leads to sign of $b_1(j)$ recovered in the opposite manner. In other words, the recovered sign of the bias $b_1(j)$ is consistent with the recovered sign of $(A_1)_j$.

Recall we assume that the   nonlinear activating function $\sigma(z)$ satisfies the property such that $\sigma(z) = 1 - \sigma(-z)$. Many popular activating functions such as step function, sigmoid function and tanh function satisfy this property. Given this property, the sign ambiguity in parameters $A_1$ and $b_1$ which leads to opposite sign in input $z$ to the activating function $\sigma(\cdot)$ can be now compensated by the sign of $a_2$ and value of $b_2$, which is recovered through least squares.
\end{remark}

\bprfof{Lemma~\ref{lem:A1-guarantee}}
From Lemma~\ref{lem:moment}, we know that the exact cross-moment $\tl{T} = \E[\tl{y} \cdot \Sc_3(x)]$ has rank-one components as columns of matrix $A_1$; see Equation~\eqref{eqn:cross-moment} for the tensor decomposition form. 
We apply a  tensor decomposition method in NN-LIFT to estimate the columns of $A_1$. We employ noisy tensor decomposition guarantees in~\citet{anandkumar2014guaranteed}. They show that when the perturbation tensor is small, the tensor power iteration initialized by the SVD-based Procedure~\ref{algo:SVD init} recovers the rank-1 components up to some small error. We also analyze the whitening step and combine it with this result leading to Lemma~\ref{lem:perturbation-bound}.

Let us now characterize the perturbation matrix and tensor. By Lemma~\ref{lem:moment}, the CP decomposition form is given by $\tl{T} = \E[\tl{y} \cdot \Sc_3(x)]$, and thus, the perturbation tensor is written as
\begin{equation} \label{eqn:TensorPert-Est}
E := \tl{T} - \widehat{T} = \E[\tl{y} \cdot \Sc_3(x)] - \frac{1}{n} \sum_{i \in [n]} \tl{y}_i \cdot \Pc_3(x_i),
\end{equation}
where $\hT = \frac{1}{n} \sum_{i \in [n]} \tl{y}_i \cdot \Pc_3(x_i)$ is the empirical form used in NN-LIFT Algorithm~\ref{algo:NN-LIFT}. Notice that in the realizable setting, the neural network output $\tl{y}$ is observed and thus, it is used in forming  the empirical tensor. Similarly, the perturbation of second order moment $\tl{M}_2 = \E[\tl{y} \cdot \Sc_2(x)]$ is given by
\begin{equation} \label{eqn:TensorPert-Est2}
E_2 := \tl{M}_2 - \widehat{M}_2 = \E[\tl{y} \cdot \Sc_2(x)] - \frac{1}{n} \sum_{i \in [n]} \tl{y}_i \cdot \Pc_2(x_i).
\end{equation}

In order to bound $\|E\|$, we matricize it to apply matrix Bernstein's inequality. We have the matricized version as
$$\tilde{E} := \E[\tl{y} \cdot M_3(x)] - \frac{1}{n} \sum_{i \in [n]} \tl{y}_i \cdot M_3(x_i)
=  \sum_{i \in [n]} \frac{1}{n} \Bigl( \E[\tl{y} \cdot M_3(x)] -  \tl{y}_i \cdot M_3(x_i) \Bigr),$$
where $M_3(x) \in \R^{d\ \times d^2}$ is the matricization of $\Sc_3(x) \in \R^{d \times d \times d}$; see~\eqref{eqn:matricization} for the definition of matricization.
Now the norm of $\tilde{E}$ can be bounded by the matrix Bernstein's inequality. The norm of each (centered) random variable inside the summation is bounded as $\frac{\tl{y}_{\max}}{n} \E[\|M_3(x)\|]$, where $\tl{y}_{\max}$ is the bound on $|\tl{y}|$. The variance term is also bounded as
$$
\frac{1}{n^2} \Bigl\| \sum_{i \in [n]} \E \left[ \tl{y}_i^2 \cdot M_3(x_i) M_3^\top(x_i) \right] \Bigr\| \leq \frac{1}{n} \tl{y}_{\max}^2 \E \left[ \left\| M_3(x) M_3^\top(x) \right\| \right].
$$
Applying matrix Bernstein's inequality, we have w.h.p.\
\begin{equation} \label{eqn:TensorPert-Proof}
\|E\| \leq \|\tl{E}\| \leq \tl{O} \left( \frac{\tl{y}_{\max}}{\sqrt{n}} \sqrt{\E \left[ \left\| M_3(x) M_3^\top(x) \right\| \right]} \right).
\end{equation}
For the second order perturbation $E_2$, it is already a matrix, and by applying matrix Bernstein's inequality, we similarly argue that w.h.p.\
\begin{equation} \label{eqn:TensorPert-Proof2}
\|E_2\| \leq \tl{O} \left( \frac{\tl{y}_{\max}}{\sqrt{n}} \sqrt{\E \left[ \left\| \Sc_2(x) \Sc_2^\top(x) \right\| \right]} \right).
\end{equation}

There is one more remaining piece to complete the proof of tensor decomposition part. The analysis in~\citet{anandkumar2014guaranteed} does not involve any whitening step, and thus, we need to adapt the perturbation analysis of~\citet{anandkumar2014guaranteed} to our additional whitening procedure. This is done in Lemma~\ref{lem:perturbation-bound}. In the final recovery bound~\eqref{eqn:tensor-guarantees} in Lemma~\ref{lem:perturbation-bound}, there are two terms; one involving $\|E\|$, and the other involving $\|E_2\|$. We first impose a bound on sample complexity such that the bound involving $\|E\|$ dominates the bound involving $\|E_2\|$ as follows. Considering the bounds on $\|E\|$ and $\|E_2\|$ in~\eqref{eqn:TensorPert-Proof} and \eqref{eqn:TensorPert-Proof2}, and imposing the lower bound on the number of samples (third bound stated in the lemma) as
$$
n \geq
\tl{O} \left( \tl{y}_{\max}^2 \cdot \frac{\E \left[ \left\| \Sc_2(x) \Sc_2^\top(x) \right\| \right]^{3/2}}{\E \left[ \left\| M_3(x) M_3^\top(x) \right\| \right]^{1/2}} \cdot \frac{1}{\tl{\lambda}_{\min}^2 \cdot s_{\min}^3(A_1)} \right),
$$
leads to this goal. By doing this, we do not need to impose the bound on $\|E_2\|$ anymore, and applying the perturbation bound in~\eqref{eqn:TensorPert-Proof} to the required bound on $\|E\|$ in Lemma~\ref{lem:perturbation-bound} leads to sample complexity bound (second bound stated in the lemma)
$$
n \geq
\tl{O} \left( \tl{y}_{\max}^2 \cdot \E \left[ \left\| M_3(x) M_3^\top(x) \right\| \right] \cdot \left( \frac{\tl{\lambda}_{\max}}{\tl{\lambda}_{\min}} \right)^3 \frac{1}{\lambda_{\min}^2 \cdot s_{\min}^6(A_1)} \cdot k \right).
$$
Finally, applying the result of Lemma~\ref{lem:perturbation-bound}, we have the column-wise error guarantees (up to permutation)
$$
\|(A_1)_j - (\hA_1)_j \|
\leq \tl{O} \left( \frac{s_{\max}(A_1)}{\lambda_{\min} } \frac{\tl{\lambda}_{\max}^2}{\sqrt{\tl{\lambda}_{\min}}}  \frac{\tl{y}_{\max}}{ \tl{\lambda}_{\min}^{1.5} \cdot s_{\min}^3(A_1)} \frac{\sqrt{\E \left[ \left\| M_3(x) M_3^\top(x) \right\| \right]}}{\sqrt{n}} \right)
\leq \tl{O} \left( \tl{\epsilon}_1 \right),
$$
where in the first inequality we also substituted the bound on $\|E\|$ in~\eqref{eqn:TensorPert-Proof}, 
and the first bound on $n$ stated in the lemma is used in the last inequality.
\eprfof


\subsubsection{Whitening analysis} \label{appendix:whitening}

The perturbation analysis of proposed tensor decomposition method in Algorithm~\ref{alg:robustpower} with the corresponding SVD-based initialization in Procedure~\ref{algo:SVD init} is provided in~\citet{anandkumar2014guaranteed}. But, they do not consider the effect of whitening proposed in Procedures~\ref{algo:whitening}~and~\ref{algo:Un-whitening}. Thus, we need to adapt the perturbation analysis of~\citet{anandkumar2014guaranteed} when the whitening procedure is incorporated. We perform it in this section.

We first elaborate on the whitening step, and analyze how the proposed Procedure~\ref{algo:whitening} works. We then analyze the inversion of whitening operator showing how the components in the whitened space are translated back to the original space as stated in Procedure~\ref{algo:Un-whitening}. We finally provide the perturbation analysis of whitening step when estimations of moments are given.

\paragraph{Whitening procedure:}
Consider second order moment $\tl{M}_2$ which is used to whiten third order tensor
\begin{equation} \label{eqn:third-moment}
\tl{T} = \sum_{j \in [k]} \lambda_j \cdot (A_1)_j \otimes (A_1)_j \otimes (A_1)_j
\end{equation}
in Procedure~\ref{algo:whitening}. It is constructed such that it has the same decomposition form as target tensor $\tl{T}$, i.e., we have
\begin{equation} \label{eqn:second-moment}
\tl{M}_2 = \sum_{j \in [k]} \tl{\lambda}_j \cdot (A_1)_j \otimes (A_1)_j.
\end{equation}
We propose two options for constructing $\tl{M}_2$ in Procedure~\ref{algo:whitening}. First option is to use second order score function and construct $\tl{M}_2 := \Ebb \left[ \tl{y} \cdot \Pc_2(x) \right]$ for which we have
\begin{equation} \label{eqn:second-moment-score}
\tl{M}_2 := \Ebb \left[ \tl{y} \cdot \Pc_2(x) \right] = \sum_{j \in [k]} \tl{\lambda}_j \cdot (A_1)_j \otimes (A_1)_j,
\end{equation}
where
\begin{equation} \label{eqn:cross-moment2-coeffs}
\tl{\lambda}_j = \E \left[ \sigma''(z_j) \right] \cdot a_2(j),
\end{equation}
for vector $z := A_1^\top x+b_1$ as the input to the nonlinear operator $\sigma(\cdot)$. This is proved similar to Lemma~\ref{lem:moment}. Second option leads to the same form for $\tl{M}_2$ as~\eqref{eqn:second-moment} with coefficient modified as $\tl{\lambda}_j = \lambda_j \cdot \inner{(A_1)_j,\theta}$.

Let matrix $W \in \R^{d \times k}$ denote the whitening matrix in the noiseless case, i.e.,
the whitening matrix $W$ in Procedure~\ref{algo:whitening} is constructed such that $W^\top \tl{M}_2 W = I_k$. Applying whitening matrix $W$ to the noiseless tensor $\tl{T} = \sum_{j \in [k]} \lambda_j \cdot (A_1)_j \otimes (A_1)_j \otimes (A_1)_j$, we have
\begin{equation} \label{eqn:tensor-whitening-noiseless}
\tl{T}(W,W,W)
= \sum_{j \in [k]} \lambda_j \left( W^\top (A_1)_j \right)^{\otimes 3}
= \sum_{j \in [k]} \frac{\lambda_j}{\tl{\lambda}_j^{3/2}} \left( W^\top (A_1)_j \sqrt{\tl{\lambda}_j} \right)^{\otimes 3}
= \sum_{j \in [k]} \mu_j v_j^{\otimes 3},
\end{equation}
where 
we define
\begin{equation} \label{eqn:mu}
\mu_j := \frac{\lambda_j}{\tl{\lambda}_j^{3/2}}, \quad v_j := W^\top (A_1)_j \sqrt{\tl{\lambda}_j}, \quad j \in [k],
\end{equation}
in the last equality.
Let  $V := [v_1 \ v_2 \ \dotsb \ v_k] \in \R^{k \times k}$ denote the factor matrix for $\tl{T}(W,W,W)$. We have
\begin{equation} \label{eqn:V}
V := W^\top A_1 \Diag(\tl{\lambda}^{1/2}),
\end{equation}
and thus,
$$
V V^\top = W^\top A_1 \Diag(\tl{\lambda}) A_1^\top W = W^\top \tl{M}_2 W =   I_k.
$$
Since $V$ is a square matrix, it is also concluded that $V^\top V = I_k$, and therefore, tensor $\tl{T}(W,W,W)$ is whitened such that the rank-1 components $v_j$'s form an orthonormal basis. This discussion clarifies how the whitening procedure works.



\paragraph{Inversion of the whitening procedure:}
Let us also analyze the inversion procedure on how to transform $v_j$'s to $(A_1)_j$'s. The main step is stated in Procedure~\ref{algo:Un-whitening}. According to whitening Procedure~\ref{algo:whitening}, let $\tl{M}_2=U \Diag(\gamma) U^\top$, $U \in \R^{d \times k}$, $\gamma \in \R^k$, denote the rank-k SVD of $\tl{M}_2$. Substituting whitening matrix $W:= U \Diag(\gamma^{-1/2})$ in~\eqref{eqn:V}, and multiplying $U \Diag(\gamma^{1/2})$ from left, we have
$$
U \Diag(\gamma^{1/2}) V = U U^\top A_1 \Diag(\tl{\lambda}^{1/2}).
$$
Since the column spans of $A_1 \in \R^{d \times k}$ and $U \in \R^{d \times k}$ are the same (given their relations to $\tl{M}_2$), $A_1$ is a fixed point for the projection operator on the subspace spanned by the columns of $U$. This projector operator is $UU^\top$ (since columns of $U$ form an orthonormal basis), and therefore, $UU^\top A_1 = A_1$. Applying this to the above equation, we have
$$
A_1 = U \Diag(\gamma^{1/2}) V \Diag(\tl{\lambda}^{-1/2}),
$$
i.e.,
\begin{equation} \label{eqn:un-whitening}
(A_1)_j = \frac{1}{\sqrt{\tl{\lambda}_j}} U \Diag(\gamma^{1/2}) v_j, \quad j \in [k].
\end{equation}

The above discussions describe the details of whitening and unwhitening procedures. We now provide the guarantees of tensor decomposition given noisy versions of moments $\tl{M}_2$ and $\tl{T}$.

\begin{lemma} \label{lem:perturbation-bound}
Let $\hM_2 = \tl{M}_2 - E_2$ and $\hT = \tl{T} - E$ respectively denote the noisy versions of
\begin{equation}
\tl{M}_2 = \sum_{j \in [k]} \tl{\lambda}_j \cdot (A_1)_j \otimes (A_1)_j, \quad
\tl{T} = \sum_{j \in [k]} \lambda_j \cdot (A_1)_j \otimes (A_1)_j \otimes (A_1)_j.
\end{equation}
Assume the second and third order perturbations satisfy the bounds
\begin{align*}
\|E_2\| & \leq \tl{O}\left( \lambda_{\min}^{1/3}  \frac{\tl{\lambda}_{\min}^{7/6} }{\sqrt{\tl{\lambda}_{\max}}}  s_{\min}^2(A_1) \frac{1}{k^{1/6}} \right), \\
\|E\| & \leq \tl{O}\left( \lambda_{\min}  \left( \frac{\tl{\lambda}_{\min}}{\tl{\lambda}_{\max}}\right)^{1.5}  s_{\min}^3(A_1) \frac{1}{\sqrt{k}} \right).
\end{align*}
Then, the proposed tensor decomposition algorithm recovers estimations of rank-1 components $(A_1)_j$'s satisfying error
\begin{equation} \label{eqn:tensor-guarantees}
\|(A_1)_j - (\hA_1)_j\|
 \leq \tl{O} \left( \frac{s_{\max}(A_1)}{\lambda_{\min} } \cdot \frac{\tl{\lambda}_{\max}^2}{\sqrt{\tl{\lambda}_{\min}}} \cdot \left[ \frac{\|E_2\|^3}{\tl{\lambda}_{\min}^{3.5} \cdot s_{\min}^6(A_1)} + \frac{\|E\|}{\tl{\lambda}_{\min}^{1.5} \cdot s_{\min}^3(A_1)} \right] \right), \quad j \in [k].
\end{equation}
\end{lemma}

\bprf
We do not have access to the true matrix $\tl{M}_2$ and the true tensor $\tl{T}$, and the perturbed versions $\hM_2 = \tl{M}_2 - E_2$ and $\hT = \tl{T} - E$ are used in the whitening procedure. Here, $E_2 \in \R^{d \times d}$ denotes the perturbation matrix, and $E \in \R^{d \times d \times d}$ denotes the perturbation tensor.
Similar to the noiseless case, let $\hW \in \R^{d \times k}$ denotes the whitening matrix constructed by Procedure~\ref{algo:whitening} such that $\hW^\top \hM_2 \hW = I_k$, and thus it orthogonalizes the noisy matrix $\hM_2$.
Applying the whitening matrix $\hW$ to the tensor $\hT$, we have
\begin{align}
\hT(\hW,\hW,\hW) &= \tl{T}(W,W,W) - \tl{T}(W-\hW,W-\hW,W-\hW) - E(\hW,\hW,\hW) \nn \\
&= \sum_{j \in [k]} \mu_j v_j^{\otimes 3} - E_W, \label{eqn:tensor-whitening}
\end{align}
where we used Equation~\eqref{eqn:tensor-whitening-noiseless}, and we defined
\begin{equation} \label{eqn:perturbation-whitening}
E_W := \tl{T}(W-\hW,W-\hW,W-\hW) + E(\hW,\hW,\hW)
\end{equation}
as the perturbation tensor after whitening. Note that the perturbation is from two sources; one is from the error in computing whitening matrix reflected in $W-\hW$, and the other is the error in tensor $\hT$ reflected in $E$.

We know that the rank-1 components $v_j$'s form an orthonormal basis, and thus, we have a noisy orthogonal tensor decomposition problem in~\eqref{eqn:tensor-whitening}. We apply the result of \citet{anandkumar2014guaranteed} where they show that if
$$\|E_W\| \leq \frac{\mu_{\min} \sqrt{\log k}}{\alpha_0 \sqrt{k}},$$
for some constant $\alpha_0 >1$, then the tensor power iteration (applied to the whitened tensor) recovers the tensor rank-1 components with bounded error (up to the permutation of columns)
\begin{equation} \label{eqn:v-error}
\|v_j - \hv_j\| \leq \tl{O} \left( \frac{\|E_W\|}{\mu_{\min}} \right).
\end{equation}

We now relate the norm of $E_W$ to the norm of original perturbations $E$ and $E_2$. For the first term in~\eqref{eqn:perturbation-whitening}, from Lemmata 4 and 5 of~\citet{SongEtal:NonparametricTensorDecomp}, we have
$$
\| \tl{T}(W-\hW,W-\hW,W-\hW) \| \leq \frac{64 \|E_2\|^3}{\tilde{\lambda}_{\min}^{3.5} \cdot s_{\min}^6(A_1)}.
$$
For the second term, by the sub-multiplicative property we have
$$
\|E(\hW,\hW,\hW)\| \leq \|E\| \cdot \|\hW\|^3
\leq 8 \|E\| \cdot \|W\|^3
\leq \frac{8 \|E\|}{s_{\min}^3(A_1) \tl{\lambda}_{\min}^{3/2}}.
$$
Here in the last inequality, we used
$$\|W\| = \frac{1}{\sqrt{s_{k} (\tl{M}_2)}} \leq \frac{1}{s_{\min}(A_1) \sqrt{\tl{\lambda}_{\min}}},$$
where $s_k(\tl{M}_2)$ denotes the $k$-th largest singular value of $\tl{M}_2$.
Here, the equality is from the definition of $W$ based on rank-$k$ SVD of $\tl{M}_2$ in Procedure~\ref{algo:whitening}, and the inequality is from $\tl{M}_2 = A_1 \Diag(\tl{\lambda}) A_1^\top$.

Substituting these bounds, we finally need the condition
$$
\frac{64 \|E_2\|^3}{\tl{\lambda}_{\min}^{3.5} \cdot s_{\min}^6(A_1)} + \frac{8 \|E\|}{\tl{\lambda}_{\min}^{1.5} \cdot s_{\min}^3(A_1)} \leq \frac{\lambda_{\min} \sqrt{\log k}}{\alpha_0 \tl{\lambda}_{\max}^{1.5} \sqrt{k}},
$$
where we also substituted bound $\mu_{\min} \geq \lambda_{\min} / \tl{\lambda}_{\max}^{1.5}$, given Equation~\eqref{eqn:mu}.
The bounds stated in the lemma ensures that each of the terms on the left hand side of the inequality are bounded by the right hand side. Thus, by the result of~\citet{anandkumar2014guaranteed}, we have
$\|v_j - \hv_j\| \leq \tl{O} \left( \|E_W\|/\mu_{\min} \right)$.
On the other hand, by the unwhitening relationship in~\eqref{eqn:un-whitening}, we have
\begin{equation} \label{eqn:error-transform}
\|(A_1)_j - (\hA_1)_j\|
= \frac{1}{\sqrt{\tl{\lambda}_j}} \|\Diag(\gamma^{1/2}) \cdot [v_j - \hv_j] \|
 \leq \sqrt{\frac{\gamma_{\max}}{\tl{\lambda}_{\min}}} \cdot \|v_j - \hv_j\|
 \leq s_{\max}(A_1) \cdot \sqrt{\frac{\tl{\lambda}_{\max}}{\tl{\lambda}_{\min}}} \cdot \|v_j - \hv_j\|.
\end{equation}
where in the equality, we use the fact that orthonormal matrix $U$ preserves the $\ell_2$ norm, and the sub-multiplicative property is exploited in the first inequality. The last inequality is also from $\gamma_{\max} = s_{\max}(\tl{M}_2) \leq s_{\max}^2(A_1) \cdot \tl{\lambda}_{\max}$, which is from $\tl{M}_2 = A_1 \Diag(\tl{\lambda}) A_1^\top$.
Incorporating the error bound on $\|v_j - \hv_j\|$ in~\eqref{eqn:v-error}, we have
$$
\|(A_1)_j - (\hA_1)_j\|
\leq \tl{O} \left( s_{\max}(A_1) \cdot \sqrt{\frac{\tl{\lambda}_{\max}}{\tl{\lambda}_{\min}}} \cdot \frac{\|E_W\|}{\mu_{\min}} \right)
 \leq \tl{O} \left( \frac{s_{\max}(A_1)}{\lambda_{\min} } \cdot \frac{\tl{\lambda}_{\max}^2}{\sqrt{\tl{\lambda}_{\min}}} \cdot  \|E_W\| \right),
$$
where we used the bound $\mu_{\min} \geq \lambda_{\min}/\tl{\lambda}_{\max}^{1.5}$ in the last step.
\eprf

\subsection{Fourier analysis guarantees} \label{appendix:Fourier}

The analysis of Fourier method for estimating parameter $b_1$ includes the following two lemmas. In the first lemma, we argue the mean of random variable $v$ introduced in Algorithm~\ref{algo:NN-LIFT} in the  realizable setting. This clarifies why the phase of $v$ is related to unknown parameter $b_1$. In the second lemma, we argue the concentration of $v$ around its mean leading to the sample complexity result. Note that $v$ is denoted by $\tl{v}$ in the realizable setting.

The Fourier method can be also used to estimate the weight vector $a_2$ since it appears in the magnitude of complex number $v$. In this section, we also provide the analysis of estimating $a_2$ with Fourier method which can be used as an alternative, while we primarily estimate $a_2$ by the ridge regression analyzed in Appendix~\ref{appendix:LLSQ}.

\restate{lem:v-mean}

\bprf
Let $\tl{F}(\omega)$ denote the Fourier transform of label function $\tl{f}(x) := \E[\tl{y}|x] = \inner{a_2, \sigma(A_1^\top x + b_1)}$ which is~\citep{MarksArabshahi1994}
\begin{equation} \label{eqn:Fourier-NN}
\tl{F}(\omega) = \sum_{j \in [k]} \frac{a_2(j)}{|A_1(d,j)|} \Sigma \left( \frac{\omega_d}{A_1(d,j)} \right) e^{j 2 \pi b_1(j) \frac{\omega_d}{A_1(d,j)}} \delta \left( \omega_{-} - \frac{\omega_d}{A_1(d,j)} A_1(\setminus d,j) \right),
\end{equation}
where $\Sigma(\cdot)$ is the Fourier transform of $\sigma(\cdot)$, $u_{-}^\top = [u_1,u_2,\dotsc,u_{d-1}]$ is vector $u^\top$ with the last entry removed, $A_1(\setminus d,j) \in \R^{d-1}$ is the $j$-th column of matrix $A_1$ with the $d$-th (last) entry removed, and finally $\delta(u) = \delta(u_1) \delta(u_2) \dotsb \delta(u_d)$.

Let $p(\omega)$ denote the probability density function of frequency $\omega$.
We have
\begin{align*}
\E[\tl{v}] &= \E_{x,\tl{y},\omega} \left[ \frac{\tl{y}}{p(x)} e^{-j \inner{\omega, x}} \right] \\
&= \E_{x,\omega} \left[ \E_{\tl{y}|\{x,\omega\}} \left[ \frac{\tl{y}}{p(x)} e^{-j \inner{\omega, x}} \Big| x,\omega \right] \right] \\
&= \E_{x,\omega} \left[ \frac{\tl{f}(x)}{p(x)} e^{-j \inner{\omega, x}} \right] \\
&= \int_{\Omega_l} \int \tl{f}(x) e^{-j \inner{\omega, x}} p(\omega) dx d\omega \\
&= \int_{\Omega_l} \tl{F}(\omega) p(\omega) d\omega,
\end{align*}
where the second equality uses the law of total expectation, the third equality exploits the label-generating function definition $\tl{f}(x) := \E[\tl{y}|x]$, and the final equality is from the definition of Fourier transform.
The variable $\omega \in \R^d$ is drawn from a $d-1$ dimensional manifold $\Omega_l \subset \R^d$. In order to compute the above integral, we define $d$ dimensional set
$$\Omega_{l;\nu} := \left\{ \omega \in \R^d : \frac{1}{2} - \frac{\nu}{2} \leq \|\omega\| \leq \frac{1}{2} + \frac{\nu}{2},  \bigl|  \inner{\omega,(\hA_1)_l} \bigr| \geq \frac{1-\tl{\epsilon}_1^2/2}{2} \right\},$$
for which $\Omega_l = \lim_{\nu \rightarrow 0^+} \Omega_{l;\nu}$. Assuming $\omega$'s are uniformly drawn from $\Omega_{l;\nu}$, we have
\begin{align*}
\E[\tl{v}] &= \lim_{\nu \rightarrow 0^+} \int_{\Omega_l;\nu} \tl{F}(\omega) p(\omega) d\omega \\
&= \lim_{\nu \rightarrow 0^+} \frac{1}{|\Omega_{l;\nu}|} \int_{-\infty}^{+\infty} \tl{F}(\omega) 1_{\Omega_{l;\nu}}(\omega) d\omega.
\end{align*}
The second equality is from uniform draws of $\omega$ from set $\Omega_{l;\nu}$ such that $p(\omega) = \frac{1}{|\Omega_{l;\nu}|} 1_{\Omega_{l;\nu}}(\omega)$, where $1_{S}(\cdot)$ denotes the indicator function for set $S$. Here, $|\Omega_{l;\nu}|$ denotes the volume of $d$ dimensional subspace $\Omega_{l;\nu}$, for which in the limit $\nu \rightarrow 0^+$, we have $|\Omega_{l;\nu}| = \nu \cdot |\Omega_l|$, where $|\Omega_l|$ denotes the surface area of $d-1$ dimensional manifold $\Omega_l$.

For small enough $\tl{\epsilon}_1$ in the definition of $\Omega_{l;\nu}$, only the delta function for $j=l$ in the expansion of $\tl{F}(\omega)$ in~\eqref{eqn:Fourier-NN} is survived from the above integral, and thus,
$$
\E[\tl{v}] = \lim_{\nu \rightarrow 0^+} \frac{1}{|\Omega_{l;\nu}|} \int_{- \infty}^{+ \infty} \frac{a_2(l)}{|A_1(d,l)|} \Sigma \left( \frac{\omega_d}{A_1(d,l)} \right) e^{j 2 \pi b_1(l) \frac{\omega_d}{A_1(d,l)}} \delta \left( \omega_{-} - \frac{\omega_d}{A_1(d,l)} A_1(\setminus d,l) \right) 1_{\Omega_{l;\nu}}(\omega) d \omega.
$$
In order to simplify the notations, in the rest of the proof we denote $l$-th column of matrix $A_1$ by vector $\alpha$, i.e., $\alpha := (A_1)_l$. Thus, the goal is to compute the integral
$$
I := \int_{- \infty}^{+ \infty} \frac{1}{|\alpha_d|} \Sigma \left( \frac{\omega_d}{\alpha_d} \right) e^{j 2 \pi b_1(l) \frac{\omega_d}{\alpha_d}} \delta \left( \omega_{-} - \frac{\omega_d}{\alpha_d} \alpha_- \right) 1_{\Omega_{l;\nu}}(\omega) d \omega,
$$
and note that $\E[\tl{v}] = a_2(l) \cdot \lim_{\nu \rightarrow 0^+} \frac{I}{|\Omega_{l;\nu}|}$. The rest of the proof is about computing the above integral. The integral involves delta functions where the final value is expected to be computed at a single point specified by the intersection of line $\omega_{-} = \frac{\omega_d}{\alpha_d} \alpha_-$, and sphere $\|\omega\|=\frac{1}{2}$ (when we consider the limit $\nu \rightarrow 0^+$). This is based on the following integration property of delta functions such that for function $g(\cdot):\R \rightarrow \R$,
\begin{equation} \label{eqn:delta-intprop}
\int_{- \infty}^{+ \infty} g(t) \delta(t) dt = g(0).
\end{equation}
We first expand the delta function as follows.
\begin{align*}
I &= \int_{- \infty}^{+ \infty} \frac{1}{|\alpha_d|} \Sigma \left( \frac{\omega_d}{\alpha_d} \right) e^{j 2 \pi b_1(l) \frac{\omega_d}{\alpha_d}} \delta \left( \omega_1 - \frac{\alpha_1}{\alpha_d} \omega_d \right) \dotsb \delta \left( \omega_{d-1} - \frac{\alpha_{d-1}}{\alpha_d} \omega_d \right) 1_{\Omega_{l;\nu}}(\omega) d \omega, \\
&= \int \dotsb \int_{- \infty}^{+ \infty} \Sigma \left( \frac{\omega_d}{\alpha_d} \right) e^{j 2 \pi b_1(l) \frac{\omega_d}{\alpha_d}} \delta \left( \omega_1 - \frac{\alpha_1}{\alpha_d} \omega_d \right) \dotsb \delta \left( \omega_{d-2} - \frac{\alpha_{d-2}}{\alpha_d} \omega_d \right) \\
& \hspace{3in} 1_{\Omega_{l;\nu}}(\omega) \cdot
\delta \left(\alpha_d \omega_{d-1} - \alpha_{d-1} \omega_d \right) d \omega_1 \dotsb \omega_d,
\end{align*}
where we used the property $\frac{1}{|\beta|} \delta(t) = \delta(\beta t)$ in the second equality.
Introducing new variable $z$, and applying the change of variable $\omega_d = \frac{1}{\alpha_{d-1}} (\alpha_d \omega_{d-1} - z)$, we have
\begin{align*}
I &= \int \dotsb \int_{- \infty}^{+ \infty} \Sigma \left( \frac{\omega_d}{\alpha_d} \right) e^{j 2 \pi b_1(l) \frac{\omega_d}{\alpha_d}} \delta \left( \omega_1 - \frac{\alpha_1}{\alpha_d} \omega_d \right) \dotsb \delta \left( \omega_{d-2} - \frac{\alpha_{d-2}}{\alpha_d} \omega_d \right) \\
& \hspace{2in} 1_{\Omega_{l;\nu}}(\omega) \cdot \delta(z) d\omega_1 \dotsb d\omega_{d-1} \frac{dz}{\alpha_{d-1}}, \\
&= \int \dotsb \int_{- \infty}^{+ \infty} \frac{1}{\alpha_{d-1}} \Sigma \left( \frac{\omega_{d-1}}{\alpha_{d-1}} \right) e^{j 2 \pi b_1(l) \frac{\omega_{d-1}}{\alpha_{d-1}}} \delta \left( \omega_1 - \frac{\alpha_1}{\alpha_{d-1}} \omega_{d-1} \right) \dotsb \delta \left( \omega_{d-2} - \frac{\alpha_{d-2}}{\alpha_{d-1}} \omega_{d-1} \right) \\
& \hspace{2in} 1_{\Omega_{l;\nu}} \left( \left[ \omega_1, \omega_2, \dotsc, \omega_{d-1}, \frac{\alpha_d}{\alpha_{d-1}} \omega_{d-1} \right] \right) d\omega_1 \dotsb d\omega_{d-1}.
\end{align*}
For the sake of simplifying the mathematical notations, we did not substitute all the $\omega_d$'s with $z$ in the first equality, but note that all $\omega_d$'s are implicitly a function of $z$ which is finally considered in the second equality where the delta integration property in~\eqref{eqn:delta-intprop} is applied to variable $z$ (note that $z=0$ is the same as $\frac{\omega_d}{\alpha_d} = \frac{\omega_{d-1}}{\alpha_{d-1}}$). Repeating the above process several times, we finally have
\begin{equation*}
I = \int_{- \infty}^{+ \infty} \frac{1}{\alpha_1} \Sigma \left( \frac{\omega_1}{\alpha_1} \right) e^{j 2 \pi b_1(l) \frac{\omega_1}{\alpha_1}} \cdot 1_{\Omega_{l;\nu}} \left( \left[ \omega_1, \frac{\alpha_2}{\alpha_1} \omega_1, \dotsc, \frac{\alpha_{d-1}}{\alpha_1} \omega_1, \frac{\alpha_d}{\alpha_1} \omega_1 \right] \right) d\omega_1.
\end{equation*}
There is a line constraint as $\frac{\omega_1}{\alpha_1} = \frac{\omega_2}{\alpha_2} = \dotsb = \frac{\omega_d}{\alpha_d}$ in the argument of indicator function. This implies that $\|\omega\| = \frac{\| \alpha\|}{\alpha_1} \omega_1 = \frac{\omega_1}{\alpha_1}$, where we used $\|\alpha\| = \|(A_1)_l\|=1$. Incorporating this in the norm bound imposed by the definition of $\Omega_{l;\nu}$, we have $\frac{\alpha_1}{2} (1-\nu) \leq \omega_1 \leq \frac{\alpha_1}{2} (1+\nu)$, and hence,
\begin{equation*}
I = \int_{\frac{\alpha_1}{2} (1-\nu)}^{\frac{\alpha_1}{2} (1+\nu)} \frac{1}{\alpha_1} \Sigma \left( \frac{\omega_1}{\alpha_1} \right) e^{j 2 \pi b_1(l) \frac{\omega_1}{\alpha_1}} d\omega_1.
\end{equation*}
We know $\E[\tl{v}] = a_2(l) \cdot \lim_{\nu \rightarrow 0^+} \frac{I}{|\Omega_{l;\nu}|}$, and thus,
$$\E[\tl{v}] = a_2(l) \cdot \frac{1}{\nu \cdot |\Omega_l|} \cdot \alpha_1 \nu \frac{1}{\alpha_1} \Sigma \left( \frac{1}{2} \right) e^{j 2 \pi b_1(l) \frac{1}{2}}
= \frac{1}{|\Omega_l|} a_2(l) \Sigma \left( \frac{1}{2} \right) e^{j\pi b_1(l)},$$
where in the first step we use $|\Omega_{l;\nu}| = \nu \cdot |\Omega_l|$, and write the integral $I$ in the limit $\nu \rightarrow 0^+$. This finishes the proof.
\eprf

%

In the following lemma, we argue the concentration of $v$ around its mean which leads to the sample complexity bound for estimating the parameter $b_1$ (and also $a_2$) within the desired error.

\begin{lemma} \label{lem:Fourier-guarantee}
If the sample complexity
\begin{equation} \label{eqn:samplecomp-fourier}
n \geq O \left( \frac{\tl{\zeta}_{\tl{f}}}{\psi \tl{\epsilon}_2^2} \log \frac{k}{\delta} \right)
\end{equation}
holds for small enough $\tl{\epsilon}_2 \leq \tl{\zeta}_{\tl{f}}$,
then the estimates $\ha_2(l) = \frac{|\Omega_l|}{|\Sigma(1/2)|} |\tl{v}|$, and $\hb_1(l) = \frac{1}{ \pi} (\angle \tl{v} - \angle \Sigma(1/2))$ for $l \in [k]$, in NN-LIFT Algorithm~\ref{algo:NN-LIFT} (see the definition of $\tl{v}$ in \eqref{eqn:v-appendix}) satisfy with probability at least $1-\delta$,
$$|a_2(l) - \ha_2(l) | \leq \frac{|\Omega_l|}{|\Sigma(1/2)|} O(\tl{\epsilon}_2), \quad
|b_1(l) - \hb_1(l) | \leq \frac{|\Omega_l|}{\pi |\Sigma(1/2)| |a_2(l)|} O(\tl{\epsilon}_2).$$
\end{lemma}

\bprf
The result is proved by arguing the concentration of variable $\tl{v}$ in~\eqref{eqn:v-appendix} around its mean characterized in~\eqref{eqn:v-mean}. We use the Bernstein's inequality to do this. Let $\tl{v} := \sum_{i \in [n]} \tl{v}_i$ where  $\tl{v}_i = \frac{1}{n} \frac{\tl{y}_i}{p(x_i)} e^{-j \inner{\omega_i, x_i}}$.
By the lower bound $p(x) \geq \psi$ assumed in Theorem~\ref{thm:guarantees-sample}
 and labels $\tl{y}_i$'s being bounded, the magnitude of centered $\tl{v}_i$'s ($\tl{v}_i - \E[\tl{v}_i]$) are bounded by  $O(\frac{1}{\psi n})$. The variance term is also bounded as
\begin{align*}
\sigma^2 = \Bigl| \sum_{i \in [n]} \Ebb \left[ (\tl{v}_i - \E[\tl{v}_i])(\overline{\tl{v}_i - \E[\tl{v}_i]}) \right] \Bigr|,
\end{align*}
where $\overline{u}$ denotes the complex conjugate of complex number $u$. This is bounded as
$$
\sigma^2 \leq \sum_{i \in [n]} \E \left[ \tl{v}_i \overline{\tl{v}_i} \right]
= \frac{1}{n^2} \sum_{i \in [n]} \E \left[ \frac{\tl{y}_i^2}{p(x_i)^2} \right]
$$
Since output $\tl{y}$ is a binary label ($\tl{y} \in \{0,1\}$), we have $\E[\tl{y}^2|x] = \E[\tl{y}|x] = \tl{f}(x)$, and thus,
$$\E \left[ \frac{\tl{y}^2}{p(x)^2} \right] = \E \left[ \E \left[ \frac{\tl{y}^2}{p(x)^2}|x \right] \right] = \E \left[ \frac{\tl{f}(x)}{p(x)^2} \right] \leq \frac{1}{\psi} \int_{\R^d} \tl{f}(x) dx = \frac{\tl{\zeta}_{\tl{f}}}{\psi},$$
where the inequality uses the bound $p(x) \geq \psi$ and the last equality is from definition of $\tl{\zeta}_{\tl{f}}$.
This provides us the bound on variance as
$$\sigma^2 \leq \frac{\tl{\zeta}_{\tl{f}}}{\psi n}.$$
Applying Bernstein's inequality concludes the concentration bound such that with probability at least $1 - \delta$, we have
$$|\tl{v} - \E[\tl{v}]| \leq O \left( \frac{1}{\psi n} \log \frac{1}{\delta} + \sqrt{\frac{\tl{\zeta}_{\tl{f}}}{\psi n} \log \frac{1}{\delta}} \right) \leq O(\tl{\epsilon}_2),$$
where the last inequality is from sample complexity bound.
This implies that $||\tl{v}| - |\E[\tl{v}]|| \leq O(\tl{\epsilon}_2)$. Substituting $|\E[\tl{v}]|$ from~\eqref{eqn:v-mean} and considering estimate $\ha_2(l) = \frac{|\Omega_l|}{|\Sigma(1/2)|} |\tl{v}|$, we have
$$|\ha_2(l) - a_2(l)| \leq \frac{|\Omega_l|}{|\Sigma(1/2)|} O(\tl{\epsilon}_2),$$
which finishes the first part of the proof.
For the phase, we have
$\phi := \angle \tl{v} - \angle \E[\tl{v}] = \pi (\hb_1(l) - b_1(l)).$
On the other hand, for small enough error $\tl{\epsilon}_2$ (and thus small $\phi$), we have the approximation $\phi \sim \tan(\phi) \sim \frac{|\tl{v}-\E[\tl{v}]|}{|\E[\tl{v}]|}$ (note that this is actually an upper bound such that $\phi \leq \tan(\phi)$). Thus,
$$|\hb_1(l) - b_1(l)| \leq \frac{1}{\pi |\E[\tl{v}]|} O(\tl{\epsilon}_2) \leq \frac{|\Omega_l|}{\pi |\Sigma(1/2)| |a_2(l)|} O(\tl{\epsilon}_2).$$
This finishes the proof of second bound.
\eprf

\subsection{Ridge regression analysis and guarantees} \label{appendix:LLSQ}

Let $h := \sigma ( A_1^\top x + b_1)$ denote the neuron or hidden layer variable. With slightly abuse of notation, in the rest of analysis in this section, we append variable $h$ by the dummy variable 1 to represent the bias, and thus, $h \in \R^{k+1}$.
We write the output as $\tl{y} = h^\top \beta + \eta$, where
$$\beta := [a_2, b_2] \in \R^{k+1}.$$
Given the estimated parameters of first layer denoted by $\hA_1$ and $\hb_1$, the neurons are estimated as $\hh := \sigma ( \hA_1^\top x + \hb_1)$. In addition, the dummy variable 1 is also appended, and thus, $\hh \in \R^{k+1}$. Because of this estimated encoding of neurons, we expand the output $\tl{y}$ as
\begin{equation} \label{eqn:ridge-regression}
\tl{y} = \hh^\top \beta + \underbrace{(h^\top - \hh^\top ) \beta}_{\text{bias (approximation)}: \ b(\hh)} + \underbrace{\eta}_{\text{noise}} = \tl{f}(\hh) + \eta,
\end{equation}
where $\tl{f}(\hh) := \E[\tl{y}|\hh] = \hh^\top \beta + b(\hh)$.
Here, we have a noisy linear model with additional bias (approximation).
Let $\hat{\beta}_\lambda$ denote the ridge regression estimator for some regularization parameter $\lambda \geq 0$, which is defined as the minimizer of the regularized empirical mean squared error, i.e.,
$$
\hat{\beta}_\lambda := \argmin_\beta \frac{1}{n} \sum_{i \in [n]} \left( \inner{\beta,\hh_i} - \tl{y}_i \right)^2 + \lambda \|\beta\|^2.
$$
We know this estimator is given by (when $\hat{\Sigma}_{\hh} + \lambda I \succ 0$)
$$\hat{\beta}_\lambda = \left( \hat{\Sigma}_{\hh} + \lambda I \right)^{-1} \cdot \hat{\E}(\hh \tl{y}),$$
where $\hat{\Sigma}_{\hh} := \frac{1}{n} \sum_{i \in [n]}  \hh_i \hh_i^\top$ is the empirical covariance of $\hh$, and $\hat{\E}$ denotes the empirical mean operator. The analysis of ridge regression 
leads to the following expected prediction error (risk) bound on the estimation of the output.

\begin{lemma}[Expected prediction error of ridge regression]
Suppose the parameter recovery results in Lemmata~\ref{lem:A1-guarantee}~and~\ref{lem:Fourier-guarantee} on $A_1$ and  $b_1$ hold. In addition, assume the nonlinear activating function $\sigma(\cdot)$ satisfies the Lipschitz property  such that $|\sigma(u) - \sigma(u')| \leq L \cdot |u-u'|$, for $u,u' \in \R$. The following noise, approximation and statistical leverage conditions also hold.  Then, by choosing the optimal $\lambda>0$ in the $\lambda$-regularized ridge regression (which estimates the parameters $\ha_2$ and $\hb_2$), the  estimated output as $\hf(x) = \ha_2^\top \sigma(\hA_1^\top x + \hb_1) + \hb_2$ satisfies the risk bound
$$\E[|\hf(x) - \tl{f}(x)|^2] \leq O \left( \frac{k \|\beta\|^2}{n} \right) + O  \left( \sqrt{ \frac{2 k \|\beta\|^2}{n} \left( \E[b(\hh)^2] + \sigma_{\text{noise}}^2/2 \right)} \right) + \E[b(\hh)^2],$$
where
$$
\E[b(\hh)^2] \leq \left[ r + \frac{|\Omega_l|}{\pi |\Sigma(1/2)| |a_2(l)|} \right]^2 \|\beta\|^2 L^2 k O(\tl{\epsilon}^2).
$$
\end{lemma}

\bprf
Since $\hh := \sigma ( \hA_1^\top x + \hb_1)$, we equivalently argue the bound on $\E[(\hh^\top \hat{\beta}_\lambda - \tl{f}(\hh))^2]$, where $\hf(x) = \hf(\hh) = \hh^\top \hat{\beta}_\lambda$.
From standard results in the study of inverse problems, we know (see Proposition 5 in~\citet{hsu2014random})
$$
\E[(\hh^\top \hat{\beta}_\lambda - \tl{f}(\hh))^2] = \E[(\hh^\top \beta - \tl{f}(\hh))^2] + \| \hat{\beta}_\lambda - \beta \|_{\Sigma_{\hh}}^2.
$$
Here, for positive definite matrix $\Sigma \succ 0$, the vector norm $\|\cdot\|_\Sigma$ is defined as $\|v\|_\Sigma := \sqrt{v^\top \Sigma v}$.
For the first term, by the definition of $\tl{f}(\hh)$ as $\tl{f}(\hh) := \E[\tl{y}|\hh] = \hh^\top \beta + b(\hh)$, we have
$$
\E[(\hh^\top \beta - \tl{f}(\hh))^2] = \E[b(\hh)^2].
$$
Lemma~\ref{lem:LLSQ-bounded-approx} bounds $\E[b(\hh)^2]$ and bounding $\| \hat{\beta}_\lambda - \beta \|_{\Sigma_{\hh}}^2$ is argued in Lemma~\ref{lem:LLSQ-bounded-excess} and Remark~\ref{remark:optimal-lambda}. Combining these bounds finishes the proof.
\eprf

In order to have final risk bounded as $\E[|\hf(x) - \tl{f}(x)|^2] \leq \tl{O} (\epsilon^2)$, for some $\epsilon >0$, the above lemma imposes sample complexity as (some of other parameters considered in~\eqref{eqn:samplecomp-tensordecomp}, \eqref{eqn:samplecomp-fourier} are not repeated here)
\begin{equation} \label{eqn:samplecomp-ridge1}
n \geq \tl{O} \left( L \frac{k \|\beta\|^2}{\epsilon^2} (1 + \sigma_{\text{noise}}^2) \right).
\end{equation}

\begin{lemma}[Bounded approximation] \label{lem:LLSQ-bounded-approx}
Suppose the parameter recovery results in Lemmata~\ref{lem:A1-guarantee}~and~\ref{lem:Fourier-guarantee} on $A_1$ and  $b_1$ hold. In addition, assume the nonlinear activating function $\sigma(\cdot)$ satisfies the Lipschitz property  such that $|\sigma(u) - \sigma(u')| \leq L \cdot |u-u'|$, for $u,u' \in \R$.
Then, the approximation term is bounded as
$$
\E[b(\hh)^2] \leq \left[ r + \frac{|\Omega_l|}{\pi |\Sigma(1/2)| |a_2(l)|} \right]^2 \|\beta\|^2 L^2 k O(\tl{\epsilon}^2).
$$
\end{lemma}

\bprf
We have
\begin{equation} \label{eqn:LLSQ-term1}
\E[b(\hh)^2] = \E[\inner{h-\hh,\beta}^2] \leq \|\beta\|^2 \cdot \E[\|h-\hh\|^2].
\end{equation}
Define $\tl{\epsilon} := \max\{\tl{\epsilon}_1,\tl{\epsilon}_2 \}$, where $\tl{\epsilon}_1$ and $\tl{\epsilon}_2$ are the corresponding bounds in Lemmata~\ref{lem:A1-guarantee}~and~\ref{lem:Fourier-guarantee}, respectively.
Using the Lipschitz property of nonlinear function $\sigma(\cdot)$, we have
\begin{align*}
|h_l - \hh_l | &= |\sigma(\inner{(A_1)_l, x} + b_1(l)) - \sigma( \inner{(\hA_1)_l, x} + \hb_1(l))| \\
&\leq L \cdot \left[ |\inner{(A_1)_l - (\hA_1)_l, x}| + |b_1(l)-\hb_1(l)| \right] \\
&\leq L \cdot \left[ r O(\tl{\epsilon}) + \frac{|\Omega_l|}{\pi |\Sigma(1/2)| |a_2(l)|} O(\tl{\epsilon}) \right],
\end{align*}
where in the second inequality, we use the bounds in Lemmata~\ref{lem:A1-guarantee}~and~\ref{lem:Fourier-guarantee}, and bounded $x$ such that $\|x\| \leq r$. Applying this to~\eqref{eqn:LLSQ-term1} concludes the proof.
\eprf


We now assume the following additional conditions to bound $\|\hat{\beta}_\lambda - \beta \|_{\Sigma_{\hh}}^2$. The following discussions are along the results of \citet{hsu2014random}.

We define the effective dimensions of the covariate $\hh$ as
$$
k_{p,\lambda} := \sum_{j \in [k]} \left( \frac{\lambda_j}{\lambda_j + \lambda} \right)^p, \quad p \in \{1,2\},
$$
where $\lambda_j$'s denote the (positive) eigenvalues of $\Sigma_{\hh}$, and $\lambda$ is the regularization parameter of ridge regression.


\begin{itemize}
\item Subgaussian noise: there exists a finite $\sigma_{\text{noise}} \geq 0$ such that, almost surely,
$$
\E_{\eta}[\exp(\alpha \eta)| \hh] \leq \exp(\alpha^2 \sigma_{\text{noise}}^2/2), \quad \forall \alpha \in \R,
$$
where $\eta$ denotes the noise in the output $\tl{y}$.
\item Bounded statistical leverage: there exists a finite $\rho_\lambda \geq 1$ such that, almost surely,
$$\frac{\sqrt{k}}{\sqrt{(\inf\{\lambda_j\} + \lambda)k_{1,\lambda}}} \leq \rho_\lambda.$$
\item Bounded approximation error at $\lambda$: there exists a finite $B_{\text{bias},\lambda} \geq 0$ such that, almost surely,
$$\rho_\lambda \left( B_{\max} + \sqrt{k} \|\beta\| \right) \leq B_{\text{bias},\lambda},$$
where $|b(\hh)| \leq B_{\max}$. Note that the approximation term $b(\hh)$ is bounded in Lemma~\ref{lem:LLSQ-bounded-approx}. The parameter $B_{\text{bias},\lambda}$ only contributes to the lower order terms in the analysis of ridge regression.
\end{itemize}

\begin{lemma}[Bounding excess mean squared error: Theorem 2 of \citet{hsu2014random}] \label{lem:LLSQ-bounded-excess}
Fix some $\lambda \geq 0$, and suppose the above noise, approximation and statistical leverage conditions hold, and in addition,
\begin{equation} \label{eqn:samplecomp-ridge2}
n \geq \tl{O} (\rho_\lambda^2 k_{1,\lambda}).
\end{equation}
Then, we have
$$
\| \hat{\beta}_\lambda - \beta \|_{\Sigma_{\hh}}^2 \leq \tl{O} \left( \frac{k}{\lambda n} \left( \E[b(\hh)^2] + \sigma_{\text{noise}}^2/2 \right) + \frac{\lambda \|\beta\|^2}{2} \left( 1 + \frac{k/\lambda + 1}{n} \right) \right) + o(1/n),
$$
where $\E[b(\hh)^2]$ is bounded in Lemma~\ref{lem:LLSQ-bounded-approx}.
\end{lemma}

In the above lemma, we also used the discussions in Remarks 12 and 15 of \citet{hsu2014random} which include comments on the simplification of the general result.

\begin{remark}[Optimal $\lambda$] \label{remark:optimal-lambda}
In addition, along the discussion in Remark 15 of~\citet{hsu2014random}, by choosing the optimal $\lambda > 0$ that minimizes the bound in the above lemma, we have
$$
\| \hat{\beta}_\lambda - \beta \|_{\Sigma_{\hh}}^2 \leq O \left( \frac{k \|\beta\|^2}{n} \right) + O  \left( \sqrt{ \frac{2 k \|\beta\|^2}{n} \left( \E[b(\hh)^2] + \sigma_{\text{noise}}^2/2 \right)} \right).
$$
\end{remark}

\section{Proof of Theorem~\ref{thm:approx-guarantees}} \label{appendix:proof-risk}

Before we provide the proof, we first state the details of bound on $C_f$. We require
\begin{align} \label{eqn:Cf-bound-tensordecomp}
C_f
 \leq \min & \left\{ \tl{O} \left( \frac{1}{r} \left( \frac{1}{\sqrt{k}} + \delta_1 \right)^{-1} \frac{1}{\sqrt{\E [ \left\| \Sc_3(x) \right\|^2 ]}} \cdot \frac{\tl{\lambda}_{\min}^2}{\tl{\lambda}_{\max}^2} \cdot \lambda_{\min} \cdot \frac{s_{\min}^3(A_1)}{s_{\max}(A_1)} \cdot \tl{\epsilon}_1 \right), \right. \\
& \quad \tl{O}\left( \frac{1}{r} \left( \frac{1}{\sqrt{k}} + \delta_1 \right)^{-1} \frac{1}{\sqrt{\E [ \left\| \Sc_3(x) \right\|^2 ]}} \cdot \lambda_{\min}  \left( \frac{\tl{\lambda}_{\min}}{\tl{\lambda}_{\max}}\right)^{1.5}  s_{\min}^3(A_1) \cdot \frac{1}{\sqrt{k}} \right), \nn \\
& \quad \left. O \left( \frac{1}{r} \left( \frac{1}{\sqrt{k}} + \delta_1 \right)^{-1} \frac{\E [ \left\| \Sc_3(x) \right\|^2 ]^{1/4}}{\E [ \left\| \Sc_2(x) \right\|^2 ]^{3/4}} \cdot \tl{\lambda}_{\min} \cdot s_{\min}^{1.5}(A_1) \right) \right\}. \nn
\end{align}

\bprfof{Theorem~\ref{thm:approx-guarantees}}
We first argue that the perturbation involves both estimation and approximation parts.
\paragraph{Perturbation decomposition into approximation and estimation parts:}
Similar to the estimation part analysis, we need to ensure the perturbation from exact means is small enough to  apply the analysis of Lemmas~\ref{lem:A1-guarantee}~and~\ref{lem:Fourier-guarantee}. Here, in addition to the empirical estimation of quantities (estimation error), the approximation error also contributes to the perturbation. This is because there is no realizable setting here, and the observations are from an arbitrary function $f(x)$. We address this for both the tensor decomposition and the Fourier parts as follows.

Recall that we use notation $\tl{f}(x)$ (and $\tl{y}$) to denote the output of a neural network. For arbitrary function $f(x)$, we refer to the neural network satisfying the approximation error provided in Theorem~\ref{thm:approx-Barron} by $\tl{y}_f$. The ultimate goal of our analysis is to show that NN-LIFT recovers the parameters of this specific neural network with small error. More precisely, note that these are a class of neural networks satisfying the approximation bound in Theorem~\ref{thm:approx-guarantees}, and it suffices to say that the output of the algorithm is close enough to one of them.


\paragraph{Tensor decomposition:}
There are two perturbation sources in the tensor analysis. One is from the approximation part and the other is from the estimation part.
By Lemma~\ref{lem:moment}, the CP decomposition form is given by $\tl{T}_f = \E[\tl{y}_f \cdot \Sc_3(x)]$, and thus, the perturbation tensor is written as
$$
E := \tl{T}_f - \widehat{T} = \E[\tl{y}_f \cdot \Sc_3(x)] - \frac{1}{n} \sum_{i \in [n]} y_i \cdot \Pc_3(x_i),
$$
where $\hT = \frac{1}{n} \sum_{i \in [n]} y_i \cdot \Pc_3(x_i)$ is the empirical form used in NN-LIFT Algorithm~\ref{algo:NN-LIFT}. Note that the observations are from the arbitrary function $y=f(x)$. The perturbation tensor can be expanded as
$$
E = \underbrace{\E[\tl{y}_f \cdot \Sc_3(x)] - \E[y \cdot \Sc_3(x)]}_{:= E_{\text{apx.}}} + \underbrace{\E[y \cdot \Sc_3(x)] - \frac{1}{n} \sum_{i \in [n]} y_i \cdot \Pc_3(x_i)}_{:= E_{\text{est.}}},
$$
where $E_{\text{apx.}}$ and $E_{\text{est.}}$ respectively denote the perturbations from approximation and estimation parts. 

We also desire to use the exact second order moment $\tl{M}_{2,f} = \E[\tl{y}_f \cdot \Sc_2(x)]$ for the whitening Procedure~\ref{algo:whitening} in the tensor decomposition method. But, we have an empirical version for which the perturbation matrix $E_2 := \tl{M}_{2,f} - \widehat{M}_2$ is expanded as
$$
E_2 = \underbrace{\E[\tl{y}_f \cdot \Sc_2(x)] - \E[y \cdot \Sc_2(x)]}_{:= E_{2,\text{apx.}}} + \underbrace{\E[y \cdot \Sc_2(x)] - \frac{1}{n} \sum_{i \in [n]} y_i \cdot \Pc_2(x_i)}_{:= E_{2,\text{est.}}},
$$
where $E_{2,\text{apx.}}$ and $E_{2,\text{est.}}$ respectively denote the perturbations from approximation and estimation parts.

In Theorem~\ref{thm:guarantees-sample} where there is no approximation error, we only need to analyze the estimation perturbations characterized in~\eqref{eqn:TensorPert-Est} and \eqref{eqn:TensorPert-Est2} since the neural network output is directly observed (and thus, we use $\tl{y}$ to denote the output).
Now, the goal is to argue that the norm of perturbations $E$ and $E_2$ are small enough (see Lemma~\ref{lem:perturbation-bound}), ensuring the tensor power iteration recovers the rank-1 components of $\tl{T}_f = \E[\tl{y}_f \cdot \Sc_3(x)]$ with bounded error. Again recall from Lemma~\ref{lem:moment} that the rank-1 components of tensor $\tl{T}_f = \E[\tl{y}_f \cdot \Sc_3(x)]$ are the desired components to recover.

The estimation perturbations $E_{\text{est.}}$ and $E_{2,\text{est.}}$ are similarly bounded as in Lemma~\ref{lem:A1-guarantee} (see~\eqref{eqn:TensorPert-Proof} and \eqref{eqn:TensorPert-Proof2}), and thus, we have w.h.p.\
\begin{align*}
\|E_{\text{est.}}\| &\leq \tl{O} \left( \frac{y_{\max}}{\sqrt{n}} \sqrt{\E \left[ \left\| M_3(x) M_3^\top(x) \right\| \right]} \right), \\
\|E_{2,\text{est.}} \| &\leq \tl{O} \left( \frac{y_{\max}}{\sqrt{n}} \sqrt{\E \left[ \left\| \Sc_2(x) \Sc_2^\top(x) \right\| \right]} \right),
\end{align*}
where $M_3(x) \in \R^{d \times d^2}$ denotes the matricization of score function tensor $\Sc_3(x) \in \R^{d \times d \times d}$, and $y_{\max}$ is the bound on $|f(x)|=|y|$.

The norm of approximation perturbation $E_{\text{apx.}} := \E[ (\tl{y}_f-y) \cdot \Sc_3(x)]$ is bounded as
\begin{align*}
\left\| E_{\text{apx.}} \right\| &= \left\| \E[ (\tl{y}_f-y) \cdot \Sc_3(x)] \right\| \\
&\leq \E[  \left\| (\tl{y}_f-y) \cdot \Sc_3(x) \right\|] \\
&= \E[ | \tl{y}_f-y | \cdot \left\| \Sc_3(x) \right\|] \\
&\leq \left( \E[ |\tl{y}_f-y|^2 ] \cdot \E [ \left\| \Sc_3(x) \right\|^2 ] \right)^{1/2},
\end{align*}
where the first inequality is from the Jensen's inequality applied to convex norm function, and we used Cauchy-Schwartz in the last inequality. Applying the approximation bound in Theorem~\ref{thm:approx-Barron}, we have
\begin{equation} \label{eqn:TensorApprox-Proof}
\left\| E_{\text{apx.}} \right\|
\leq O(r C_f) \cdot \left(\frac{1}{\sqrt{k}} + \delta_1 \right) \cdot \sqrt{\E [ \left\| \Sc_3(x) \right\|^2 ]},
\end{equation}
and similarly,
$$
\left\| E_{2,\text{apx.}} \right\|
\leq O(r C_f) \cdot \left(\frac{1}{\sqrt{k}} + \delta_1 \right) \cdot \sqrt{\E [ \left\| \Sc_2(x) \right\|^2 ]},
$$

We now need to ensure the overall perturbations $E = E_{\text{est.}} + E_{\text{apx.}}$ and $E_2 = E_{2,\text{est.}} + E_{2,\text{apx.}}$ satisfies the required bounds in Lemma~\ref{lem:perturbation-bound}. 
Note that similar to what we do in Lemma~\ref{lem:A1-guarantee}, we first impose a bound such that the term involving $\|E\|$ is dominant in~\eqref{eqn:tensor-guarantees}.
Bounding the estimation part $\|E_{\text{est.}}\|$ provides similar sample complexity as in estimation Lemma~\ref{lem:A1-guarantee} with $\tl{y}_{\max}$ substituted by $y_{\max}$.

For the approximation error, by imposing (third bound stated in the theorem)
$$
C_f \leq O \left( \frac{1}{r} \left( \frac{1}{\sqrt{k}} + \delta_1 \right)^{-1} \frac{\E [ \left\| \Sc_3(x) \right\|^2 ]^{1/4}}{\E [ \left\| \Sc_2(x) \right\|^2 ]^{3/4}} \cdot \tl{\lambda}_{\min} \cdot s_{\min}^{1.5}(A_1) \right),
$$
we ensure that the term involving $\|E\|$ is dominant in the final recovery error in~\eqref{eqn:tensor-guarantees}.
By doing this, we do not need to impose the bound on $\|E_{2,\text{apx.}}\|$ anymore, and applying the bound in~\eqref{eqn:TensorApprox-Proof} to the required bound on $\|E\|$ in Lemma~\ref{lem:perturbation-bound} leads to bound (second bound stated in the theorem)
$$
C_f \leq
\tl{O}\left( \frac{1}{r} \left( \frac{1}{\sqrt{k}} + \delta_1 \right)^{-1} \frac{1}{\sqrt{\E [ \left\| \Sc_3(x) \right\|^2 ]}} \cdot \lambda_{\min}  \left( \frac{\tl{\lambda}_{\min}}{\tl{\lambda}_{\max}}\right)^{1.5}  s_{\min}^3(A_1) \cdot \frac{1}{\sqrt{k}} \right).
$$
Finally, applying the result of Lemma~\ref{lem:perturbation-bound}, we have the column-wise error guarantees (up to permutation)
\begin{align*}
\|(A_1)_j - (\hA_1)_j \|
&\leq \tl{O} \left( \frac{s_{\max}(A_1)}{\lambda_{\min} } \frac{\tl{\lambda}_{\max}^2}{\sqrt{\tl{\lambda}_{\min}}}  \frac{\| E_{\text{est.}} \| +\|  E_{\text{apx.}} \|}{ \tl{\lambda}_{\min}^{1.5} \cdot s_{\min}^3(A_1)} \right), \\
&\leq \tl{O} \left( \frac{\tl{\lambda}_{\max}^2}{\tl{\lambda}_{\min}^2} \frac{s_{\max}(A_1)}{\lambda_{\min} \cdot s_{\min}^3(A_1)} \left[ \frac{y_{\max}}{\sqrt{n}} \sqrt{\E \left[ \left\| M_3(x) M_3^\top(x) \right\| \right]} \right. \right. \\
& \qquad\qquad\qquad\qquad\qquad\qquad\qquad + \left. \left. r C_f \cdot \left(\frac{1}{\sqrt{k}} + \delta_1 \right) \cdot \sqrt{\E [ \left\| \Sc_3(x) \right\|^2 ]} \right] \right)  \\
&\leq \tl{O} \left( \tl{\epsilon}_1 \right),
\end{align*}
where in the second inequality we substituted the earlier bounds on $\|E_{\text{est.}} \|$ and $\|  E_{\text{apx.}}\|$, 
and the first bounds on $n$ and $C_f$ stated in the theorem are used in the last inequality.


\paragraph{Fourier part:}
Let
$$\tl{v}_f := \frac{1}{n} \sum_{i \in [n]} \frac{(\tl{y}_f)_i}{p(x_i)} e^{-j \inner{\omega_i, x_i}}.$$
Note that this a realization of $\tl{v}$ defined in~\eqref{eqn:v-appendix} when the output is generated by a neural network satisfying approximation error provided in Theorem~\ref{thm:approx-Barron} denoted by $\tl{y}_f$; see the discussion in the beginning of the proof.

The perturbation is now
$$e := \E[\tl{v}_f] - \underbrace{\frac{1}{n} \sum_{i \in [n]} \frac{y_i}{p(x_i)} e^{-j \inner{\omega_i, x_i}}}_{=: v}.$$
Similar to the tensor decomposition part, it can be expanded to estimation and approximation parts as
$$
e := \underbrace{\E[\tl{v}_f] - \E [v]}_{e_{\text{apx.}}}
+ \underbrace{\E[v] - v}_{e_{\text{est.}}}.
$$
Similar to Lemma~\ref{lem:Fourier-guarantee}, the estimation error is w.h.p. bounded as
$$|e_{\text{est.}}| \leq O(\tl{\epsilon}_2),$$
if the sample complexity satisfies $n \geq \tl{O} \left( \frac{\zeta_f}{\psi \tl{\epsilon}_2^2} \right)$, where $\zeta_{f} := \int_{\R^d} f(x)^2 dx$. Notice the difference between $\zeta_f$ and $\tl{\zeta}_{\tl{f}}$.
The approximation part is also bounded as
$$|e_{\text{apx.}}| \leq \frac{1}{\psi} \E[|\tl{y}_f - y|] \leq \frac{1}{\psi} \sqrt{\E[|\tl{y}_f - y|^2]} \leq \frac{1}{\psi} O(r C_f) \cdot \left(\frac{1}{\sqrt{k}} + \delta_1 \right),$$
where the last inequality is from the approximation bound in Theorem~\ref{thm:approx-Barron}. Imposing the condition
\begin{equation} \label{eqn:Cf-bound-fourier}
C_f \leq \frac{1}{r} \left( \frac{1}{\sqrt{k}} +\delta_1 \right)^{-1} \cdot O( \psi \tl{\epsilon}_2)
\end{equation}
satisfies the desired bound $|e_{\text{apx.}}| \leq O(\tl{\epsilon}_2)$. The rest of the analysis is the same as Lemma~\ref{lem:Fourier-guarantee}.

\paragraph{Ridge regression:}
It introduces an additional approximation term in the linear regression formulated in~\eqref{eqn:ridge-regression}. Given the above bounds on $C_f$, the new approximation term only contributes to lower order terms.

Combining the analyzes for tensor decomposition, Fourier and ridge regression parts finishes the proof.
\eprfof

\subsection{Discussion on Corollary~\ref{cor:kernel}} \label{appendix:kernel}

Similar to the specific Gaussian kernel function, we can also provide the results for other kernel functions and in general for positive definite functions as follows.
$f(x)$ is said to be positive definite if $\sum_{j,l} x_i x_j f(x_j-x_l) \geq 0,$ for all $x_j,x_l \in \R^d$.
\citet{Barron93} shows that positive definite functions have $C_f$ bounded as
\begin{equation*} 
C_f \leq \sqrt{-f(0) \cdot \nabla^2 f(0)},
\end{equation*}
where $\nabla^2 f(x) := \sum_{i \in [d]} \partial^2 f(x)/ \partial x_i^2$. Note that the operator $\nabla^2$ is different from the derivative operator $\nabla^{(2)}$ that we defined in~\eqref{eqn:derivativedef}. Applying this to the proposed bound in~\eqref{eqn:Cf-bound}, we conclude that our algorithm can train a neural network which approximates a class of positive definite and kernel functions with similar bounds as in Theorem~\ref{thm:approx-guarantees}. 
Corollary~\ref{cor:kernel} is for the special case of Gaussian kernel function.

\bprfof{Corollary~\ref{cor:kernel}}
For the location and scale mixture $f(x) :=  \int K(\alpha(x+\beta)) G(d\alpha,d\beta)$, we have~\citep{Barron93}
$C_f \leq C_K \cdot \int |\alpha| \cdot |G|(d\alpha,d\beta),$
where $C_K$ denotes the corresponding parameter for $K(x)$. For the standard Gaussian kernel function $K(x)$ considered here, we have~\citep{Barron93}
$C_K \leq \sqrt{d}$, which concludes
$$C_f \leq \sqrt{d} \cdot \int |\alpha| \cdot |G|(d\alpha,d\beta).$$
We now apply the required bound in~\eqref{eqn:Cf-bound} to finish the proof.
But, in this specific setting, we also have the following simplifications for the bound in~\eqref{eqn:Cf-bound}.

For the Gaussian input $x \sim \mathcal{N} (0,\sigma_x^2 I_d)$, the score function is
$$\Sc_3(x) = \frac{1}{\sigma_x^6} x^{\otimes 3} - \frac{1}{\sigma_x^4} \sum_{j \in [d]} \left( x \otimes e_j \otimes e_j +  e_j \otimes x \otimes e_j + e_j \otimes e_j \otimes x \right),$$
which has expected square norm as $\E[\|\Sc_3(x)\|^2] = \tl{O}(d^3/\sigma_x^6)$.

Given the input is Gaussian and  the activating function is the step function, we can also write the coefficients $\lambda_j$ and $\tl{\lambda}_j$ as
\begin{align*}
\lambda_j &= a_2(j) \cdot \frac{1}{\sqrt{2 \pi} \sigma_x^3} \cdot \exp \left( - \frac{b_1(j)^2}{2 \sigma_x^2} \right) \cdot \left( \frac{b_1(j)^2}{\sigma_x^2} - 1 \right), \\
\tl{\lambda}_j &= a_2(j) \cdot \frac{b_1(j)}{\sqrt{2 \pi} \sigma_x^3} \cdot \exp \left( - \frac{b_1(j)^2}{2 \sigma_x^2} \right).
\end{align*}
Given the bounds on coefficients as $|b_1(j)| \leq 1$, $|a_2(j)| \leq 2 C_f$, $j \in [k]$, we have
\begin{align*}
\frac{\tl{\lambda}_{\min}}{\tl{\lambda}_{\max}} &\geq \frac{(a_2)_{\min} \cdot (b_1)_{\min}}{2C_f} \exp(-1/(2 \sigma_x^2)), \\
\lambda_{\min} & \geq \frac{(a_2)_{\min}}{\sqrt{2 \pi} \sigma_x^3} \exp(-1/(2 \sigma_x^2)) \cdot \min_{j \in [k]} |b_1(j)^2/\sigma_x^2 - 1|.
\end{align*}

Recall that the columns of $A_1$ are randomly drawn from the Fourier spectrum of $f(x)$ as described in~\eqref{eqn:random_freq}. Given $f(x)$ is the Gaussian kernel, the Fourier spectrum $\|\omega\| \cdot |F(\omega)|$ corresponds to a sub-gaussain distribution. Thus, the singular values of $A_1$ are bounded as~\citep{rudelson2009smallest}
$$\frac{s_{\min}(A_1)}{s_{\max}(A_1)} \geq \frac{1-\sqrt{k/d}}{1+\sqrt{k/d}} \geq O(1),$$
where the last inequality is from $k=Cd$ for some small enough $C<1$.

Substituting these bounds in the required bound in~\eqref{eqn:Cf-bound} finishes the proof.
\eprfof


\end{document}